\pdfoutput=1

\documentclass[11pt]{article}

\usepackage{acl}
\usepackage{url}

\usepackage{times}
\usepackage{latexsym}

\usepackage{tabularx}
\usepackage{booktabs}
\usepackage{caption}
\usepackage{booktabs}
\usepackage{tabularx} 

\usepackage{makecell}   
\usepackage{multirow}
\usepackage{booktabs}
\usepackage{subcaption}

\usepackage[T1]{fontenc}

\usepackage[utf8]{inputenc}
\usepackage{tcolorbox}

\usepackage{microtype}
\usepackage{hyperref}

\usepackage{inconsolata}

\usepackage{graphicx}
\usepackage{microtype}
\usepackage{subcaption} 
\usepackage{caption} 
\usepackage{booktabs} 

\usepackage{multirow}
\usepackage{array}

\usepackage{algorithm}
\usepackage{algorithmic}
\usepackage{makecell}  

\usepackage{tabularx}
\usepackage{amsmath}
\usepackage{amssymb}
\usepackage{mathtools}
\usepackage{amsthm}
\usepackage{booktabs}  

\usepackage{amsmath}

\title{TinySQL: A Progressive Text-to-SQL Dataset for Mechanistic Interpretability Research}

\author{
\textbf{Abir Harrasse}\textsuperscript{*1,5} \quad
\textbf{Philip Quirke}\textsuperscript{*1} \quad
\textbf{Clement Neo}\textsuperscript{2} \\
\textbf{Dhruv Nathawani}\textsuperscript{3} \quad
\textbf{Luke Marks}\textsuperscript{1} \quad
\textbf{Amir Abdullah}\textsuperscript{4,1} \\
\textsuperscript{1}Martian \quad
\textsuperscript{2}Apart Research \quad
\textsuperscript{3}NVIDIA \\
\textsuperscript{4}Thoughtworks \quad
\textsuperscript{5}Mohammed VI Polytechnic University \\
\textsuperscript{*}\textit{Equal contribution} \\
\textit{Correspondence to:} \texttt{\{philip@withmartian.com, abirharrasse@gmail.com\}}
}

\renewcommand{\thefootnote}{\fnsymbol{footnote}}

\makeatletter
\renewcommand\@makefnmark{}
\makeatother

\begin{document}
\maketitle

\begin{abstract}

\footnote{Proceedings of the 2025 Conference on Empirical Methods in Natural Language Processing (EMNLP 2025)}

Mechanistic interpretability research faces a gap between analyzing simple circuits in toy tasks and discovering features in large models. To bridge this gap, we propose text-to-SQL generation as an ideal task to study, as it combines the formal structure of toy tasks with real-world complexity. We introduce TinySQL, a synthetic dataset, progressing from basic to advanced SQL operations, and train models ranging from 33M to 1B parameters to establish a comprehensive testbed for interpretability. We apply multiple complementary interpretability techniques, including Edge Attribution Patching and Sparse Autoencoders, to identify minimal circuits and components supporting SQL generation. We compare circuits for different SQL subskills, evaluating their minimality, reliability, and identifiability. Finally, we conduct a layerwise logit lens analysis to reveal how models compose SQL queries across layers: from intent recognition to schema resolution to structured generation. Our work provides a robust framework for probing and comparing interpretability methods in a structured, progressively complex setting.
\end{abstract}

\section{Introduction}
The circuit discovery approach in mechanistic interpretability (MI) has largely focused on small models solving simple tasks, such as arithmetic \citep{quirkeunderstanding, nandaprogress} and indirect object identification \citep{wanginterpretability}. While these studies revealed key mechanisms, reliance on toy tasks limits validation and comparison across interpretability methods. Recent work has shifted toward feature discovery using sparse autoencoders (SAEs) \citep{huben2023sparse, templeton2024scaling}, but linking these insights to circuit-level understanding remains difficult without intermediate tasks that support both approaches.

Text-to-SQL generation provides an ideal middle ground, as SQL’s formal structure makes it more tractable than general language generation while still requiring natural language understanding. This enables systematic comparisons of interpretability methods within a task complex enough to reflect real-world challenges.

Existing text-to-SQL datasets like Spider \citep{yu2019spiderlargescalehumanlabeleddataset} and WikiSQL \citep{zhongSeq2SQL2017} are too complex and noisy for rigorous interpretability analysis. To address this, we introduce TinySQL, a curated dataset that enables controlled analysis of how transformers learn and generate SQL queries. TinySQL progresses through text-to-SQL tasks of increasing complexity, isolating key aspects of the generation process while maintaining structural consistency. Alongside the dataset, we train and release models of 33M, 0.5B, and 1B parameters, providing a comprehensive testbed for interpretability research.

Our analysis combines multiple complementary interpretability techniques to study SQL generation. Using Edge Attribution Patching (EAP), we identify minimal circuits supporting SQL query generation. In parallel, our SAE analysis reveals consistent patterns in how the same base model fine-tuned on different tasks yield similar interpretable heads. Lastly, our logit lens analysis reveals a layered computation pattern: early layers capture query intent, middle layers resolve schema elements, and later layers synthesize structured SQL outputs.

\renewcommand{\thefootnote}{\fnsymbol{footnote}}
\renewcommand{\thefootnote}{\arabic{footnote}}

\begin{figure*}[h!]
    \centering
    \includegraphics[width=\textwidth]{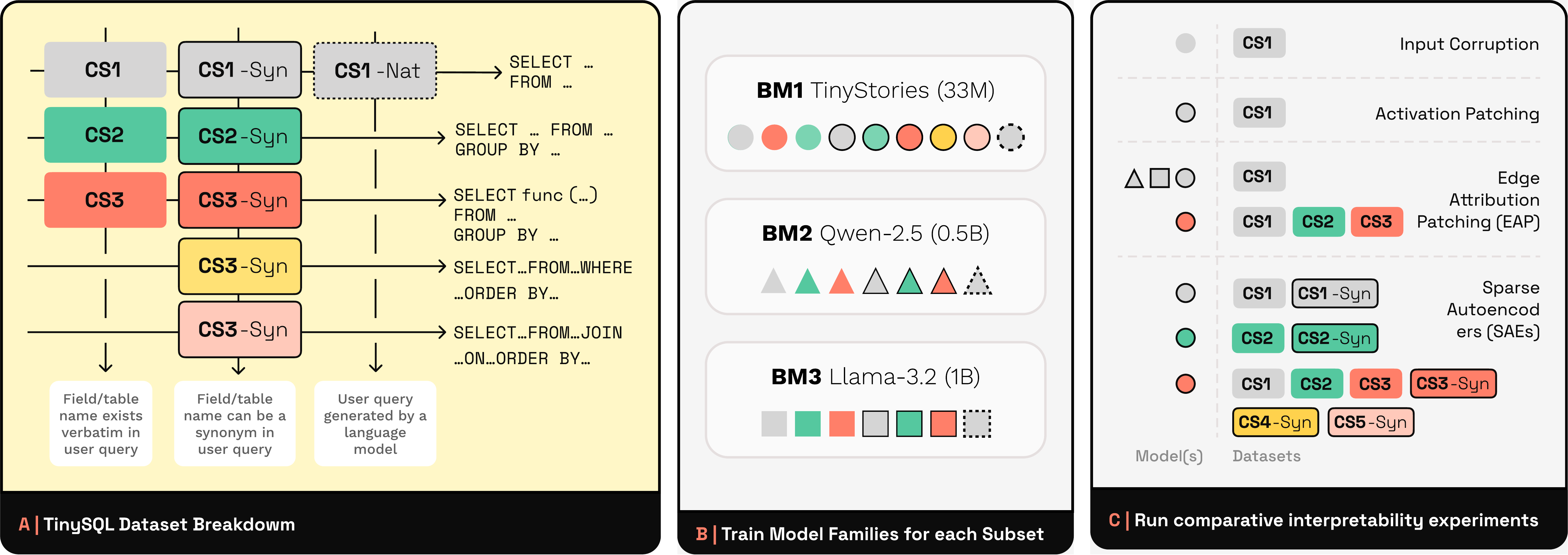}
    \caption{(a) TinySQL is broken down into 9 subsets of varying complexities, across both SQL query and user query axes. (b) We train and release a comprehensive set of models on each dataset subset.(c) We apply MI techniques across various configurations to understand model behavior and compare results.}
    \label{fig:overview_fig1}
\end{figure*}

The contributions of our paper are as follows:


\begin{enumerate}
\item We introduce TinySQL, a structured text-to-SQL dataset bridging toy tasks and real-world applications. By controlling SQL complexity across five subskills and releasing models trained on different subsets of TinySQL, we provide a robust testbed for mechanistic interpretability.

\item We apply multiple interpretability methods (EAP and SAE) to identify circuits across SQL subskills. We compare them, show their local minimality, and validate their relevance through performance and identifiability studies.

\item We present a layerwise logit lens analysis revealing how models compose SQL queries: early layers identify intent, middle layers resolve schema references, and later layers integrate all elements into structured outputs.

\end{enumerate}

Our work advances understanding of neural networks in structured query processing and outlines a roadmap for future MI studies. By identifying limitations and breaking points, we define clear boundaries for reliable application and highlight areas needing new approaches.

\footnote{Project website with code, data, models and viewers at: \url{https://abirharrasse.github.io/tinysql/}}

\section{Background}
\subsection{Mechanistic Interpretability Approaches}
Early mechanistic interpretability (MI) research focused on discovering specific computational circuits in small language models through carefully controlled tasks. These studies identified mechanisms for indirect object identification \citep{wanginterpretability}, docstring generation \citep{docstring_circuit}, and number incrementation \citep{nanda2023progress}, but remained limited by their reliance on highly specialized tasks. However, these investigations remained limited to highly specific tasks that offered few natural paths for extension. While automated methods like ACDC \citep{conmy2023towards} and EAP \citep{syed2023attribution} helped automate circuit discovery, interpreting these circuits still required extensive manual analysis. 

Beyond mechanistic interpretability, recent work has leveraged synthetic languages and controlled data to probe model reasoning. \citet{structured_1} demonstrate that pre-training on artificial language sequences with explicit and implicit token dependencies improves downstream performance. \citet{structured_2} use CFG-based synthetic grammars to study recursive reasoning, revealing interpretable internal structures and algorithmic attention patterns in generative models. These results highlight the value of structured tasks, and support our use of Text-to-SQL as a semantically grounded, interpretable setting.

Text-to-SQL generation offers a unique middle ground for advancing MI research. The task combines the formal structure needed for circuit analysis with the semantic complexity of natural language understanding, making it an ideal testbed for bridging interpretability methods across scales.

Recent work with SAEs \citep{huben2023sparse, templeton2024scaling} has enabled large-scale analysis of model behavior. \citet{marks2024sparse} use SAE features to uncover interpretable circuits for tasks like subject-verb agreement, demonstrating the promise of this approach. However, such studies often focus on a single model and task, leaving questions about generalization unresolved. To address this, we introduce multiple Text-to-SQL variants across models, enabling systematic investigation of behavioral variation.

\subsection{Text-to-SQL Generation}
Text-to-SQL is a task where models generate SQL queries from natural language requests. 
Given a database schema and a query like ``Show me all employee salaries'', the model must produce the correct SQL query ``\texttt{SELECT salary FROM employees}''.

Text-to-SQL is a step up from toy tasks in MI research while retaining structure for rigorous analysis. Each query maps to a single answer, allowing clear evaluation, unlike general code generation. The task also exhibits systematic patterns, such as ``how many” mapping to \texttt{COUNT} and “highest” to \texttt{ORDER BY DESC}.

\subsection{Text-to-SQL Datasets}

The field evolved from simpler beginnings with \texttt{WikiSQL} \citep{zhongSeq2SQL2017}, which focused on single-table queries from Wikipedia to enable large-scale evaluation. \texttt{Spider} \citep{yu2019spiderlargescalehumanlabeleddataset} established a foundation for text-to-SQL research by focusing on cross-domain generalization through multiple tables and schemas. Its variants explore different aspects of complexity: \texttt{Spider-Syn} \citep{spider-syn} tests semantic understanding by replacing schema terms with synonyms, while \texttt{Spider-Real} \citep{spider_real} aligns more closely with natural user queries by omitting explicit column names. These datasets collectively enable research into model robustness across synthetic and real-world scenarios. 

Recent text-to-SQL datasets have advanced the field by incorporating conversational structure and domain-specific challenges. CoSQL \citep{more_text_to_sql_cosql} introduces multi-turn interactions grounded in real user dialogues. SParC \citep{more_text_to_sql_sparc} similarly emphasizes context-dependent SQL generation, where models must resolve references across sequential questions in cross-domain settings. In the medical domain, MIMICSQL \citep{more_text_to_sql_treqs} features naturally varied questions about electronic health records, including paraphrasing and under-specification. 

Alongside these realistic benchmarks, recent efforts like Gretel's \texttt{ Synthetic-Text-To-SQL} \citep{gretel-synthetic-text-to-sql-2024} and \texttt{SQL-create-context} \citep{b-mc2_2023_sql-create-context} seek to synthesize datasets that capture naturalistic query patterns while retaining control over generation. These works reflect growing emphasis on bridging the gap between controlled data creation and real-world applicability.

However, these existing datasets prioritize training high-performance models through diverse, complex queries rather than supporting controlled experiments. Even synthetic resources like \texttt{Sythnetic-Text-To-SQL} lack the systematic progression of complexity needed for mechanistic interpretability research. This gap motivates our development of TinySQL, which draws inspiration from machine learning's long history of using synthetic data to control task complexity and enable clear evaluation. By providing carefully controlled progression of query complexity while maintaining consistent structure, TinySQL bridges the divide between performance-focused datasets and the controlled environment needed for interpretability research.

\section{The TinySQL Dataset}

To enable rigorous MI research, we need datasets that progress systematically from toy-like tasks suitable for circuit analysis to realistic tasks that capture core text-to-SQL challenges. Each level must provide enough examples for both model training and detailed experimentation. TinySQL implements this through two independent axes: SQL command complexity and language variation.

This design enables comparative analysis - training models on different complexity levels to study base task handling or examining how language variation impacts SQL generation.

\subsection{Dataset Structure}

TinySQL structures tasks along two dimensions: SQL complexity and query language variation, isolating specific model behaviors while maintaining consistent task patterns.

\paragraph{SQL Command Levels.} Tasks are organized into five levels of increasing SQL complexity:
\begin{itemize}
\item \textbf{CS1 (Basic Queries)} focuses on fundamental \texttt{SELECT-FROM} operations. Example: ``Show me the salary from employees'' → \texttt{SELECT salary FROM employees}

\item \textbf{CS2 (Ordered Queries)} introduces \texttt{ORDER BY} clauses. Example: ``Show employee salaries from highest to lowest'' → \texttt{SELECT salary FROM employees ORDER BY salary DESC}

\item \textbf{CS3 (Aggregation Queries)} adds aggregation functions and grouping. Example: ``How many employees are in each department?'' → \texttt{SELECT COUNT(id) FROM employees GROUP BY department}

\item \textbf{CS4 (Filter Queries)}adds type-aware \texttt{WHERE} clauses to 80\% of queries. Example: ``Find entries where salt contains 'z' ordered by score and location'' → \texttt{\detokenize{SELECT salt, max_score FROM data WHERE salt LIKE '\%z\%' ORDER BY max_score DESC, location DESC}}

\item \textbf{CS5 (Join Queries)} introduces multi-table queries with \texttt{JOIN} clauses on matching column types. Example: ``List workflow templates joined with pairs on user ID ordered by score and certification'' → \texttt{\detokenize{SELECT name, city FROM workflow_templates JOIN pairs ON workflow_templates.user_id = pairs.user_id ORDER BY average_score ASC, certification ASC}}.

\end{itemize}

Examples of each Command Level are presented in Appendix.\ref{app:data-examples}

\paragraph{Query Language Variants.} For each command level, we provide three variants of increasing linguistic complexity:
\begin{itemize}
\item The \textbf{Base (CSx)} variation uses rigid templates where field and table names exactly match the schema, providing a controlled baseline for studying SQL generation mechanisms.

\item The \textbf{Synonyms (CSx\_Syn)} variation introduces semantic mapping between query terms and schema fields (e.g., ``earnings'' mapping to ``salary''), testing models' ability to handle semantic equivalences. This is inspired by synonym-based datasets like \texttt{Spider-Syn} \citep{spider-syn} and \texttt{ADVETA} \citep{adveta}. 

\item The \textbf{Natural (CSx\_Nat)} variation allows flexible natural language phrasing while targeting the same SQL operations. This most closely resembles real-world usage, and it is currently limited to CS1 queries due to the difficulties of ensuring dataset quality.
\end{itemize}

\textbf{Dataset statistics.}
The complete dataset consists of nine task variants (\texttt{CS1/2/3}, \texttt{CS1/2/3/4/5\_Syn}, and \texttt{CS1\_Nat}), each containing 100,000 examples generated through a systematic data creation pipeline that ensures consistency with our design principles. The average query length using the \texttt{meta-llama/Llama-3.2-1B-Instruct} tokenizer ranges from 16.48 tokens for CS1\_Syn, 38.76 tokens for CS2\_Syn, 42.39 tokens for CS3\_Syn, 53.73 tokens for CS4\_Syn and 61.61 tokens for CS5\_Syn, with similar figures for the non synonym variants.

Employing the approach of \citet{dataset_analysis}, we compare the English portion of TinySQL to prior Text-to-SQL datasets and find it to be both lexically and semantically more challenging than prior datasets. This is reflected in three key metrics: (a) rarity, the ratio of rare words to content words in a natural language (NL) question \cite{rarity}; (b) lexical density, the ratio of content words to total words \cite{lexical_density}; and (c) Flesch reading ease, a readability measure where lower scores indicate more complex sentences with longer structures and more syllables per word on average \cite{flesch}. As shown in Table~\ref{tab:cs_synonyms_stats}, TinySQL scores higher on rarity and lexical density, and lower on readability, compared to most previous Text-to-SQL datasets.

\begin{table}[h!]
\centering
\small
\begin{tabular}{lccc}
\toprule
\textbf{Dataset} & \textbf{Rarity} & \textbf{Lexical Density} & \textbf{Readability} \\
\midrule
CS1\_Syn   & \textbf{0.60} & 0.58 & 45.5 \\
CS2\_Syn & 0.52 & 0.56 & 29.4 \\
CS3\_Syn & 0.45 & \textbf{0.61} & 29.9 \\
CS4\_Syn & 0.41 & 0.58 & 28.7 \\
CS5\_Syn & 0.44 & 0.59 & \textbf{25.7} \\
IMDb            & 0.23 & 0.43  & 88.0 \\
Yelp            & 0.28 & 0.51  & 85.0 \\
Academic        & 0.24 & 0.54  & 65.0 \\
Spider          & 0.21 & 0.55  & 89.0 \\
Geo             & 0.29 & 0.58  & 95.0 \\
Restos          & 0.26 & 0.56  & 90.0 \\
MIMICSQL        & 0.25 & 0.61  & 68.0 \\
K-DBQA          & 0.26 & 0.59  & 75.0 \\
BIRD            & 0.30 & 0.57  & 73.0 \\
Atis            & 0.35 & 0.46  & 78.0 \\
Scholar         & 0.31 & 0.52  & 77.0 \\
Advising        & 0.25 & 0.44  & 80.0 \\
\bottomrule
\end{tabular}
\caption{Dataset Statistics Comparison to Previous Text-to-SQL Datasets}
\label{tab:cs_synonyms_stats}
\end{table}

For details on our dataset generation methodology, including schema design, query construction, and instruction templating for \texttt{CSx}, \texttt{CSx\_Syn}, and \texttt{CSx\_Nat}, please refer to Appendices \ref{app:csx-method} and \ref{app:cs-nat-method}.

\subsection{Models Trained}
\label{sec:models_trained}
We fine-tune three base models to perform the text-to-SQL task. We call them BM1, BM2, and BM3:
\begin{itemize}
    \item \textbf{BM1:} \texttt{TinyStories-33M} \citep{eldan2023tinystoriessmalllanguagemodels}, a 2-layer model with 33M parameters.
    \item \textbf{BM2:} \texttt{Qwen2.5-0.5B-Instruct} \citep{qwen2.5}, a model of approx. 500M parameters.
    \item \textbf{BM3:} \texttt{Llama-3.2-1B-Instruct} \citep{grattafiori2024llama3herdmodels}, a 1B-parameter Llama variant.
\end{itemize}
Our datasets comprise five SQL complexity levels (\texttt{CS1}, \texttt{CS2}, \texttt{CS3}, \texttt{CS4} and \texttt{CS5}), combined with either base or synonym versions for the first three variants, resulting in nine variations (\texttt{CS1}, \texttt{CS2}, \texttt{CS3}, \texttt{CS1\_Syn}, \texttt{CS2\_Syn}, \texttt{CS3\_Syn}, \texttt{CS4\_Syn}, \texttt{CS5\_Syn} and \texttt{CS1\_NAT}. We use shorthand like \texttt{BM1\_CS1} and \texttt{BM1\_CS1\_Syn} for \texttt{BM1} fine-tuned on \texttt{CS1} base and synonym versions, extending similarly to other variants. We exclude \texttt{CS1\_NAT} from all trainings. We train BM1 on all 8 remaining variants, while we train BM2 and BM3 only on the 6 CS1 through to CS3 variants. This produces 20 fine-tuned checkpoints. Refer App.\ref{app:TrainedModels} for training and architecture details, and App.\ref{app:eap-features} for examples of the model input format.

\paragraph{Performance Overview.}

All three base models reliably converge on all dataset variants. For example, fine-tuned \texttt{BM1} achieves over 85\% exact-match accuracy on CS3\_Syn, the most challenging variant, while \texttt{BM2} and \texttt{BM3} exceed 98\% on most tasks. See App.\ref{app:TrainedModels} and Tab.\ref{tab:TrainedModels} for details. These results show that even smaller transformer architectures can learn robust text-to-SQL generation on TinySQL, providing a strong foundation for MI analysis.

\section{Experiments}
Our experiments aim to uncover how language models internally represent and execute structured code generation tasks, using interpretability tools such as Edge Attribution Patching, Sparse Autoencoders, and Logit Lens analysis.

\begin{figure*}[htbp]
    \centering
    \includegraphics[width=\textwidth]{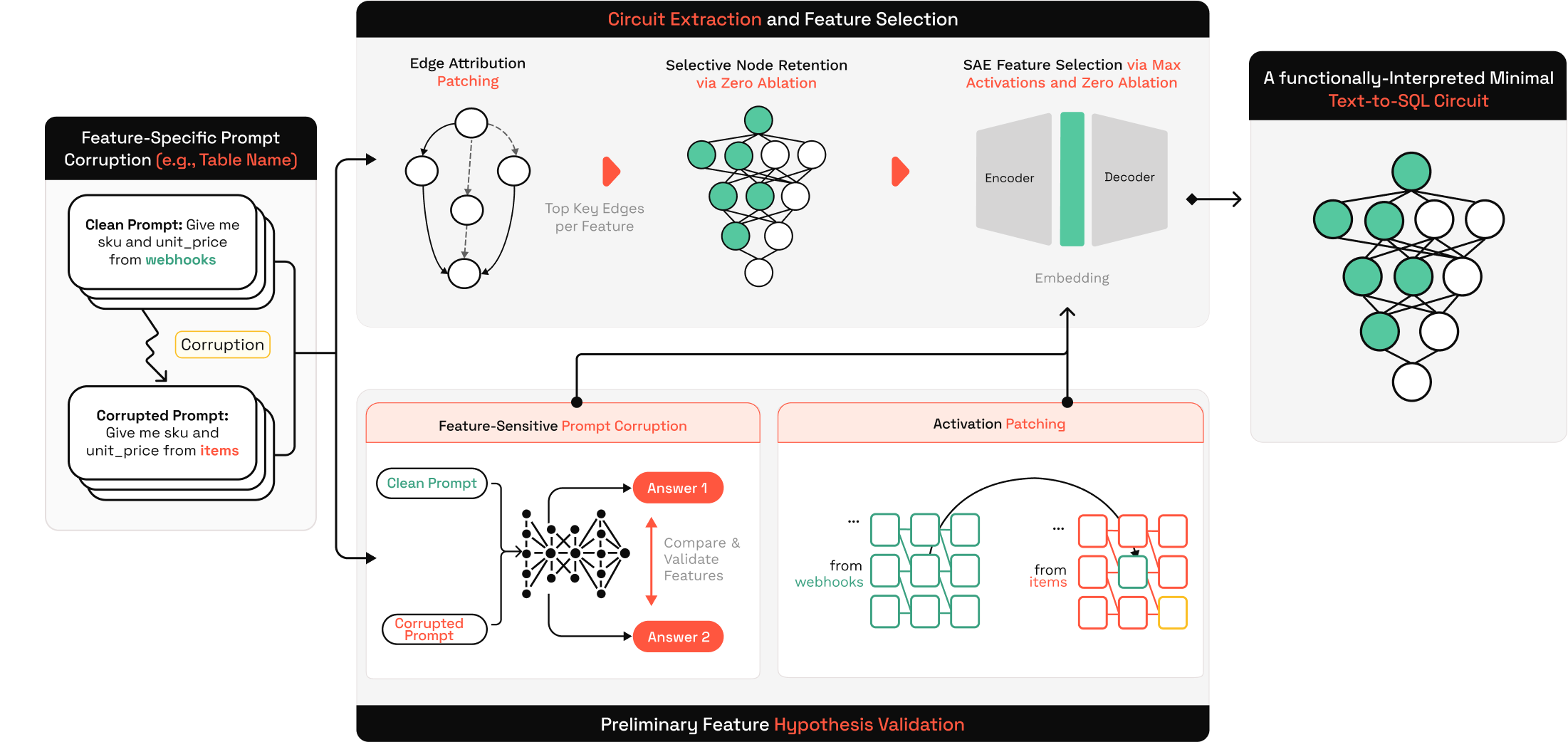}
    \caption{To extract and interpret text-to-SQL functionality, we use Edge Attribution Patching to identify key connections, Selective Node Retention to create a minimal working circuit, and SAE Feature Selection to interpret node functionality. We also use prompt corruption and activation patching to form hypotheses on how the model implements functions.}
    \label{fig:three_stage_process}
\end{figure*}

\subsection{Mechanistic Interpretability Techniques}
In circuit analysis, we view the model as a computational graph $G=(V,E)$, where circuits are subgraphs implementing specific functions \citep{conmy2023towards, wu2024interpretabilityscaleidentifyingcausal}. The resolution of these nodes varies from layers, to sublayers (attention, MLP), and to individual components (attention head, MLP neuron). 



\textbf{Edge Attribution Patching (EAP).} Given clean input $x_c$ and corrupted input $x_d$, EAP measures how each edge $E$ in the computational graph affects the model's behavior. For edge $E$ with clean activation $e_c$ and corrupted activation $e_d$, we approximate its causal effect as $\Delta_E L = (e_d - e_c) \cdot \nabla_e L(x_c)$, where $L$ is the task-specific metric. The absolute attribution score $|\Delta_E L|$ ranks edge importance, allowing identification of the most crucial edges for a given task \citep{syed2023attribution}. The subnetwork is typically obtained by keeping the top $k$ edges, where $k$ is a chosen hyperparameter.

\textbf{Sparse Autoencoder Features.} Neural networks often learn to encode multiple distinct features in overlapping directions, which makes their internal representations difficult to interpret. Sparse Autoencoders (SAEs) address this by learning an encoder $E: \mathbb{R}^n \rightarrow \mathbb{R}^m$ and decoder $D: \mathbb{R}^m \rightarrow \mathbb{R}^n$ that reconstruct model activations $x$ while enforcing sparsity in the latent space, minimizing $|x - D(E(x))|^2$ under appropriate constraints. This sparsity forces the autoencoder to learn a basis set of features that activate independently and rarely, which often correspond to human-interpretable concepts \citep{huben2023sparse}.

\subsection{Basic SQL Generation}
\label{sec:basic-sql-exp}

\paragraph{Input Corruption.} To probe how \texttt{BM1\_CS1} uses context, we replace table or field names in either the instruction or schema (e.g., "table\_a" vs. "table\_b"). If performance remains stable, the model likely ignores the corrupted source.

We find the model relies on schema context for table names and on the instruction for field names, an effective shortcut in CS1, where names match. To prevent this, we created synonym variants (\texttt{CSx\_Syn}) that require mapping between equivalent terms. We now focus on these variants to study more realistic behaviors.



\subsection{Finding Locally Minimal Text-to-SQL Circuits via EAP and Ablation}
\label{sec:minimal_circuits}

To identify the locally-minimal circuits responsible for SQL generation, we analyzed how the model process components like table name, field names, and \texttt{ORDER BY} clauses. We generated 15 batches of 100 paired clean and corrupted prompts for each SQL feature (full set of features detailed in App.~\ref{app:eap-features}). We then performed EAP, selecting the 10 most important edges ranked by the task-specific metric $L$, which produces scores near 0 for essential edges and near 1 for less relevant ones.

\begin{equation}
\label{eq:patch-loss}
L = \frac{\ell(x_\text{clean}|\text{do}(E=e_\text{patch})) - \ell(x_\text{corr})}{\ell(x_\text{clean}) - \ell(x_\text{corr})}
\end{equation}

For each feature, we ran EAP four times across all combinations of semantic field and table name variations. While we obtain 10 edges per batch, we retain only the edges that appear across batches consistently,  requiring 100\% consistency for CS1 and CS2, and 80\% for CS3 due to its increased complexity. The full results appear in Figures \ref{fig:eap_grid_cs1}, \ref{fig:eap_grid_cs2}, and \ref{fig:eap_grid_cs3} in the appendix. After running EAP for all features, we obtain a set of edges per feature. We then take the set of attention heads associated with these edges across all features, and retain them during the following ablation study.

\begin{table*}[t]
\centering
\begin{tabular}{l|l|cc|c}
\toprule
\textbf{Model} & \textbf{Dataset} & \multicolumn{2}{c|}{\textbf{Heads}} & \textbf{Recovered Accuracy} \\
\cmidrule(r){3-4}
 & & \textbf{Layer 0} & \textbf{Layer 1} &  \\
\midrule
BM1\_CS1\_Syn & CS1 & 11,3,1,8,15,14,13,7 & 13,3,7,14,15,11 & 85.4\% \\
BM1\_CS3\_Syn & CS1& 11,14,3,8,2,7,0,9,6,13,5,1 & 2,3,5,15,11,8,13 & 97.6\% \\
BM1\_CS3\_Syn & CS2 & 11,3,7,14,2,8,13,1,10,4 & 3,2,15,5,11,8 & 74.6\% \\
BM1\_CS3\_Syn & CS3 & 11,14,3,0,13,2,7,6,4,1 & 2,15,3,8,1,7,5,11 & 78.3\% \\
\midrule
& & \multicolumn{2}{c|}{\textbf{Layers}} & \\
\cmidrule(r){3-4} 
BM2\_CS1\_Syn & CS1 & \multicolumn{2}{c|}{21,8,20,9,0,1,13,16,11,14,7,18,15,23} & 71.6\% \\
BM3\_CS1\_Syn & CS1 & \multicolumn{2}{c|}{0,5,4,14,8,6,1,9,11,10,15,13,7,12} & 82\% \\
\bottomrule
\end{tabular}
\caption{Minimal circuits identified for Text-to-SQL tasks, showing the attention heads found using EAP and their recovered accuracy. For BM1 models, heads are split between Layer 0 and Layer 1, while for BM2 and BM3 they are reported across all layers.}
\label{tab:eap_minimal_circuits}
\end{table*}

\paragraph{Attention Head Ablation.} To validate our findings, we ablated all attention heads except those linked to EAP-identified edges and measured Recovered Accuracy. We tested both mean- and zero-ablation (the latter being more informative). Figure~\ref{fig:ablation_results} shows how performance varies under selective component removal.


Table~\ref{tab:eap_minimal_circuits} shows that EAP performed best on smaller BM1 models, identifying precise task-relevant heads. In larger models (BM2/BM3), top edges often involved MLPs, leading to few head selections and low recovered accuracy (5–10\%). To compensate, we retained entire layers when any head was marked important, improving accuracy at the cost of granularity. In the next section, we address this limitation using more targeted SAE circuits.

\subsection{Sparse Circuit Identification: Method and Results}

To identify interpretable latent features relevant to SQL task performance, we train Sparse Autoencoders (SAEs) on attention head output and MLP activations from the decoder blocks of our transformer models. We focus our main analysis on the BM1 family spanning all the datasets, with results for BM2 provided in Appendix.\ref{app:saes_circuits_res}.

\paragraph{Experimental Setup}

We begin by collecting attention heads and MLP activations from each of $\texttt{BM}_2\_\texttt{CS}_1\_\texttt{Syn}$ and $\texttt{BM}_1\_\texttt{CS}_i\_\texttt{Syn}$ models when applied to its corresponding $\texttt{CS}_i$ datasets, for $i \in \{1, 3, 4, 5\}$ . To analyze how models trained on more complex tasks generalize to simpler ones, we also create a \textit{descending complexity} evaluation set by applying $\texttt{BM1\_CS3\_Syn}$ to $\texttt{CS1}$ and $\texttt{CS2}$ data.

We train SAEs on the collected activations using the \citet{sparsify}'s package, following the top-$k$ training regime \citet{gao2024scaling} , with $k=16$, an expansion factor of 4, and a learning rate of $8 \times 10^{-4}$ for 6 training epochs. Across all models and datasets, we find that the Fraction of Unexplained Variance (FVU) remains below 0.05, indicating that the SAEs faithfully reconstruct the input representations. Training each SAE takes approximately 1 hour for BM1 and 2 hours for BM2, all on a single A100 GPU.

\paragraph{Feature Selection via AUC Scoring.}

To isolate task-relevant features, we follow the four-step procedure in Algorithm~\ref{alg:sae_auc}.  We evaluate each circuit’s fidelity using zero-ablation and report recovered accuracy in Table~\ref{tab:layerwise-heads}, with additional results on other combinations in Appendix.~\ref{app:saes_circuits_res}.

\begin{algorithm}[t]
\caption{SAE-Based Circuit Identification via Similarity-Guided Feature Selection and AUC Mapping}
\label{alg:sae_auc}
\begin{algorithmic}[1]

\STATE \textbf{Collect Activations:} 
Extract residual activations $\{x^{(j)}\}_{j=1}^N$ at hook $h$ and position $n$

\STATE \textbf{Compute SAE Feature Activations:} 
Project activations through SAE encoder to obtain sparse codes $\{z^{(j)}\}$

\STATE \textbf{Similarity-Guided Feature Selection:}  
Iterate $k = 1$ to $K_{\max}$: compute diff vector $\Delta_k$ by ablating lower-ranked features, apply $\Delta_k$ to the model, and measure output similarity.  
Stop when similarity $\geq$ threshold or improvement flattens.  
Set $k^\star = k$

\STATE \textbf{Map to Model Components via AUC Attribution:}  
Let $a_f = \lVert W_{\text{dec}}[f] \rVert$ for each selected feature $f \in F$, and $A = \sum_{f \in F} a_f$.  
Select smallest $m$ such that:
\[
m^\star = \operatorname*{argmin}_m \left| \frac{\sum_{i=1}^{m} a_i}{A} - \tau \right|
\]

\end{algorithmic}
\end{algorithm}

\begin{table*}[t]
\centering
\begin{tabular}{l|l|cc|c}
\toprule
\textbf{Model} & \textbf{Dataset} & \multicolumn{2}{c|}{\textbf{Heads and MLPs}} & \textbf{Recovered Accuracy} \\
\cmidrule(r){3-4}
 & & \textbf{Layer 0} & \textbf{Layer 1} &  \\
\midrule
\multirow{2}{*}{BM1\_CS1\_Syn} & \multirow{2}{*}{CS1} & 13, 8, 1, 10, 15, 3, 14, 9, 11 & -- & \multirow{2}{*}{80.64\%} \\
 & & All MLP neurons & -- & \\
\midrule
\multirow{2}{*}{BM1\_CS3\_Syn} & \multirow{2}{*}{CS1} & 13, 8, 1, 15, 10, 12, 14, 7, 9, 3 & -- & \multirow{2}{*}{85.71\%} \\
 & & 846 MLP neurons & -- & \\
\midrule
\multirow{2}{*}{BM1\_CS3\_Syn} & \multirow{2}{*}{CS3} & 13, 8, 1, 15, 10, 12, 14, 9, 7, 3 & -- & \multirow{2}{*}{87.67\%} \\
 & & All MLP neurons & -- & \\
\bottomrule
\end{tabular}
\caption{Recovered accuracy and layer-wise head selections for circuits identified via high-AUC SAE features. Each layer column is split into two rows per model (first row for attention heads and second for MLP neurons). Full results are in Appendix.~\ref{app:saes}.}
\label{tab:layerwise-heads}
\end{table*}

\section{Identifiability and Local Minimality of Derived Circuits}

\subsection{Identifiability via SAE-EAP Overlap}

To assess the identifiability of circuits extracted using sparse autoencoders (SAEs), we compare them against circuits identified by Edge Attribution Patching (EAP). We quantify overlap between the two methods and contrast it with a random baseline. This comparison provides a new identifiability measure for SAE circuits.

Our results in Table \ref{tab:overlap} reveal a consistently higher overlap between EAP and SAE circuits than between EAP and random circuits (for which the accuracies are presented in Appendix.\ref{app:eap_vs_random}). This supports the hypothesis that SAE-selected features align with the ground truth structure identified via causal intervention. Notably, Layer 2 exhibits more variability and lower overlap across methods, which is consistent with our earlier ablation study (Fig.\ref{fig:all_ablation_results}) indicating its minor role in SQL task processing, often involving only a handful of heads.

\begin{table}[h]
\centering
\begin{tabular}{l p{1.2cm} c c}
\toprule
\textbf{Model} & \textbf{Dataset} & \makecell{\textbf{Layer} \\ \textbf{0}} & \makecell{\textbf{Layer} \\ \textbf{1}} \\
\midrule
\multirow{2}{*}{BM1-CS1-Syn} 
    & CS1 & 87.5\% & 50\% \\
    &     & 52.62\%         & --   \\
\midrule
\multirow{2}{*}{BM1-CS3-Syn} 
    & CS1 & 70\%     & 57\% \\
    &     & 74.85\%  & --   \\
\midrule
\multirow{2}{*}{BM1-CS3-Syn} 
    & CS2 & 70\%     & 50\% \\
    &     & 54.81\%  & --   \\
\midrule
\multirow{2}{*}{BM1-CS3-Syn} 
    & CS3 & 50\%     & 62.5\% \\
    &     & 36.16\%  & --     \\
\bottomrule
\end{tabular}
\caption{Percentage overlap between EAP circuits and other methods across different datasets and layers. 
For each model and dataset, the first row shows overlap between SAE and EAP circuits, and the second row shows overlap between random (Rand) and EAP circuits}
\label{tab:overlap}
\end{table}

\subsection{Local Minimality of Circuits}

We also analyze the local minimality of the circuits (i.e., the property that no smaller circuit can be obtained by removing or editing components of the current ``locally minimal") circuits identified by SAEs, by measuring the percentage of the total model they utilize. This metric reflects how sparse and efficient the discovered subcircuits are in reproducing task behavior. Table~\ref{tab:minimality} summarizes these results across different combinations of models and datasets.

\begin{table}[h]
\centering
\begin{tabular}{l l c}
\toprule
\textbf{Model} & \textbf{Dataset} & \makecell{\textbf{Percentage of} \\ \textbf{Model Used}} \\

\midrule
BM2-CS1-Syn & CS1 & 69.94\% \\
BM1-CS1-Syn & CS1 & 42.69\% \\
BM1-CS3-Syn & CS1 & 37.95\% \\
BM1-CS3-Syn & CS2 & 42.69\% \\
BM1-CS3-Syn & CS3 & 43.75\% \\
BM1-CS4-Syn & CS4 & 44.79\% \\
BM1-CS5-Syn & CS5 & 34.83\% \\
\bottomrule
\end{tabular}
\caption{Percentage of Model Components Used by SAE Circuits}
\label{tab:minimality}
\end{table}

We also explored an alternative circuit extraction strategy based on mean ablation . While this approach produced more minimal circuits, their effectiveness was comparable to random circuits of the same size, suggesting poor alignment with ground-truth task structure. Thus, we opted not to include mean-ablated circuits in our main results. Detailed examples and ablation configurations are provided in Appendix.~\ref{appendix:mean_circuits}.

For example, in \texttt{BM1}-\texttt{CS3-Syn} $\rightarrow$ \texttt{CS1}, the found circuit recovered 88.63\% accuracy while using only 22.5\% of model components. Other cases show similar patterns, with best circuits retaining high performance using only ~20–30\% of the model.

Finally, on \texttt{BM2}-\texttt{CS1-Syn} $\rightarrow$ \texttt{CS1}, focusing only on attention heads yielded circuits using just 3.17\% of the model’s heads while maintaining high accuracy. However, including MLPs raised usage to 60.71\%, indicating that while attention sparsity is possible, full circuit function often relies on broader model interaction.

\section{Two-Phase SQL Generation: Intent First, Grounding Later}

To better understand how the model constructs SQL queries, we apply \textbf{LogitLens} analysis~\cite{nostalgebra2023logitlens}. We observe a clear two-phase behavior in the generation process.

In the first phase, the model focuses on identifying the \textbf{intent} of the query by emitting appropriate \textit{SQL keywords} (e.g., \texttt{SELECT}, \texttt{WHERE}, \texttt{GROUP BY}). These tokens receive higher logit probabilities in the earlier layers of the model, suggesting an early commitment to the structural form of the SQL command.

Subsequently, the model transitions into a second phase where it determines the appropriate \textbf{table names} to include in the query. This transition is evident from the delayed rise in logit probabilities for table-related tokens, which occurs at deeper layers, indicating the model is leveraging contextual information from the prompt to ground the query in schema elements.

\begin{figure}[h]
    \centering
    \includegraphics[width=\linewidth]{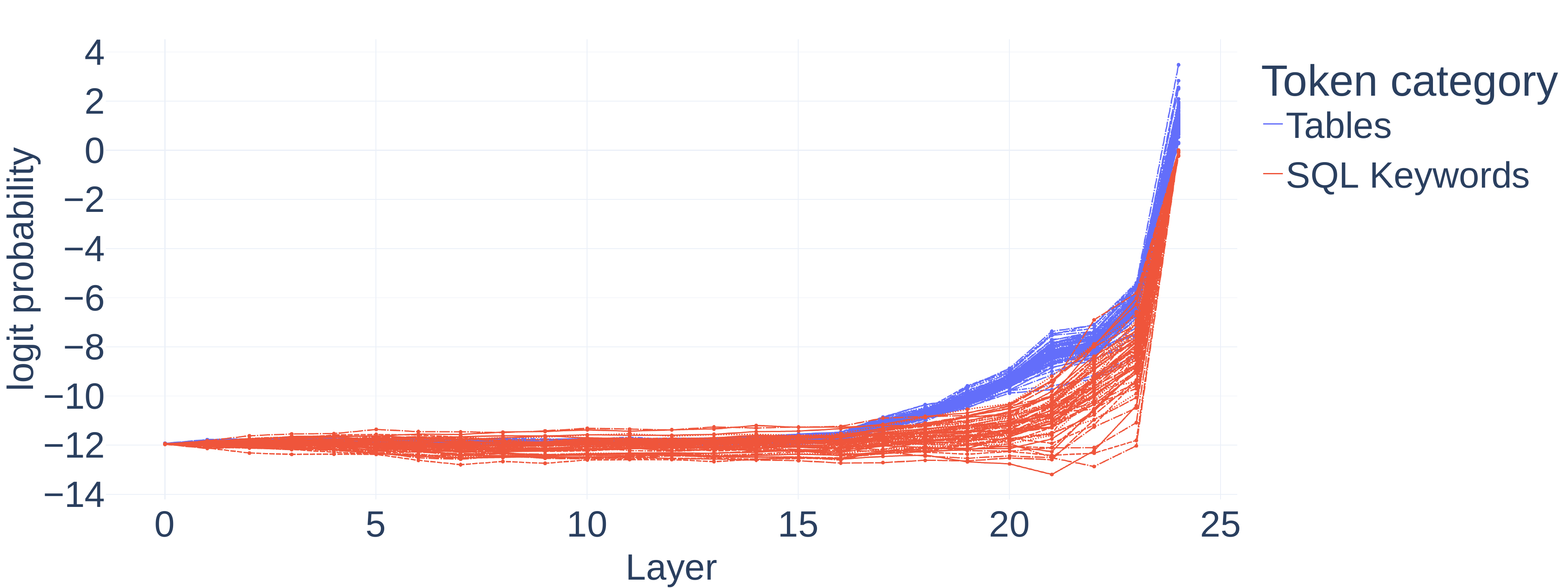}
    \caption{Emergence of SQL intent (via keywords) precedes table name resolution during generation in BM2-CS3. LogitLens reveals elevated logit probabilities for SQL keywords early, followed by table tokens at deeper layers.}
    \label{fig:logitlens-sql-keywords}
\end{figure}

We provide additional visualizations demonstrating similar trends across other benchmarks and model configurations in Appendix.~\ref{appendix:logitlens} and a detailed logit lens analysis illustrating the two-phase token emergence across layers in Appendix.~\ref{appendix:logit-lens-case-study}

\section{Discussion}

\subsection{Ablation Validates Robustness–Granularity Tradeoff}

Our ablation results reveal that zero-ablation recovers performance significantly better than random circuits, making it more reliable for identifying robust, task-relevant components. In contrast, mean-ablation yields smaller circuits but recovers performance similar to random baselines, suggesting weaker causal validity. Thus, zero-ablation is better for robustness, while mean-ablation may be useful for identifying compact, interpretable circuits when some loss in reliability is acceptable. Notably, across prompts within all classes, activation patching results vary significantly, even among prompts in the same class, highlighting the inherent variability in the circuits identified (see Appendix.\ref{app:ap-graphs} for examples).

\subsection{Distributed Computation in SQL Generation}

LogitLens analysis shows that SQL generation unfolds compositionally: early layers capture intent, middle layers resolve schema references, and later layers assemble the final query. However, EAP often reveals fragmented mechanisms scattered across layers, especially in larger models, highlighting the difficulty of isolating clean circuits and the limits of current interpretability tools at scale. This fragmentation is consistent with the observed variability in activation patching results, suggesting that circuit-level interpretations may differ notably even for similar inputs.

\section{Conclusion}
We present TinySQL, a dataset with increasingly complex SQL subsets and models of varying sizes. While not a replacement for real-world data, it supports controlled circuit discovery. Our analysis of EAP, SAE, and Logit Lens reveals both strengths and limitations, offering a testbed for advancing interpretability methods.

\section*{Acknowledgements}
 
We thank Narmeen Oozeer for her invaluable early feedback on the project and for proposing Edge Attribution Patching as a linear approximation of the ACDC idea. We are also grateful to Shriyash Upadhyay for his thoughtful feedback and assistance with communication around the project. We appreciate Antia Garcia Casal for her help with figures and for shaping the overall style of our visualizations.

 \section*{Limitations}


While our approach provides meaningful insights into SQL generation mechanisms, several limitations remain. Interpretability results are sensitive to the choice of method and ablation strategy: zero-ablation offers robustness but inflates circuit size, while mean-ablation yields smaller circuits with unclear causal impact. Our EAP technique scales less effectively to larger models, and even the best-performing circuits recover only partial accuracy, suggesting that key mechanisms may remain undetected. We applied SAEs to CS4/5 to explore shared structure, but did not apply EAP due to its computational overhead and diminishing returns in deeper tasks, which require more targeted and scalable methods.

\section*{Ethical Considerations} This research primarily impacts the technical robustness of SQL understanding models. Our datasets should lead to more reliable database query systems, benefiting software developers and database administrators. As our work focuses on technical SQL syntax understanding using synthetic data, it presents minimal risks of societal harm or misuse. 

\bibliography{refs}

\appendix

\section{Note on AI assistance}
AI assistance was used for code development and improving the phrasing of the manuscript, while all analyses and conclusions were independently derived by the authors.

\section{Working Note on Over-training}
\label{app:OverTraining}
An LLM trained \textit{from scratch} on our CSn training data should predict SQL very accurately as it can assume the answer is always ``SELECT some fields FROM some table ORDER BY some fields WHERE some clauses''. To investigate whether we have over-trained the LLM to be SQL specific, we measure the model performance, before and after refinement training, on general tasks. In the TinyStories case, we use story telling tasks and evaluations as per the original paper, as shown in Table\ref{tab:CompletionRates} 

\begin{table}[h]
\centering
\small 
\begin{tabular}{|l|c|c|c|c|c|c|}
\hline
\multirow{2}{*}{ } & \multicolumn{2}{c|}{BM1} & \multicolumn{2}{c|}{BM2} & \multicolumn{2}{c|}{BM3} \\
\cline{2-7}
 & SQL & Story & SQL & Gen & SQL & Gen \\
\hline
Base & 2\% & 95\% & 24\% & 92\% & 24\% & 91\% \\
CS1 & 93\% & 92\% & 92\% & 87\% & 90\% & 88\% \\
CS2 & 83\% & 88\% & 81\% & 82\% & 89\% & 86\% \\
CS3 & 63\% & 85\% & 75\% & 79\% & 88\% & 84\% \\
\hline
\end{tabular}
\caption{The base (unrefined) models show limited ability to perform text-to-SQL tasks (columns "SQL"). For the command sets CS1, CS2 and CS3, the refined models, especially the LLMs, perform much better, while retaining most of their general capability (columns "Story" and  "Gen"). }
\label{tab:CompletionRates}
\end{table}

\section{Dataset details}

\subsection{Dataset Generation Methodology: CSx, CSx\_Syn}
\label{app:csx-method}
We developed an automated pipeline to generate SQL queries with increasing complexity while maintaining consistent patterns and structures for systematic progression and controlled experimentation.

Each dataset variant contains 100,000 examples, split into training (76.5\%), validation (13.5\%), and test (10\%) sets. While patterns are consistent across splits, we ensure no direct overlap of specific examples. 

Each example consists of three components, the natural language instruction, the table schema context, and the target SQL query and is generated as follows:

\textbf{Step 1: Schema Generation.} Each example features a single table with a randomly chosen name from a set of 200 common options (e.g., employees, products). Columns are drawn from a larger pool of 300 database fields (e.g., user\_id, timestamp), paired with appropriate SQL types (e.g., \texttt{INTEGER}, \texttt{VARCHAR}). Tables contain 2-12 columns, with certain fields restricted to relevant tables (e.g., salary only in employment-related tables) for semantic coherence.

\textbf{Step 2: Target Query Generation.} Based on command set level, we generate SQL queries with increasing complexity. CS1 creates basic \texttt{SELECT-FROM} statements, selecting random schema columns. CS2 adds \texttt{ORDER BY} clauses (90\% probability), randomly choosing ascending or descending order. CS3 introduces aggregate functions (\texttt{COUNT}, \texttt{SUM}, \texttt{AVG}, \texttt{MIN}, \texttt{MAX}), applied to numeric columns with an 85\% probability, ensuring compatibility with data types (e.g., \texttt{SUM} and \texttt{AVG} for numeric fields). CS4 adds \texttt{WHERE} clauses (80\% probability) with 1--3 type-consistent filters per query. CS5 introduces a second table with a \texttt{JOIN} clause on a shared field when schema compatibility allows.

\textbf{Step 3: Instruction Generation.} We translate SQL queries into natural language using a set of 50 templated patterns. These patterns range from direct commands (``Show me X from Y'') to questions (``What are the X in Y?'') and complex constructions (``From Y, get me X ordered by Z''). For base variants (CS1-3), we use exact table and column names. For synonym variants (CSx\_Syn), we replace 80\% of table names and 50\% of column names with synonyms. The templates maintain consistent structure while varying surface form. Aggregate functions have their own set of 20 dedicated phrase patterns (e.g., ``average of X'' → ``\texttt{AVG(X)}").

\textbf{Step 4: Quality Checks.} We implement an automated scoring system that evaluates generated SQL along multiple dimensions. It considers structural correctness (proper clause placement), semantic validity (field name matching), and implementation accuracy (correct aggregation and ordering). The system assigns partial credit based on component-wise correctness, allowing for fine-grained evaluation of model outputs. 

\subsection{Dataset Generation Methodology: CSx\_Nat}
\label{app:cs-nat-method}
We additionally explore a synthetic dataset generation method using Gretel AI’s Data Designer\footnote{\url{https://gretel.ai/navigator/data-designer}}. By defining two \emph{categorical seed columns} (\texttt{table\_name} and \texttt{instruction\_phrase}) and several \emph{generation columns} (e.g., \texttt{column\_names}, \texttt{selected\_columns}), we prompt Gretel’s language models to produce varied table names, column sets, and natural instructions. We then assemble them into structured SQL queries, applying post-processing rules for syntactic correctness. This approach further diversifies our text-to-SQL dataset by introducing additional schema and instruction variety. 

We use Gretel AI’s Data Designer\footnote{\url{https://gretel.ai/navigator/data-designer}}, which employs a compound AI pipeline controlled by a YAML specification, to create a synthetic dataset we call \textbf{CS1\_Nat}. Our configuration includes:

\begin{itemize}
    \item \textbf{Categorical seed columns:} We define two main seed columns, \texttt{table\_name} and \texttt{instruction\_phrase}, each populated with a diverse set of possible values (e.g., 100 table names, 60 instruction patterns). These serve as anchors, ensuring rich contextual variety in the generated rows.
    \item \textbf{Generated data columns:} We specify prompts for each generated column:
    \begin{itemize}
        \item \texttt{column\_names}: Produces domain-relevant, snake\_case field names.
        \item \texttt{selected\_columns}: Selects a subset of those field names for the final SQL query.
        \item \texttt{column\_data\_types}: Pairs each chosen field name with an appropriate SQL type (e.g., \texttt{INT}, \texttt{VARCHAR}).
        \item \texttt{sql\_prompt}: Reformulates table and column references into more natural instruction phrases, injecting synonym variety.
        \item \texttt{sql\_context}: Composes a \texttt{CREATE TABLE} statement by combining field names and data types into a cohesive schema definition.
        \item \texttt{sql}: Produces a full \texttt{SELECT} query matching the generated schema and context.
    \end{itemize}
    \item \textbf{Post-processors:} We apply validation checks to confirm syntactic correctness and logical coherence, ensuring the generated \texttt{SELECT} query aligns with the declared table schema and the instruction phrase.
\end{itemize}

By adjusting prompts and carefully selecting seed column values, we ensure each generated instruction and corresponding SQL query remains unique and contextually consistent. The resulting dataset is then partitioned into training, validation, and test splits for further use in text-to-SQL modeling and MI studies. All configuration details and additional examples are available upon request, enabling reproducibility and further exploration of prompt-based synthetic data generation.

Below is a simplified sample of the configuration file used to generate CS1-Nat using Data Designer:
\begin{verbatim}
model_suite: apache-2.0

special_system_instructions: >-
  You are a SQL expert. Your role is 
  to write SQL prompts, SELECT, 
  and CREATE TABLE statements.

categorical_seed_columns:
  - name: table_name
    values:
      - users
      - user_profiles
      - ...
  - name: instruction_phrase
    values:
      - ``Construct an SQL query to''
      - ``Retrieve the''
      - ...

generated_data_columns:
  - name: sql_prompt
    generation_prompt: >-
      Generate a concise natural 
      language instruction for ..
  - name: sql_context
    generation_prompt: >-
      Generate the SQL CREATE TABLE 
      statement for a table 
      named '{table_name}'...
  - name: sql
    generation_prompt: >-
      Generate a SQL statement ...

post_processors:
  - validator: code
\end{verbatim}

This systematic approach ensures reproducible dataset generation while maintaining controlled progression in task complexity. The explicit probabilities and controlled vocabularies enable consistent generation across different implementations.

\subsection{Dataset Examples}
\label{app:data-examples}

Table~\ref{tab:dataset-examples} shows representative examples from each category (CS1–CS5) in the TinySQL dataset. Each example includes a natural language question, a SQL statement and the corresponding SQL query.

\begin{table*}
\centering
\scriptsize  
\renewcommand{\arraystretch}{1.4}
\begin{tabular}{|c|p{3.5cm}|p{4.5cm}|p{4.5cm}|}
\hline
\textbf{Dataset} & \textbf{Create Statement} & \textbf{English Prompt} & \textbf{SQL Statement} \\
\hline

\hline
CS1 & \texttt{CREATE TABLE surveys ( file\_type VARCHAR(100), expires\_at DATETIME, comment TEXT, message TEXT, birthday DATE )} & Get a readout of expires\_at, birthday, comment, message and file\_type from surveys & \texttt{SELECT expires\_at, birthday, comment, message, file\_type FROM surveys} \\
\hline

CS1-Syn & \texttt{CREATE TABLE search\_filters ( downloads BIGINT, passed BOOLEAN, website VARCHAR(255), discount DECIMAL(10,2), birthday DATE, is\_active BOOLEAN, sequence SMALLINT, last\_message TEXT, meta\_keywords VARCHAR(500), response\_id INTEGER, reference TEXT )} & Can you get me recent message, date of birth, reply id, meta\_keywords, url, downloads and referral from refinement options? & \texttt{SELECT last\_message, birthday, response\_id, meta\_keywords, website, downloads, reference FROM search\_filters} \\
\hline

\hline
CS2 & \texttt{CREATE TABLE terms ( vote\_id INTEGER, start\_date DATE, mac\_address VARCHAR(17), last\_message TEXT, signature BLOB, middle\_name VARCHAR(100), coordinates POINT, address VARCHAR(255), meta\_description TEXT )} & Can you get me coordinates, signature, address, start\_date and mac\_address from terms? top coordinates, best meta\_description, with the highest start\_date, starting with the highest last\_message, starting with the highest mac\_address & \texttt{SELECT coordinates, signature, address, start\_date, mac\_address FROM terms ORDER BY coordinates DESC, meta\_description DESC, start\_date DESC, last\_message DESC, mac\_address DESC} \\
\hline

CS2-Syn & \texttt{CREATE TABLE configurations ( subject TEXT, image TEXT, response TEXT, major VARCHAR(100), company VARCHAR(255), session\_id VARCHAR(100), city VARCHAR(100), min\_score INTEGER, actual\_time INTEGER, manufacturer VARCHAR(255) )} & Looking in setup details, show me major, real duration, title, city, maker, session\_id, photo, minimum points and organization worst company & \texttt{SELECT major, actual\_time, subject, city, manufacturer, session\_id, image, min\_score, company FROM configurations ORDER BY company ASC} \\
\hline

CS3 & \texttt{CREATE TABLE conversions ( total\_price DECIMAL(10,2), approved\_at TIMESTAMP, content LONGTEXT, estimated\_time SMALLINT, permissions TEXT, first\_name VARCHAR(100), salt VARCHAR(32), full\_name VARCHAR(150) )} & Read out total count total\_price, how many full\_name, times approved\_at, tally salt, content, instances of permissions, first\_name and how many estimated\_time from conversions arranged by first\_name, oldest permissions, structured by estimated\_time & \texttt{SELECT COUNT(total\_price) AS COUNT\_total\_price, COUNT(full\_name) AS COUNT\_full\_name, COUNT(approved\_at) AS COUNT\_approved\_at, COUNT(salt) AS COUNT\_salt, content, COUNT(permissions) AS COUNT\_permissions, first\_name, COUNT(estimated\_time) AS COUNT\_estimated\_time FROM conversions ORDER BY first\_name ASC, permissions ASC, estimated\_time ASC} \\
\hline

CS3-Syn & \texttt{CREATE TABLE product\_versions ( skills TEXT, location POINT )} & Can you get me latest skills and location from releases? sorted alphabetically by location & \texttt{SELECT MAX(skills) AS MAX\_skills, location FROM product\_versions ORDER BY location ASC} \\
\hline

CS4 & \texttt{CREATE TABLE data ( max\_score INTEGER, gender CHAR(1), region GEOMETRY, location GEOMETRY, middle\_name VARCHAR(100), salt VARCHAR(32), last\_message TEXT )} & Bring up middle\_name, last\_message, location, salt, region, max\_score and gender from information where salt is containing '\%z\%' priority ordered by max\_score, in reverse alphabetical order of location & \texttt{SELECT middle\_name, last\_message, location, salt, region, max\_score, gender FROM data WHERE salt LIKE '\%z\%' ORDER BY max\_score DESC, location DESC} \\
\hline

CS5 & \texttt{CREATE TABLE workflow\_templates ( like\_id INTEGER, heading DECIMAL(5,2), is\_verified BOOLEAN, share\_id BIGINT, average\_score FLOAT, rating DECIMAL(3,2), time TIME, certification VARCHAR(255), name VARCHAR(100), city VARCHAR(100) )} & I need a list of verified status, like\_id, time, average\_score, designation, city, heading, rating, certification and share\_id from procedure patterns join with pairs on like\_id equals user\_id ordered numerically by average\_score, ordered by date of is\_verified, alphabetically by certification, grouped by rating & \texttt{SELECT is\_verified, like\_id, time, average\_score, name, city, heading, rating, certification, share\_id FROM workflow\_templates JOIN pairs ON workflow\_templates.like\_id = pairs.user\_id ORDER BY average\_score ASC, is\_verified ASC, certification ASC, rating ASC} \\
\hline

\end{tabular}
\caption{Examples from TinySQL dataset across five compositional subsets (CS1–CS5) and CS1-Syn, CS2-Syn, CS3-Syn data.}
\label{tab:dataset-examples}
\end{table*}

\section{Trained Models}
\label{app:TrainedModels}

\paragraph{Base Model Licenses.} The base model for BM1, \texttt{TinyStories-33M} \citep{eldan2023tinystoriessmalllanguagemodels},is released under the MIT license. The base model for BM2, \texttt{Qwen2.5-0.5B-Instruct} \citep{qwen2.5}, is released under the Apache 2.0 license. The base model for BM3, \texttt{Llama-3.2-1B-Instruct} \citep{grattafiori2024llama3herdmodels}, is released under the Llama 3.2 Community License.

\paragraph{Model Architectures.}
The three models vary in size and architecture. BM1 (\texttt{TinyStories-33M}) has 2 transformer layers and 16 attention heads. BM2 (\texttt{Qwen2.5-0.5B-Instruct}) is composed of 24 layers with 14 attention heads. BM3 (\texttt{Llama-3.2-1B-Instruct}) consists of 16 layers and 32 attention heads. These configurations reflect each model's respective scale and capacity.

\paragraph{Training Details.}
We use the \texttt{transformers} library and the \texttt{trl} extension for instruction fine-tuning using the standard alpaca template. Models are trained using an \emph{AdamW} optimizer with weight decay of 0.01. The learning rate used is $1\times10^{-5}$, a value chosen based on preliminary experiments showing that higher rates led to larger validation oscillations and less stable convergence. We employ 100 warmup steps at the beginning of training. Each model is trained for a single epoch over \texttt{76,500} training examples, with \texttt{13,500} validation and \texttt{10,000} test examples in each dataset split.
We use a maximum sequence length of 512 tokens. The effective batch size is 32 per update step, achieved by running on four GPUs with a per-device batch size of 8 and a gradient accumulation step of 1. For BM2 and BM3, we enable \texttt{flash-attention v2} for efficient attention computation and reduced memory overhead. The padding side is configured based on each model’s tokenizer requirements, and new tokens (e.g.\ a dedicated \texttt{<pad>} token) are added or resized if needed to accommodate smaller or older base models (e.g.\ TinyStories). Training took about 1 hour / epoch for TinyStories and about 2 hours / epoch on Llama / Qwen models on 2 x A100 GPUs.

We trained basic and semantic models as shown in Table \ref{tab:TrainedModels}.

\begin{table*}[h!]
\centering
\begin{tabular}{|l|l|l|l|c|}
\hline
\textbf{Abbrev.} & \textbf{Model} & \textbf{CSn} & \textbf{Type} & \textbf{Exact-match Accuracy} \\
\hline
BM1\_CS0 & TinyStories & N/A & N/A & 0\%** \\
BM2\_CS0 & Qwen 2.5     & N/A & N/A & 10\%** \\
BM3\_CS0 & Llama 3.2    & N/A & N/A & 10\%** \\
\hline
BM1\_CS1\_1.8 & TinyStories & CS1 & Basic & 98.79\% \\
BM1\_CS1\_1.10 & TinyStories & CS1\_Syn & Semantic & 92.97\% \\
BM1\_CS2\_2.8 & TinyStories & CS2 & Basic & 97.62\% \\
BM1\_CS2\_2.10 & TinyStories & CS2\_Syn & Semantic & 92.02\% \\
BM1\_CS3\_3.8 & TinyStories & CS3 & Basic & 94.31\% \\
BM1\_CS3\_3.10 & TinyStories & CS3\_Syn & Semantic & 85.63\% \\
\hline
BM2\_CS1\_4.2 & Qwen 2.5 & CS1 & Basic & 100.0\% \\
BM2\_CS1\_4.3 & Qwen 2.5 & CS1\_Syn & Semantic & 99.52\% \\
BM2\_CS2\_5.2 & Qwen 2.5 & CS2 & Basic & 100.0\% \\
BM2\_CS2\_5.3 & Qwen 2.5 & CS2\_Syn & Semantic & 99.55\% \\
BM2\_CS3\_6.2 & Qwen 2.5 & CS3 & Basic & 99.82\% \\
BM2\_CS3\_6.3 & Qwen 2.5 & CS3\_Syn & Semantic & 98.88\% \\
\hline
BM3\_CS1\_7.2 & Llama 3.2 & CS1 & Basic & 100.0\% \\
BM3\_CS1\_7.3 & Llama 3.2 & CS1\_Syn & Semantic & 99.67\% \\
BM3\_CS2\_8.2 & Llama 3.2 & CS2 & Basic & 100.0\% \\
BM3\_CS2\_8.3 & Llama 3.2 & CS2\_Syn & Semantic & 99.63\% \\
BM3\_CS3\_9.2 & Llama 3.2 & CS3 & Basic & 99.94\% \\
BM3\_CS3\_9.3 & Llama 3.2 & CS3\_Syn & Semantic & 99.43\% \\
\hline
\end{tabular}
\caption{The base models (TinyStories-Instruct-2Layers-33M, Qwen2.5-0.5B-Instruct and Llama-3.2-1B-Instruct show limited ability to perform text to SQL tasks. The trained models, especially the LLMs, perform much better, while retaining most of their general capability. 
** The evaluation is strict as it requires the model provide just the answer without any additional preface or text, explaining the low zero-shot scores here for Llama and Qwen}
\label{tab:TrainedModels}
\end{table*}

\begin{figure}[h!]
\centering
\includegraphics[width=0.95\columnwidth]{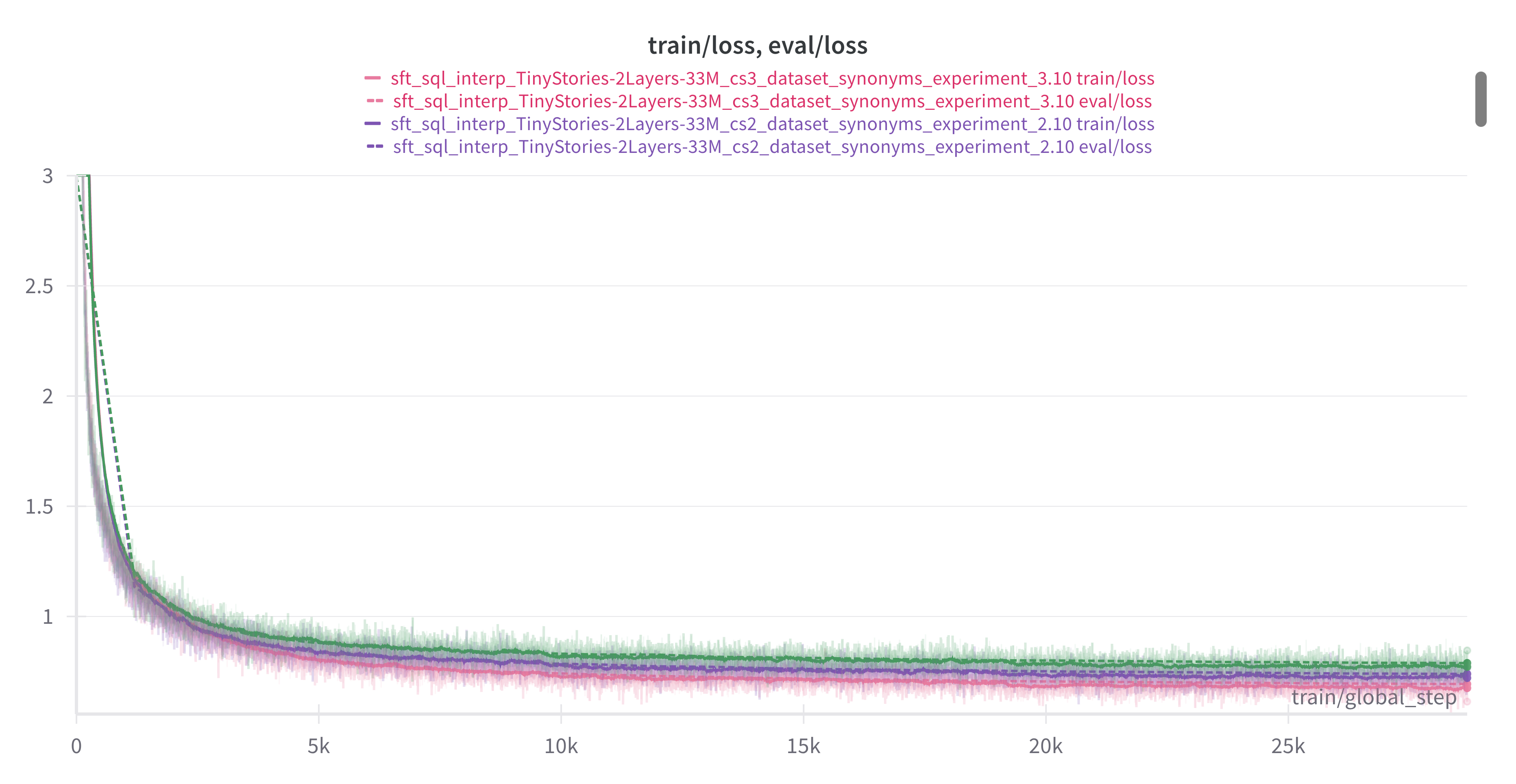}
\caption{Training and Validation loss curves for instruction tuning BM1 (TinyStories-Instruct-2Layers-33M) on CS1\_Syn, CS2\_Syn, CS3\_Syn}
\label{fig:bm1_syn_loss}
\end{figure}

\begin{figure}[h!]
\centering
\includegraphics[width=0.95\columnwidth]{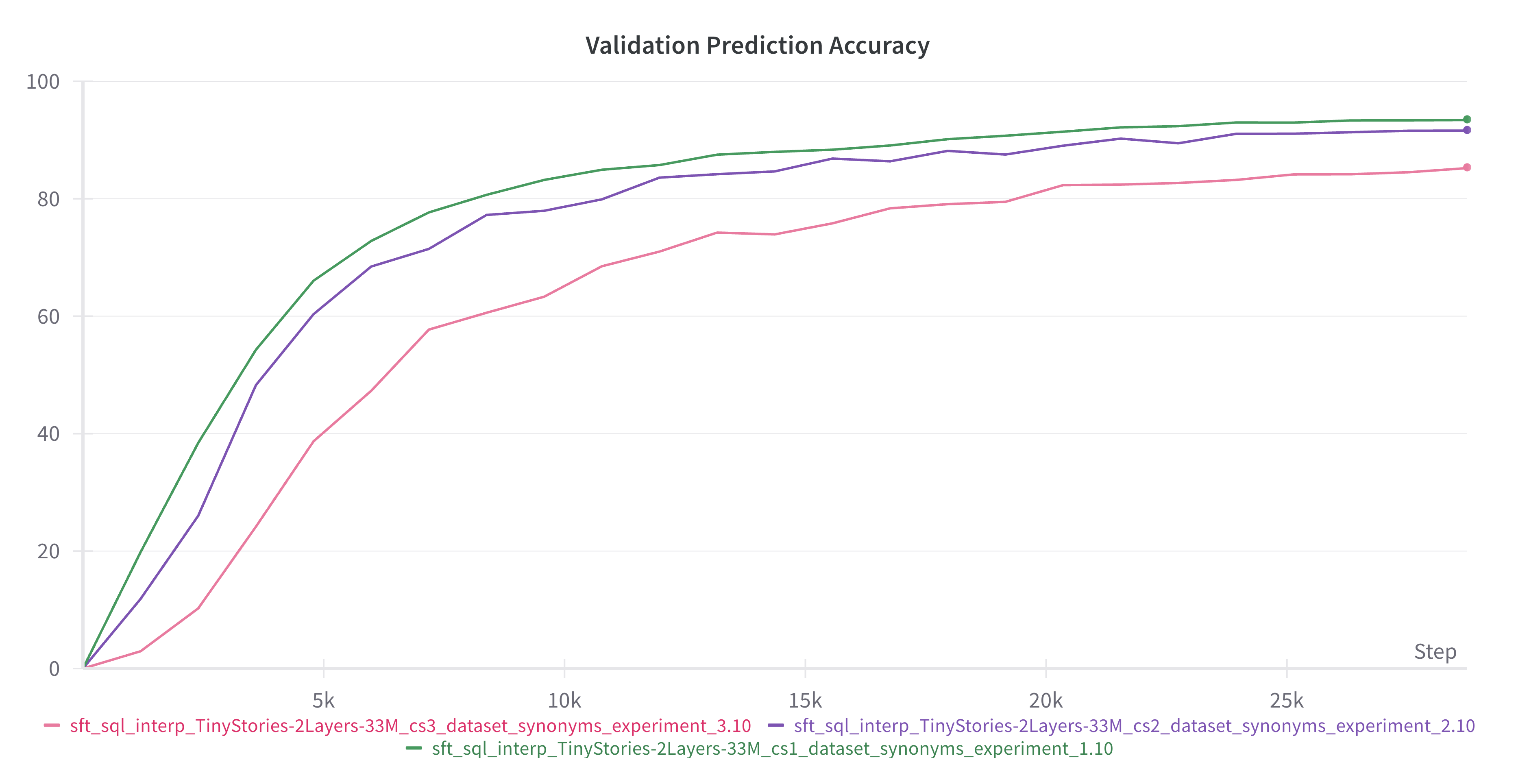}
\caption{Accuracy curves for instruction tuning BM1 (TinyStories-Instruct-2Layers-33M) on CS1\_Syn, CS2\_Syn, CS3\_Syn}
\label{fig:bm1_syn_acc}
\end{figure}

\begin{figure}[h!]
\centering
\includegraphics[width=0.95\columnwidth]{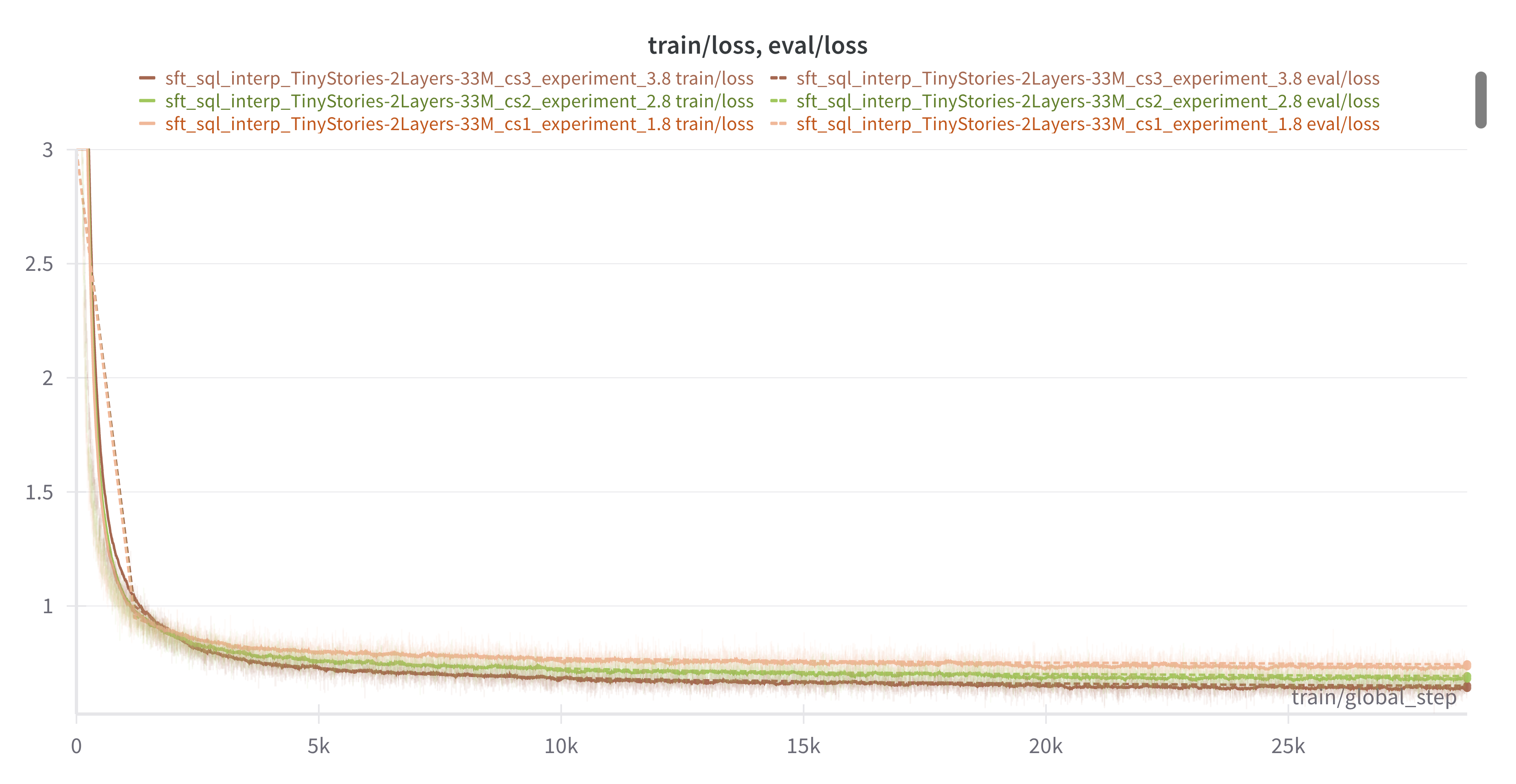}
\caption{Training and Validation loss curves for instruction tuning BM1 (TinyStories-Instruct-2Layers-33M) on CS1, CS2, CS3}
\label{fig:bm1_csx_loss}
\end{figure}

\begin{figure}[h!]
\centering
\includegraphics[width=0.95\columnwidth]{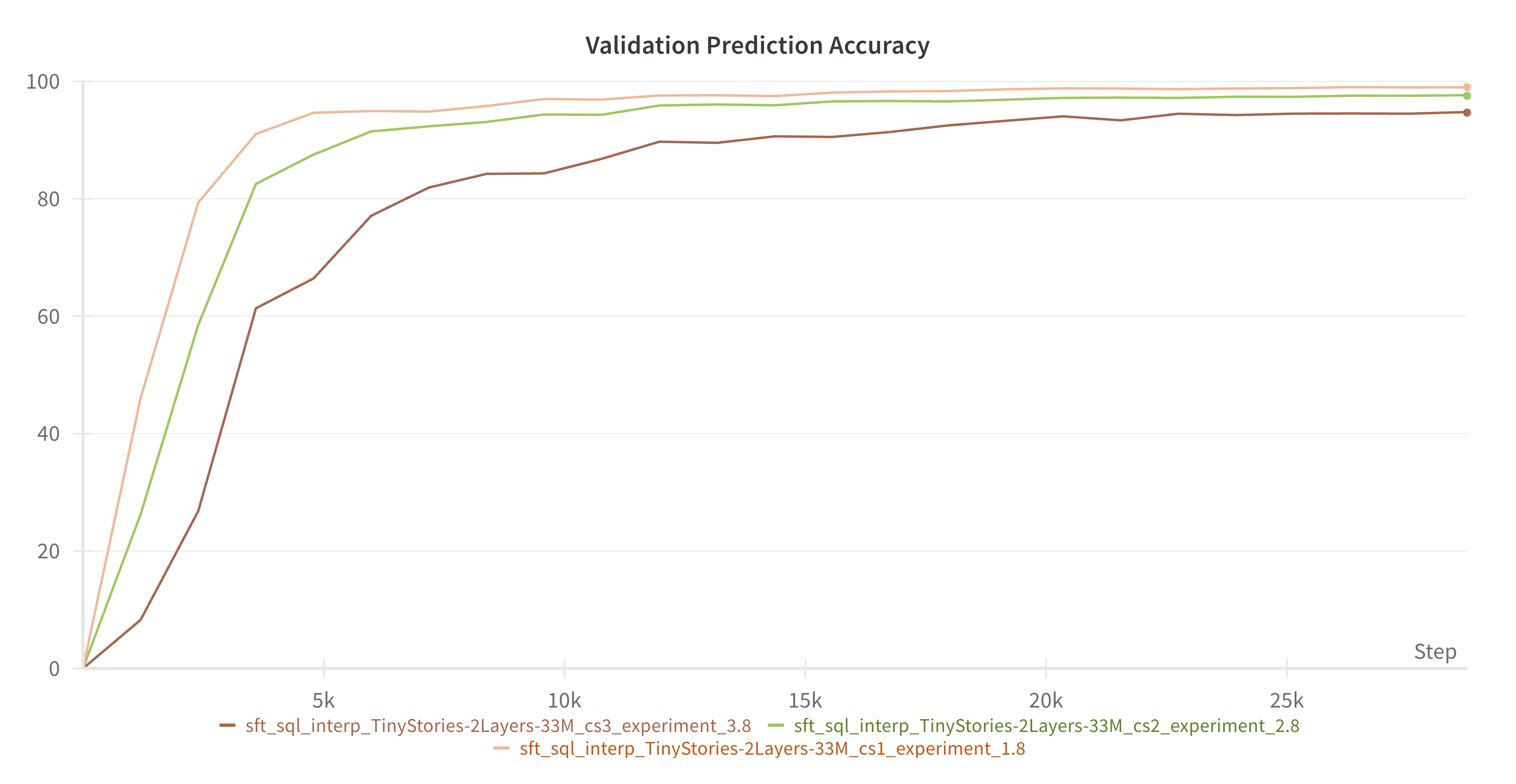}
\caption{Accuracy curves for instruction tuning BM1 (TinyStories-Instruct-2Layers-33M) on CS1, CS2, CS3}
\label{fig:bm1_syn_csx_acc}
\end{figure}

\label{sec:data_characteristics}
\section{Data set entry characteristics}
For brevity, the main text of this paper uses short examples. The dataset entries are considerably longer. These are some characteristics of the generated datasets:

\begin{itemize}
\item  All CREATE TABLE clause contains (randomly) 2 to 12 columns.
\item  All SELECT clause contains (randomly) 1 to the number of table columns.
\item  In data set CS2, the ORDER BY clause is added and contains (randomly) 1 to the number of table columns. The ASC or DESC ordering is chosen randomly. An ORDER BY clause is added to 90\% of entries.
\item  In data set CS3, each SELECT column is randomly assigned a valid (depending on the column data type) aggregate from the list: SUM, AVG, MIN, MAX, COUNT and “”. That is, \~20\% of columns have no aggregate.
\item In data set CS4, we extend CS3 by adding a type-aware WHERE clause to 80\% of queries, selecting 1–3 distinct columns and type appropriate operators/values.
\item In data set CS5, we introduce a second table and a JOIN clause that matches columns with the same base type. If there are no columns in the second table that share a common type with the table already in the statement, we do not add a JOIN clause. This results in 73.0\% of the train split for CS5 containing a JOIN.
\end{itemize}

Table \ref{tab:LongDataSetExamples} shows the maximum length of entries in each dataset.

\begin{table*}
    \centering
    \begin{tabular}{l|l|l|l}
        \toprule
        \textbf{Dataset} & \textbf{English Prompt} & \textbf{Create statement} & \textbf{SQL statement} \\
        \midrule
        CS1 & 19 words, 169 chars & 29 words, 299 chars & 15 words, 159 chars \\
        CS2 & 63 words, 476 chars & 29 words, 308 chars & 40 words, 325 chars \\
        CS3 & 78 words, 579 chars & 29 words, 296 chars & 55 words, 564 chars \\
        \bottomrule
    \end{tabular}
    \caption{Lengths of the longest entries in words and characters per dataset. For the synonym datasets, the English Prompts are \~ 10\% longer.}
    \label{tab:LongDataSetExamples}
\end{table*}

\section{Activation Patching Results}
\label{app:ap-graphs}
\begin{figure*}[htbp]
    \centering
    \begin{tabular}{cc}
        \includegraphics[width=0.45\textwidth]{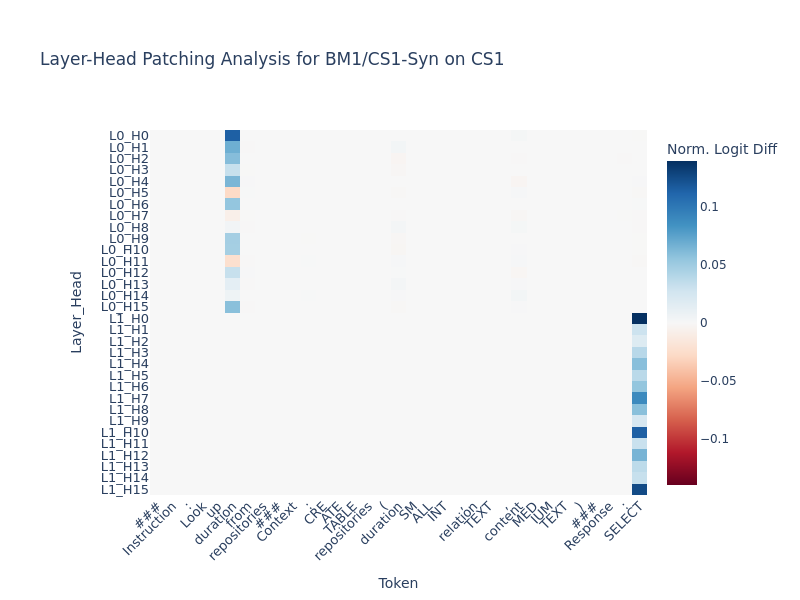} & \includegraphics[width=0.45\textwidth]{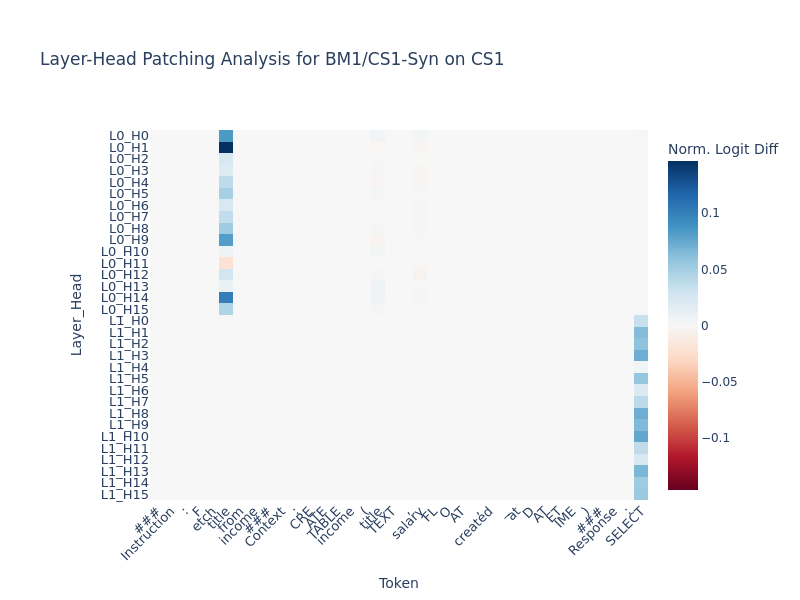} \\
        \includegraphics[width=0.45\textwidth]{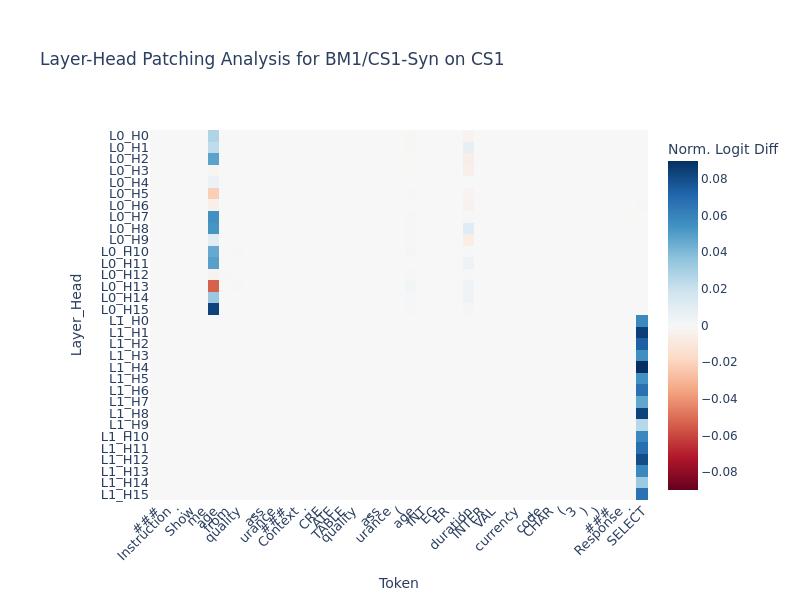} & \includegraphics[width=0.45\textwidth]{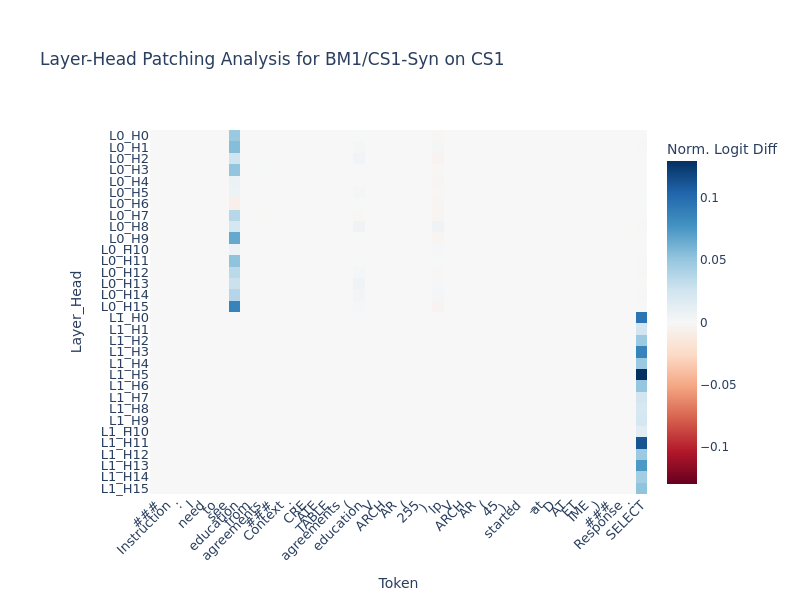} \\
    \end{tabular}
    \caption{Activation patching results for 4 prompts, where the table contains three fields and the user queries for only the first field of the table. Even when controlling for some variables, the results differ from prompt to prompt.}
    \label{fig:ap-graphs}
\end{figure*}

Across prompts across all classes, we find that the activation patching results differ quite significantly from each other, even for prompts in the same class. We provide a sample of 4 examples in Figure \ref{fig:ap-graphs}, and we note that this variation persists throughout all examples.

\section{EAP Circuits Identification}
\label{app:circuitidentification}

\subsection{Edge Attribution Patching (EAP) Results}

We run the Edge Attribution Patching (EAP) experiment as described in Sec.~\ref{sec:minimal_circuits} on a set of clean and corrupted prompt pairs. For each model, we determine which features to corrupt based on the complexity of the command set being evaluated. Typically, we corrupt one token at a time, then input the clean and corrupted prompt pairs into the EAP algorithm to identify the set of important edges for each feature.

To ensure reliability, we run EAP on 15 batches of 100 prompts each, recording only the edges that appear across all batches above a chosen threshold.

\subsubsection{Features used per dataset}
\label{app:eap-features}

For the CS1 command set, where SQL statements follow this structure:

\begin{tcolorbox}[colframe=lightgray,title=Example from CS1]
\#\#\# Instruction: show me the type and date from the orders table \#\#\# Context: CREATE TABLE orders ( type CHAR, date INT ) \#\#\# Response: SELECT type, date FROM orders
\end{tcolorbox}

We identify the following key features for correct response prediction:

\begin{itemize}
    \item \textbf{EngTableName}: Table name in the \#\#\# Instruction
    \item \textbf{EngFieldName}: Field name in the \#\#\# Instruction
    \item \textbf{DefTableName}: Table name in the \#\#\# Context
    \item \textbf{DefFieldName}: Field name in the \#\#\# Context
\end{itemize}

For each feature, we corrupt prompts by replacing only the feature with a random word while keeping everything else unchanged. We then run EAP and record the results.

For the CS2 command set, where SQL statements follow this structure:

\begin{tcolorbox}[colframe=lightgray,title=Example from CS2]
\#\#\# Instruction: show me the category and value from the links table ordered by value in ascending order \#\#\# Context: CREATE TABLE links ( category CHAR, value TEXT ) \#\#\# Response: SELECT category, value FROM links ORDER BY value ASC
\end{tcolorbox}

We use the same features as in CS1 and add:

\begin{itemize}
    \item \textbf{OrderByField}: Field used for ordering in the \#\#\# Instruction (\textit{value} in the example)
    \item \textbf{OrderByDirection}: Ordering direction in the \#\#\# Instruction (\textit{ascending} in the example)
\end{itemize}

For the CS3 command set, where SQL statements follow this structure:

\begin{tcolorbox}[colframe=lightgray,title=Example from CS3]
\#\#\# Instruction: From orders, get me least recent code with the highest code \#\#\# Context: CREATE TABLE orders ( weight CHAR, code TEXT ) \#\#\# Response: SELECT MIN(code) AS MIN\_code FROM orders ORDER BY code DESC
\end{tcolorbox}

In addition to the CS2 features, we introduce:
\begin{itemize}
    \item \textbf{AggregateField}: Field used for aggregation in the \#\#\# instruction (code in the example)
    \item \textbf{AggregateFunction}: Term indicating the aggregation function in the \#\#\# Instruction (highest in the example)
\end{itemize}

\subsubsection{EAP results}

Once we generate our set of batches for each feature, we run the EAP algorithm and record the most important edges.

For each run, we select the top 10 edges per batch and retain only the edges that appeared at least T\% of the time across batches. For CS1 and CS2, we set T = 100, while for CS3, we set T = 80.

For each feature, we run EAP four times, covering all possible truth-value combinations of use\_synonyms\_field and use\_synonyms\_table. This allows us to identify the set of edges responsible for each feature in both semantic and non-semantic cases.

\begin{figure}
    \centering
    \begin{minipage}{0.24\textwidth}
        \centering
        \includegraphics[width=\linewidth]{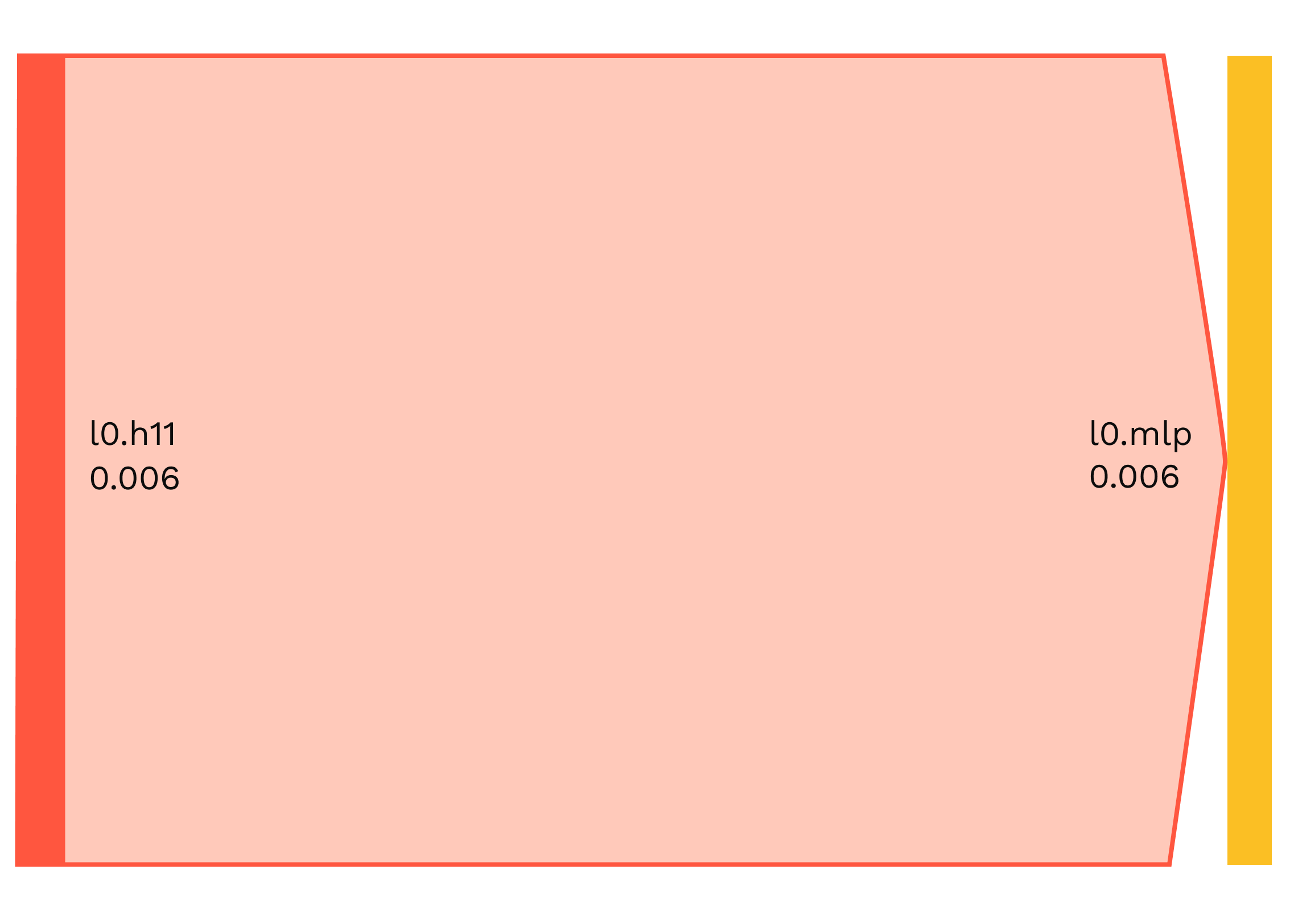}
        \label{fig:eap_fig1}
    \end{minipage}
    \hfill
    \begin{minipage}{0.24\textwidth}
        \centering
        \includegraphics[width=\linewidth]{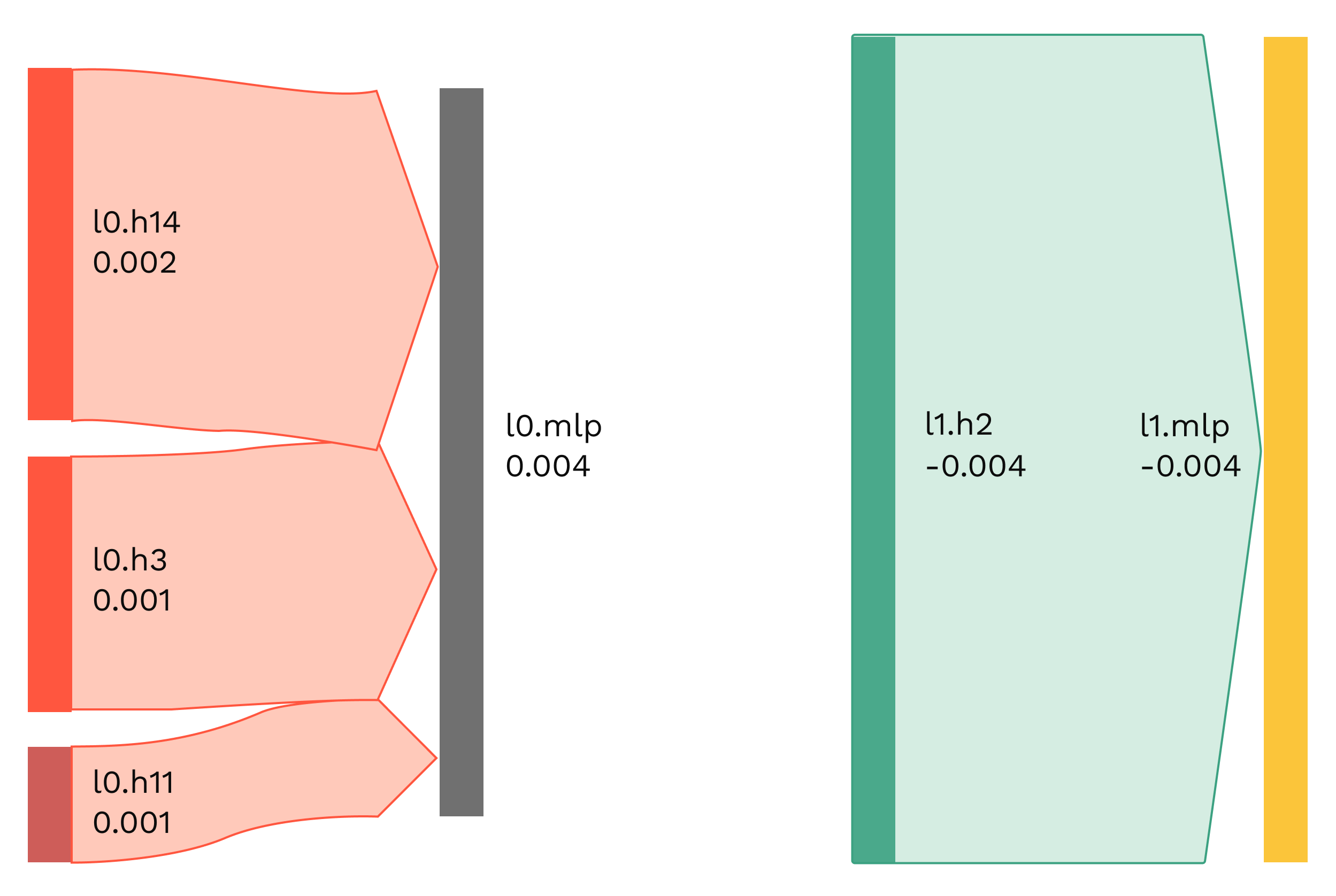}
        \label{fig:eap_fig2}
    \end{minipage}
    \hfill
    \begin{minipage}{0.24\textwidth}
        \centering
        \includegraphics[width=\linewidth]{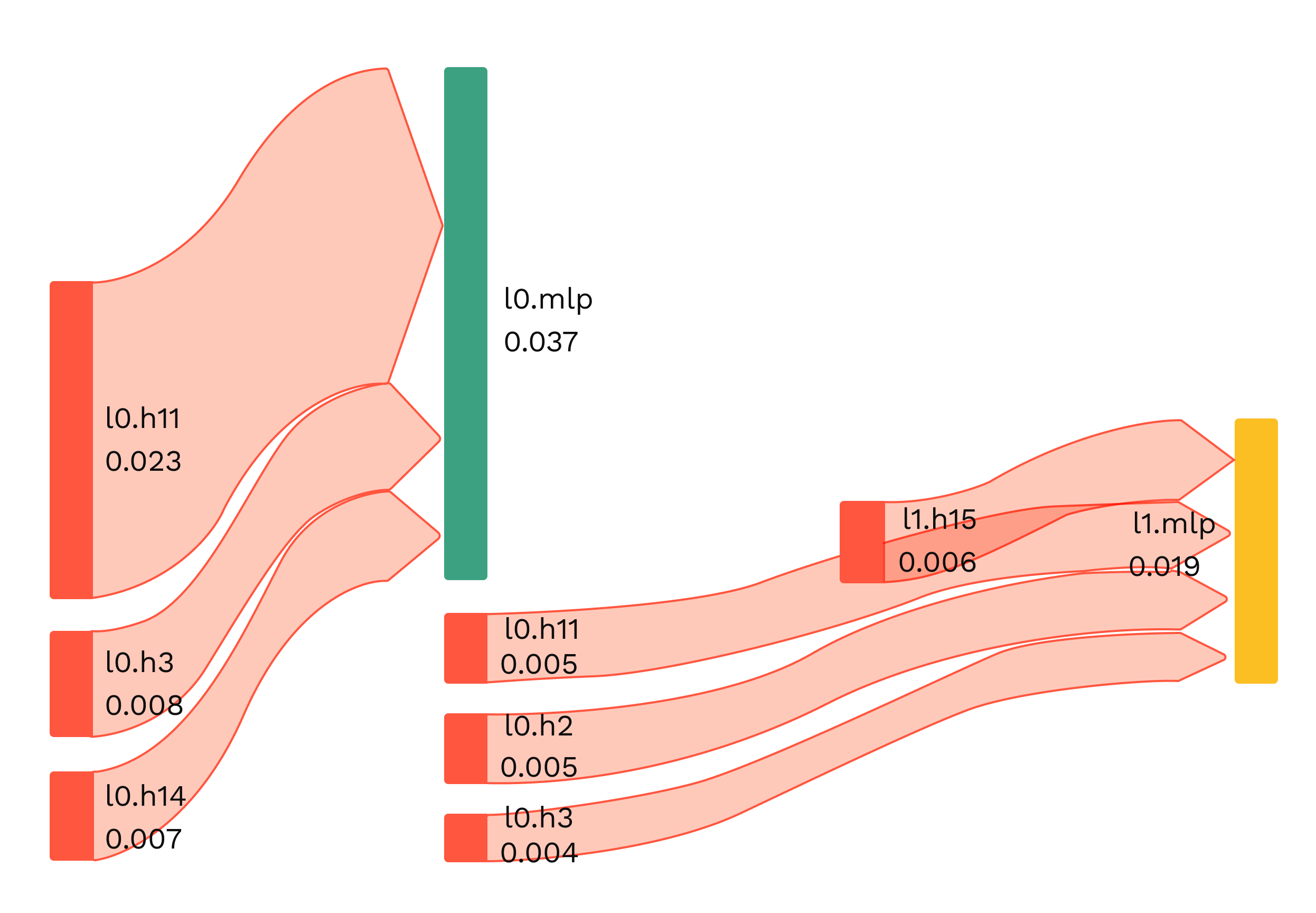}
        \label{fig:eap_fig3}
    \end{minipage}
    \hfill
    \begin{minipage}{0.24\textwidth}
        \centering
        \includegraphics[width=\linewidth]{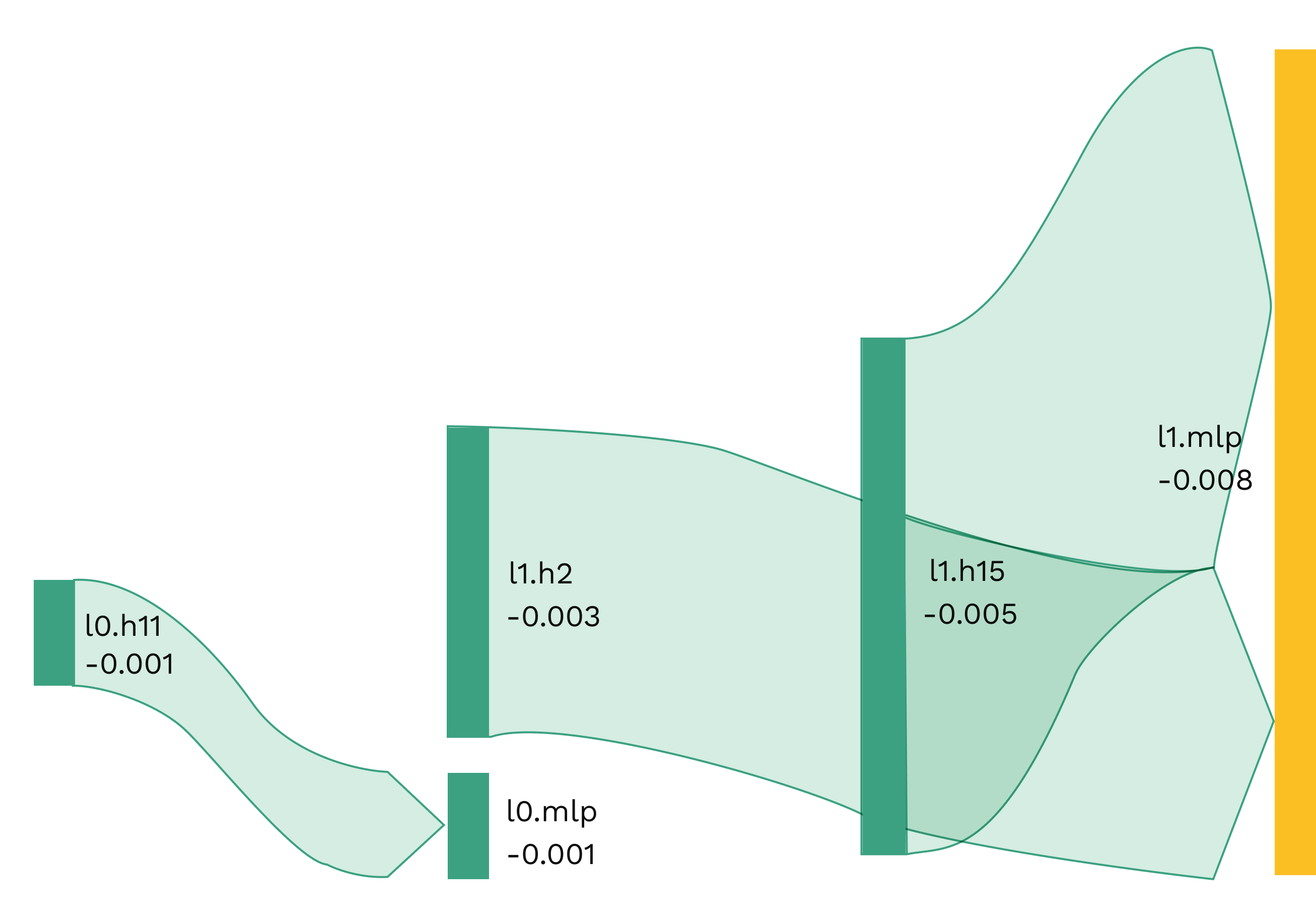}
        \label{fig:eap_fig4}
    \end{minipage}
    
    \vspace{0.5cm} 
    
    \begin{minipage}{0.24\textwidth}
        \centering
        \includegraphics[width=\linewidth]{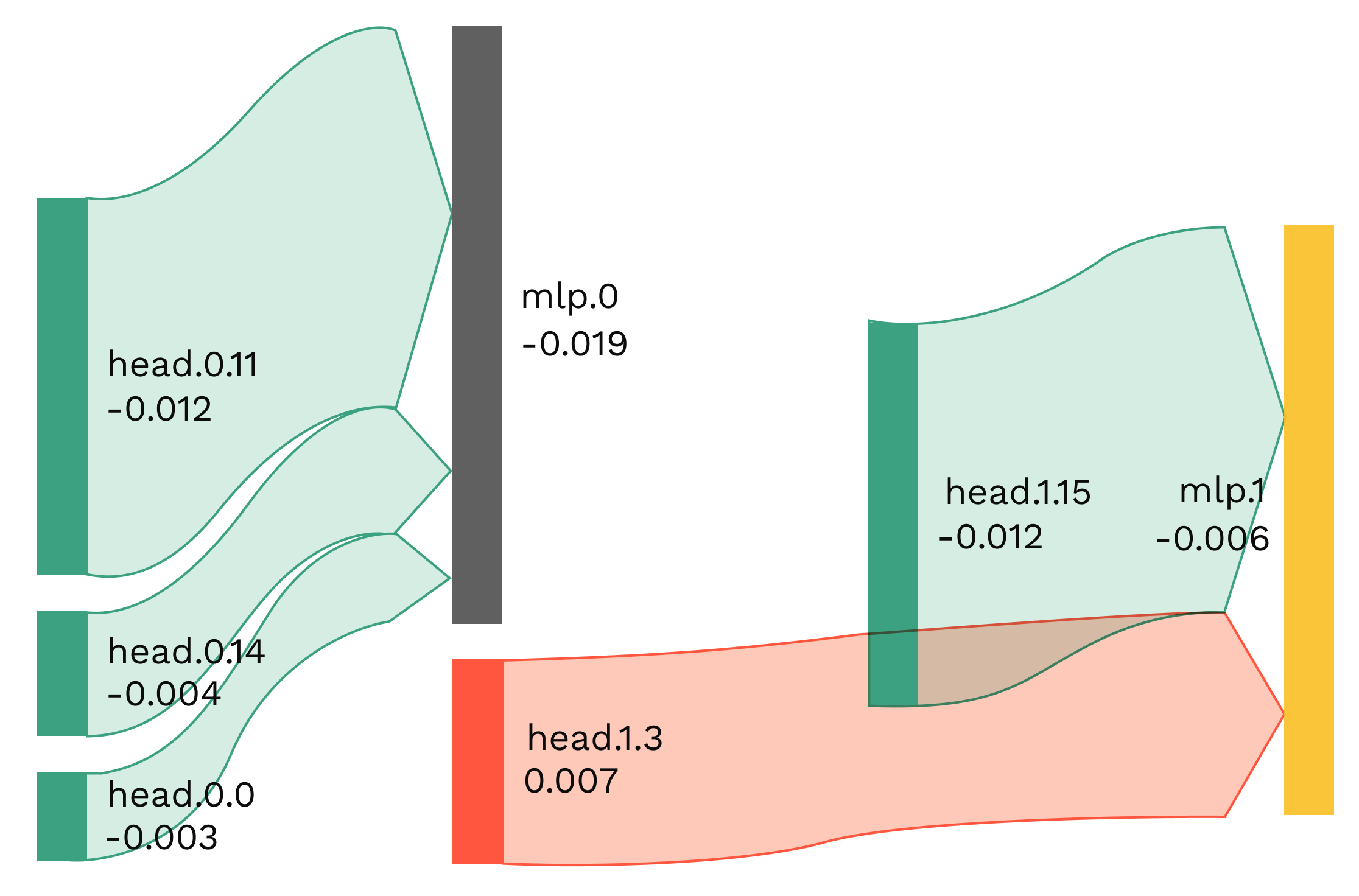}
        \label{fig:eap_fig5}
    \end{minipage}
    \hfill
    \begin{minipage}{0.24\textwidth}
        \centering
        \includegraphics[width=\linewidth]{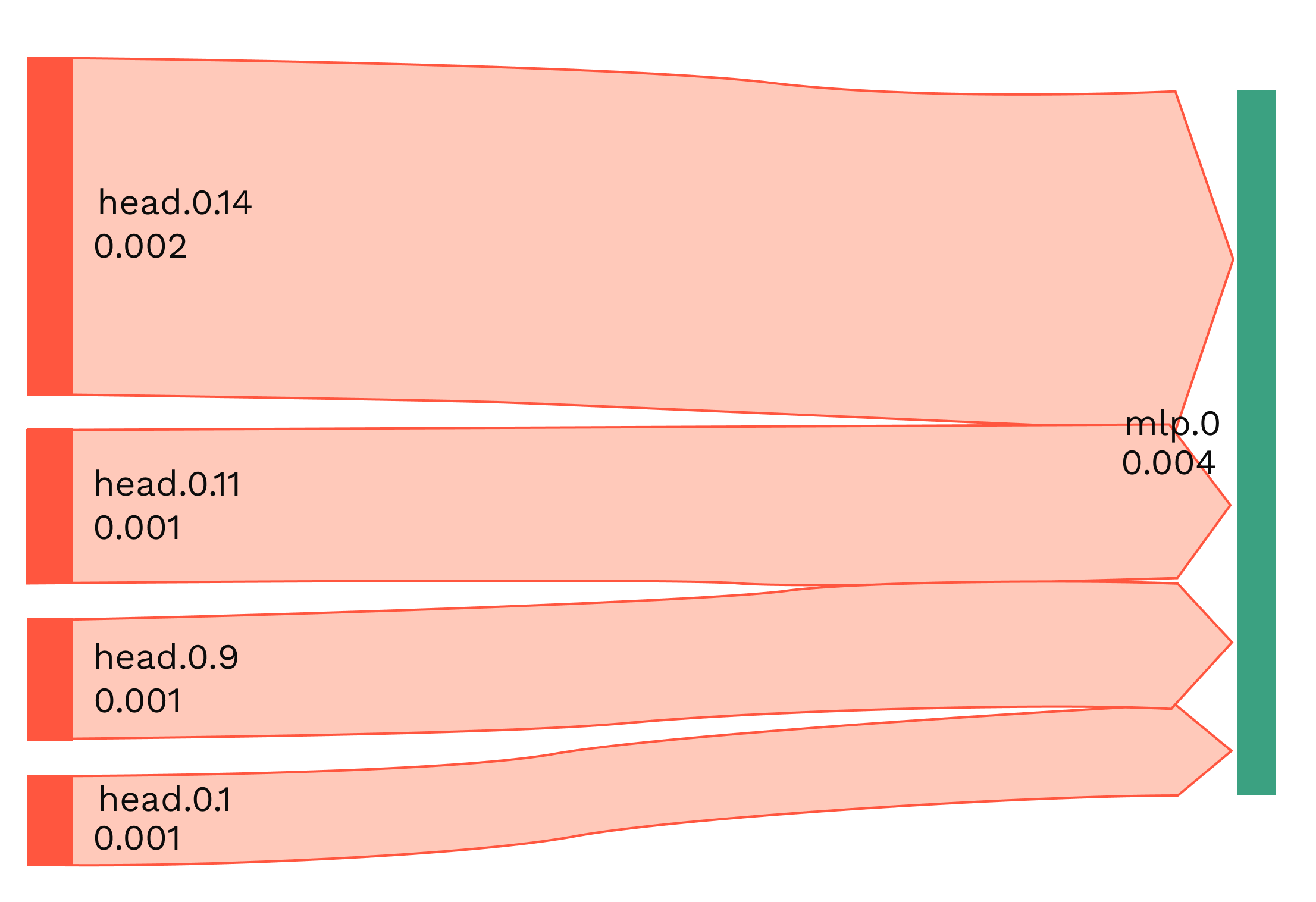}
        \label{fig:eap_fig6}
    \end{minipage}
    \hfill
    \begin{minipage}{0.24\textwidth}
        \centering
        \includegraphics[width=\linewidth]{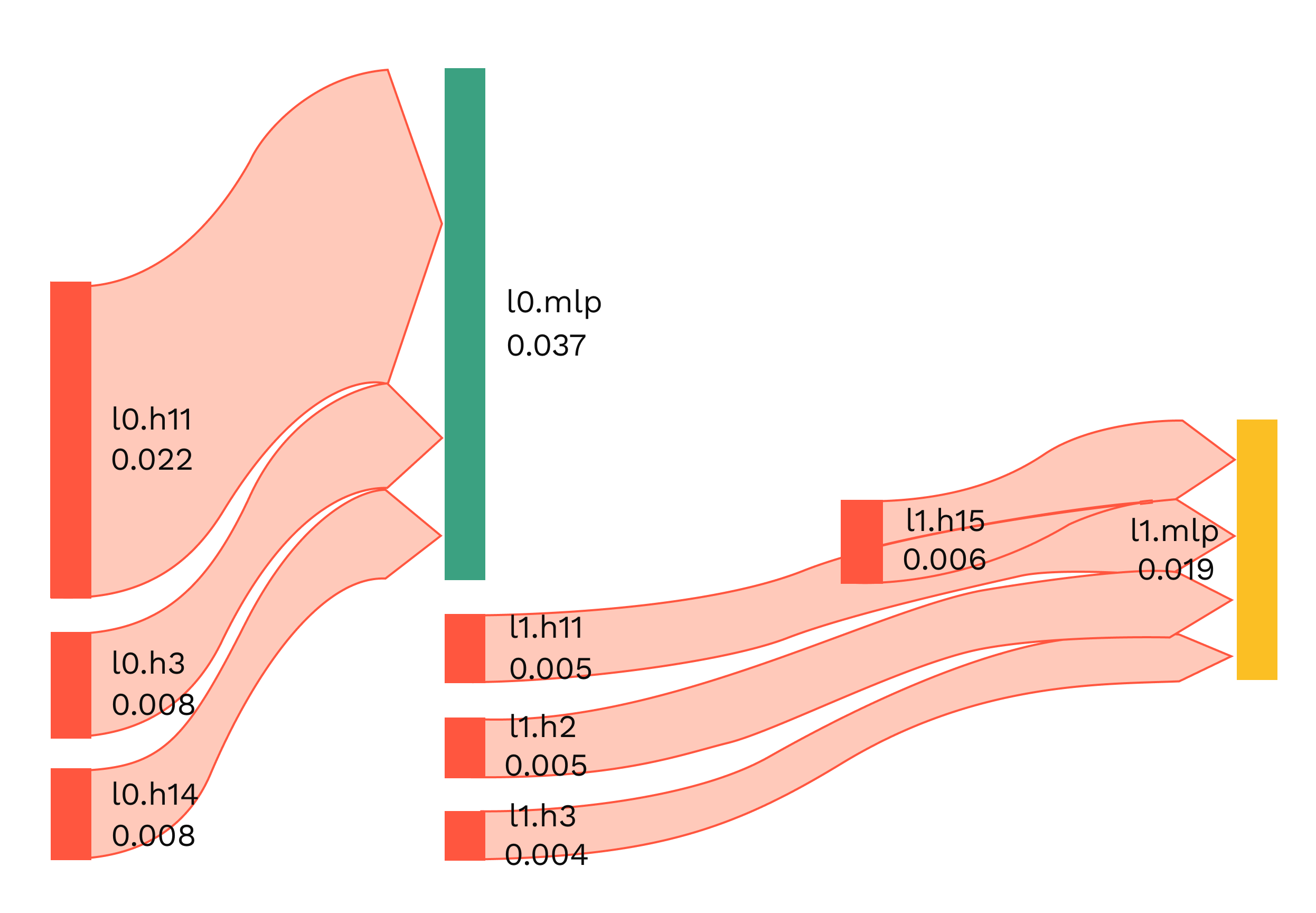}
        \label{fig:eap_fig7}
    \end{minipage}
    \hfill
    \begin{minipage}{0.24\textwidth}
        \centering
        \includegraphics[width=\linewidth]{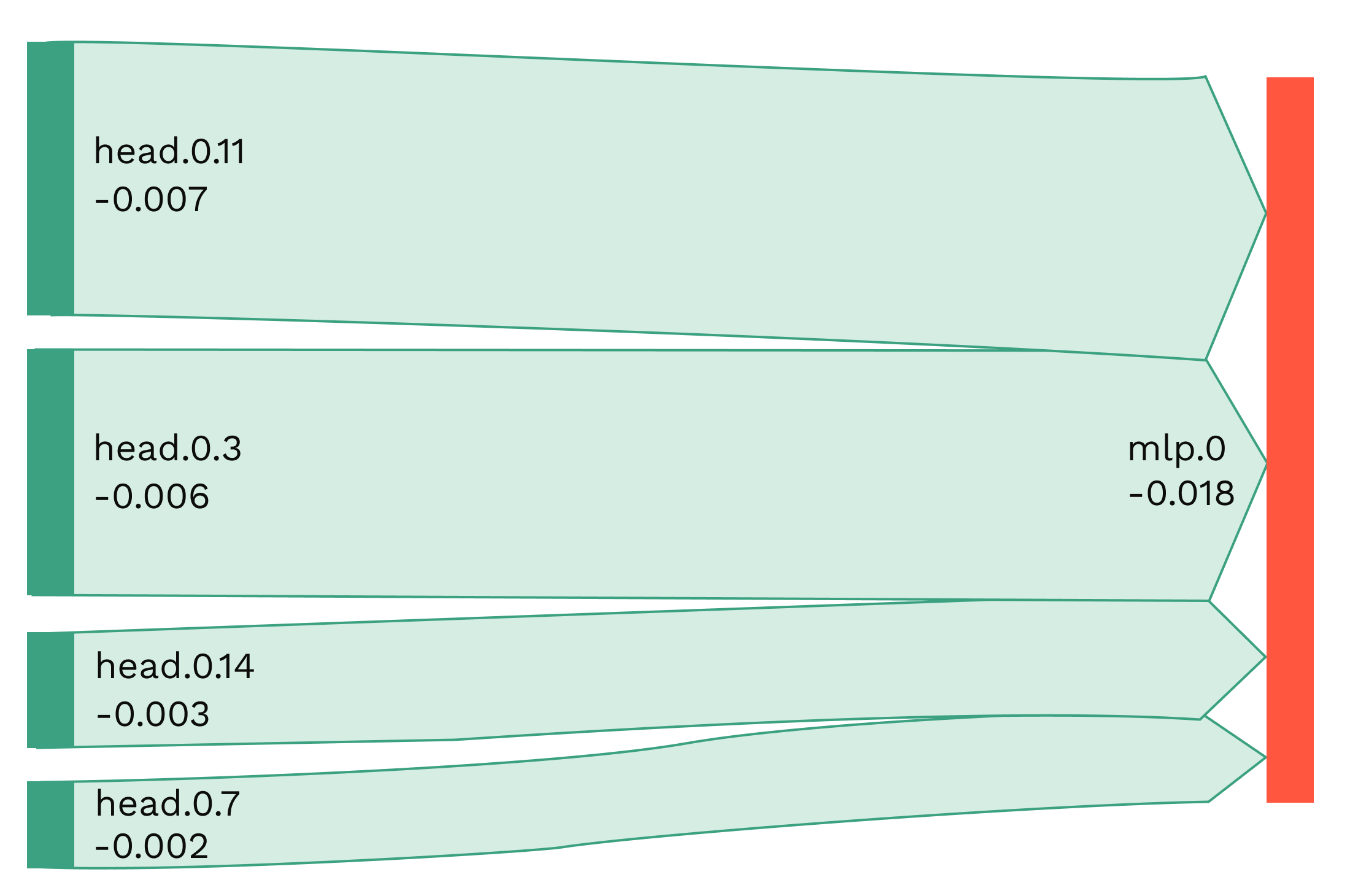}
        \label{fig:eap_fig8}
    \end{minipage}
    
    \vspace{0.5cm} 
    
    \begin{minipage}{0.24\textwidth}
        \centering
        \includegraphics[width=\linewidth]{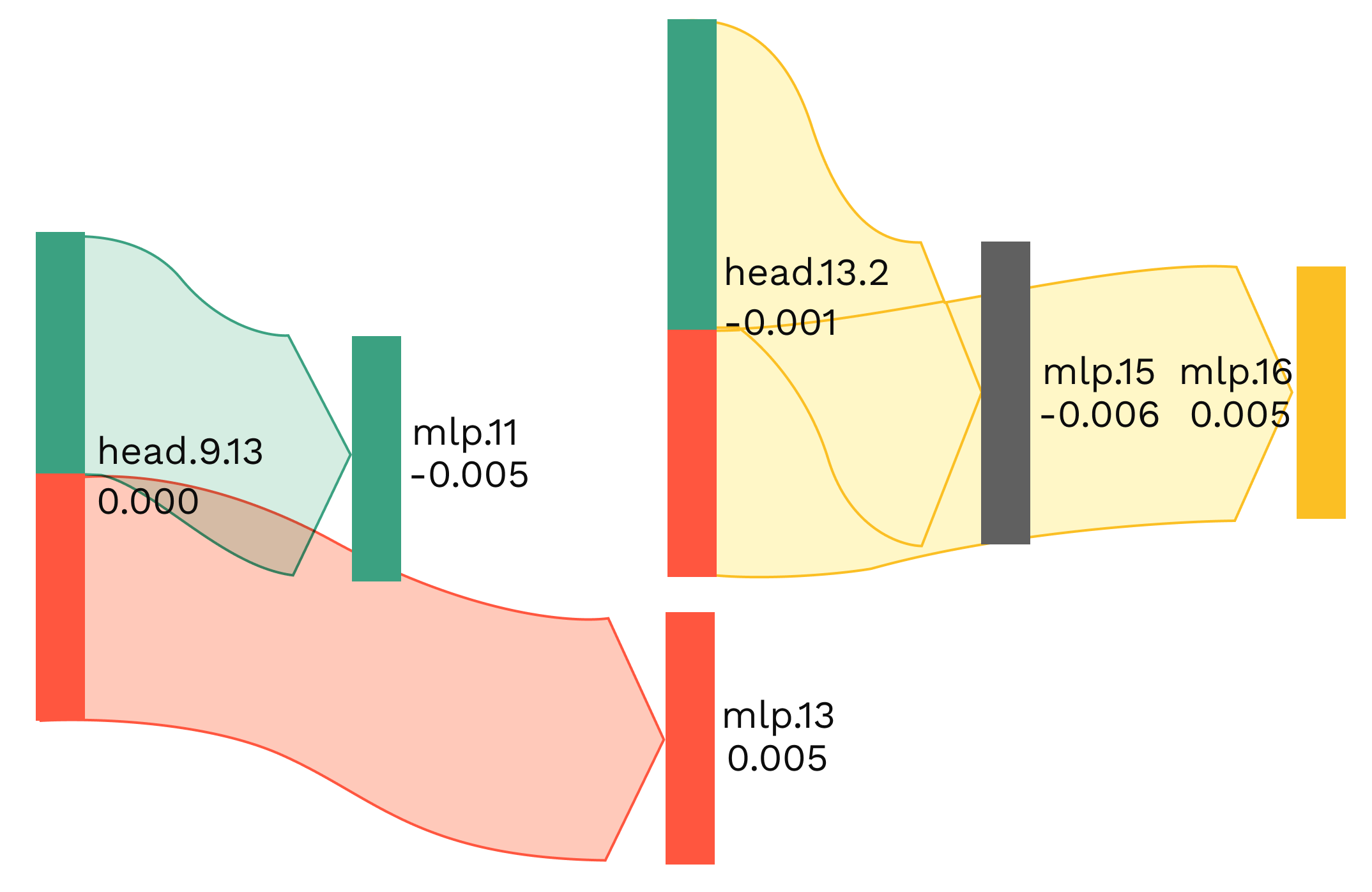}
        \label{fig:eap_fig9}
    \end{minipage}
    \hfill
    \begin{minipage}{0.24\textwidth}
        \centering
        \includegraphics[width=\linewidth]{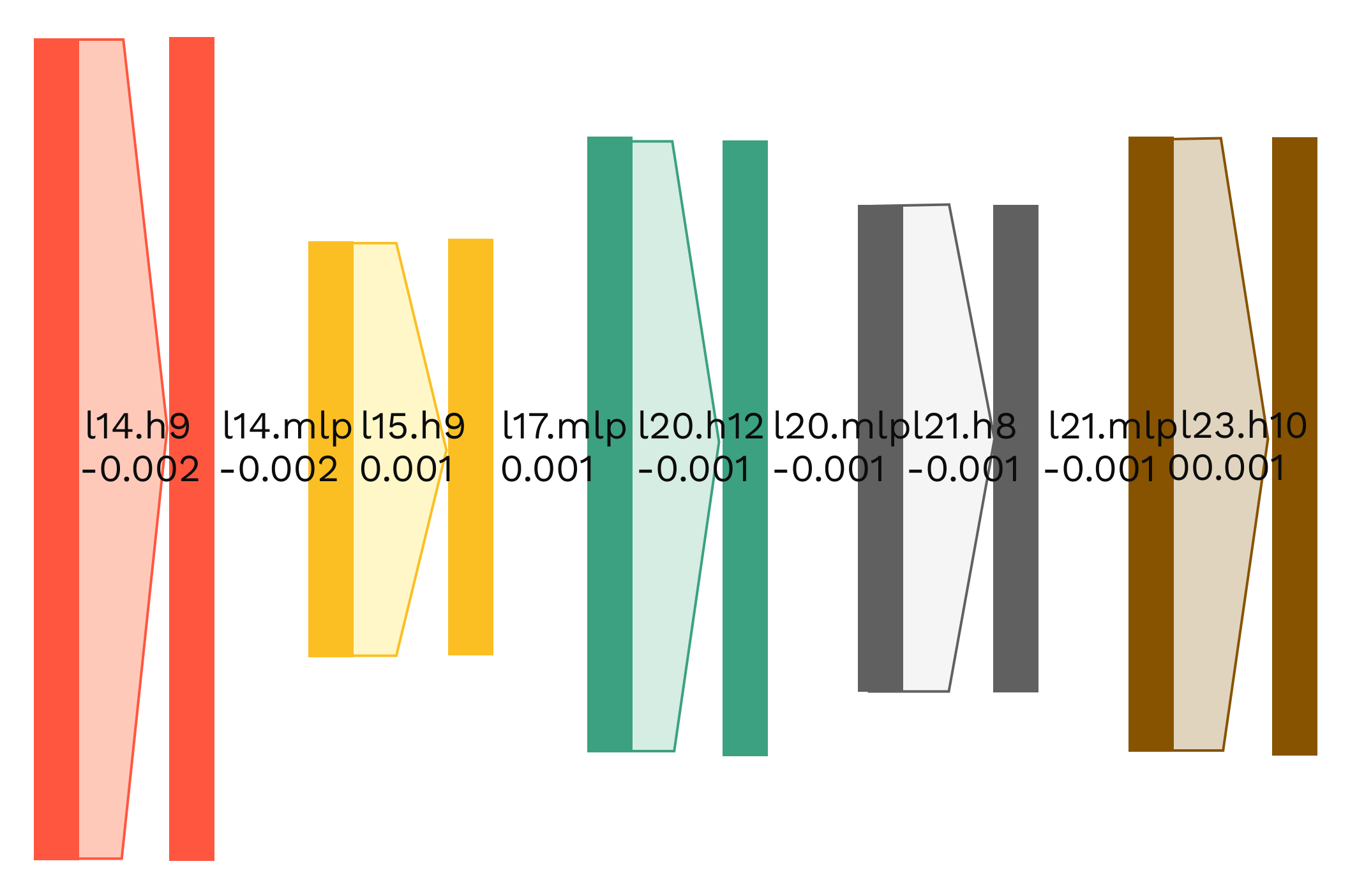}
        \label{fig:eap_fig10}
    \end{minipage}
    \hfill
    \begin{minipage}{0.24\textwidth}
        \centering
        \includegraphics[width=\linewidth]{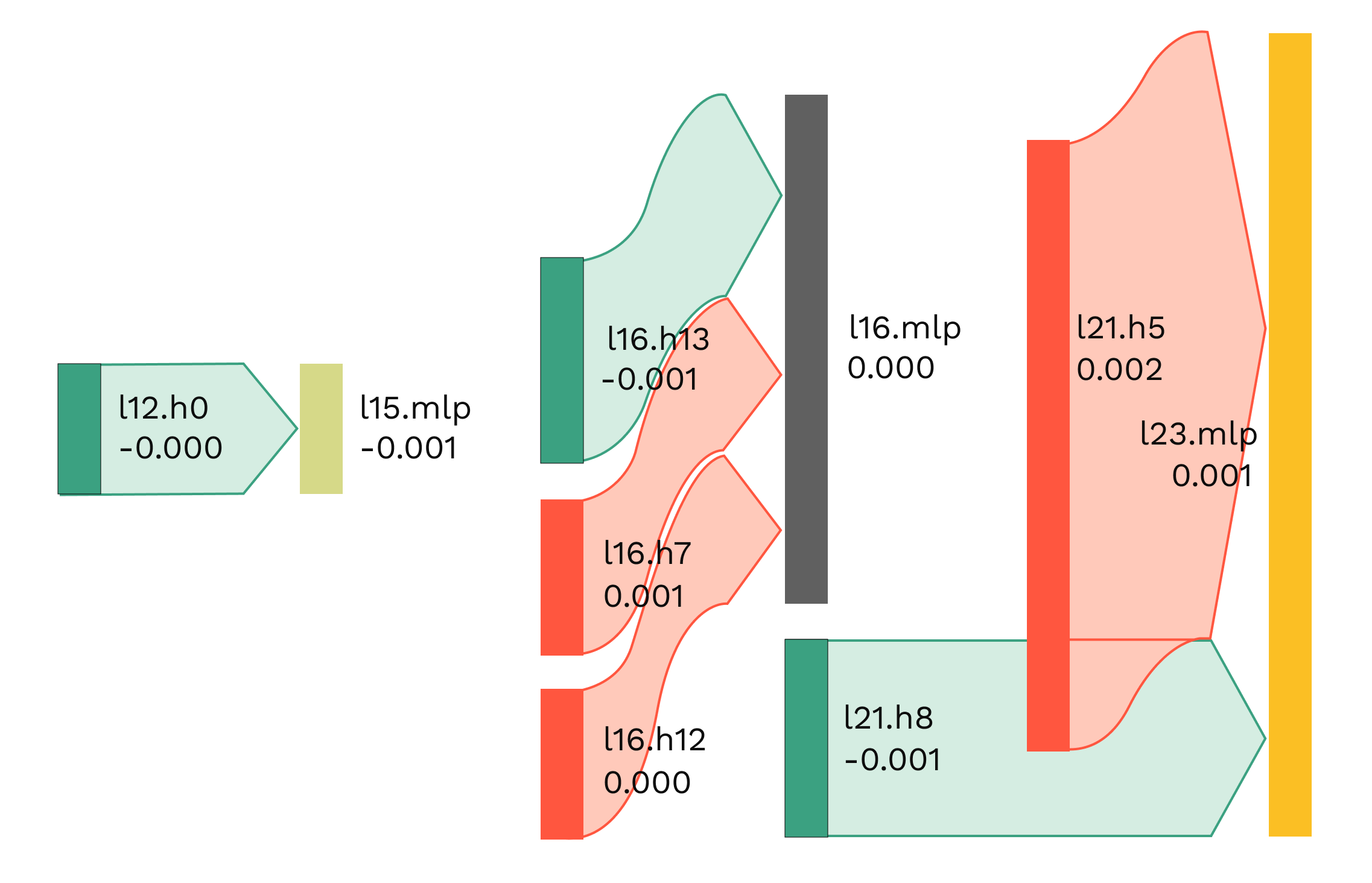}
        \label{fig:eap_fig11}
    \end{minipage}
    \hfill
    \begin{minipage}{0.24\textwidth}
        \centering
        \includegraphics[width=\linewidth]{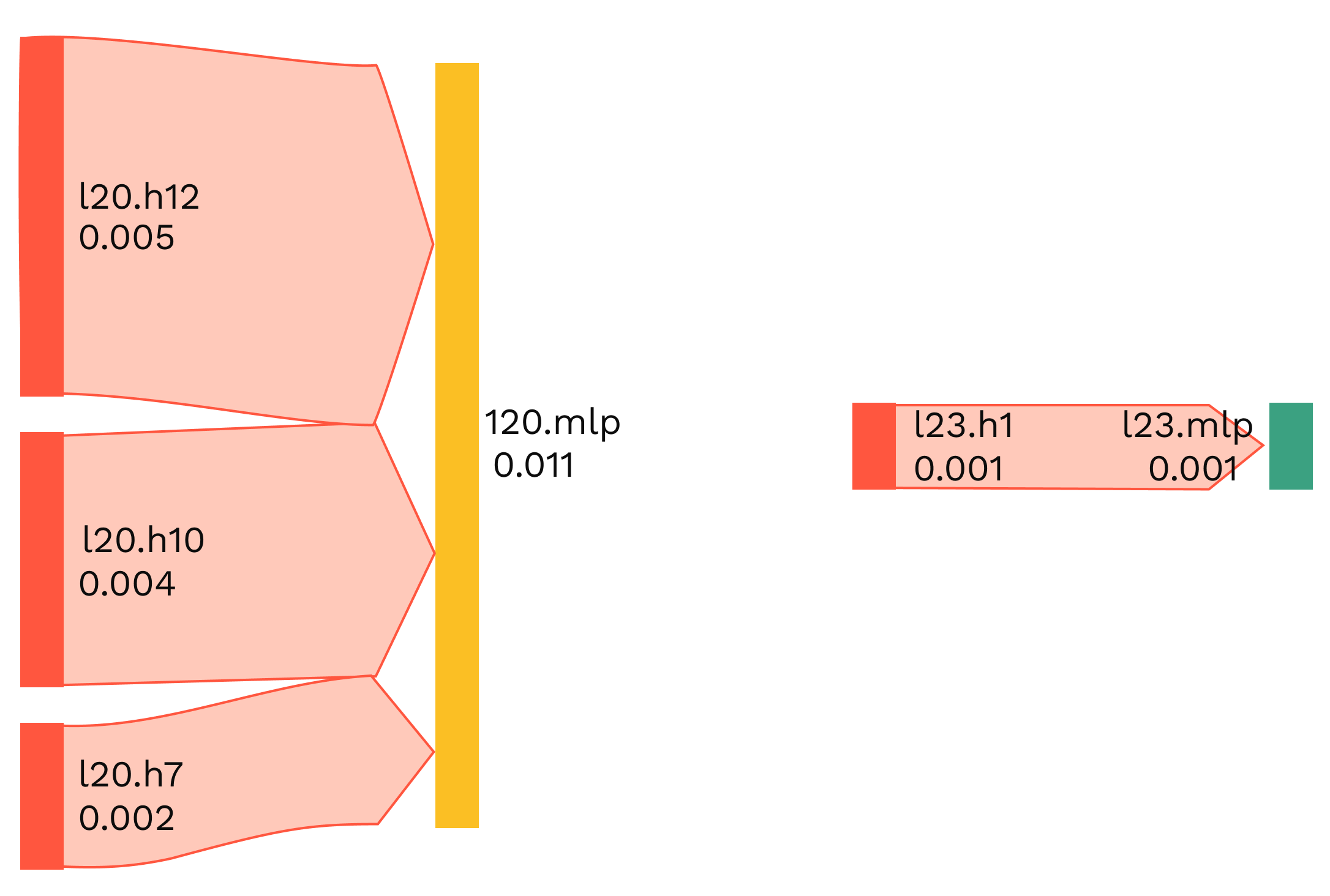}
        \label{fig:eap_fig12}
    \end{minipage}

    \vspace{0.5cm} 
    
    \begin{minipage}{0.24\textwidth}
        \centering
        \includegraphics[width=\linewidth]{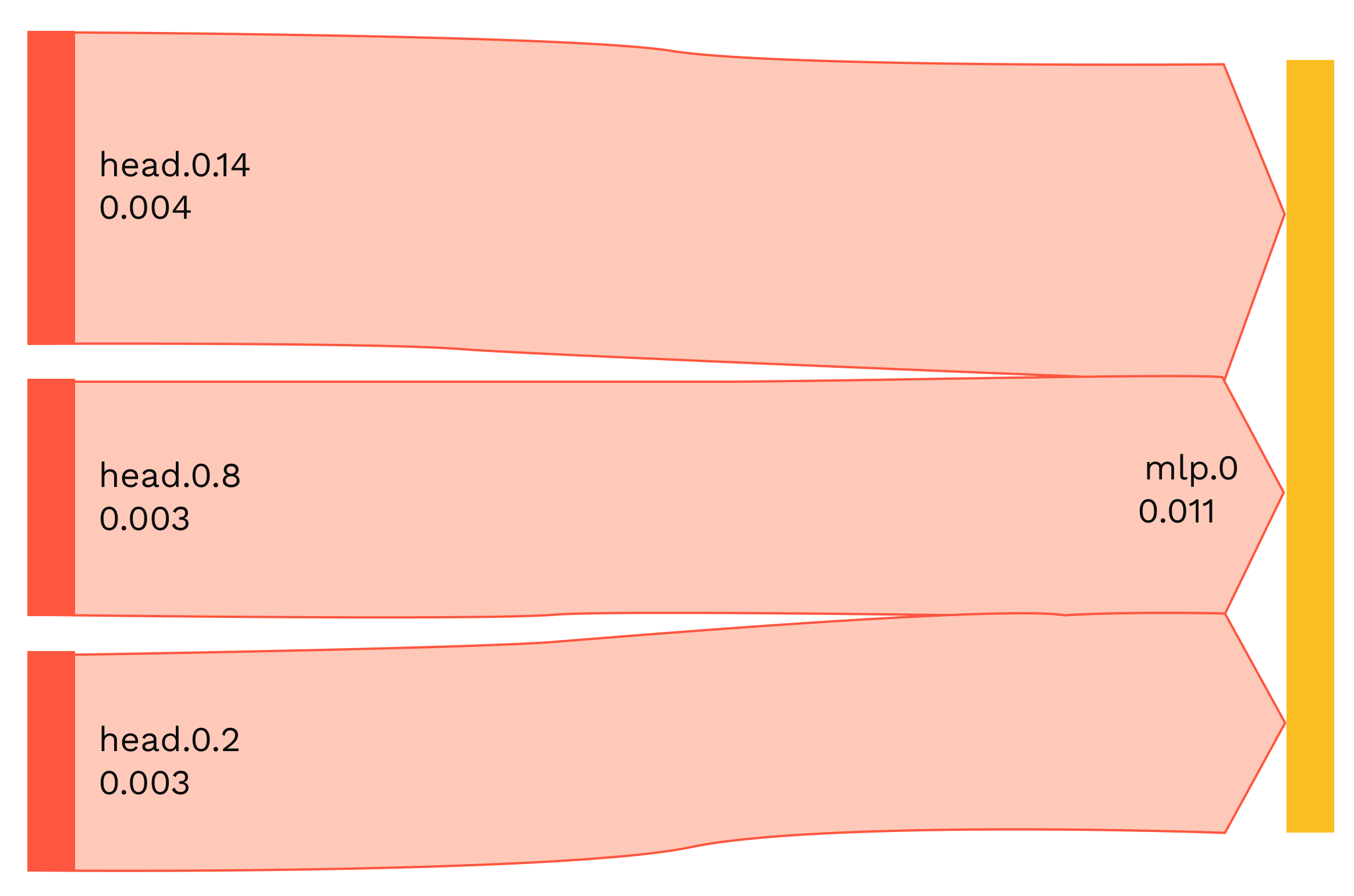}
        \label{fig:eap_fig13}
    \end{minipage}
    \hfill
    \begin{minipage}{0.24\textwidth}
        \centering
        \includegraphics[width=\linewidth]{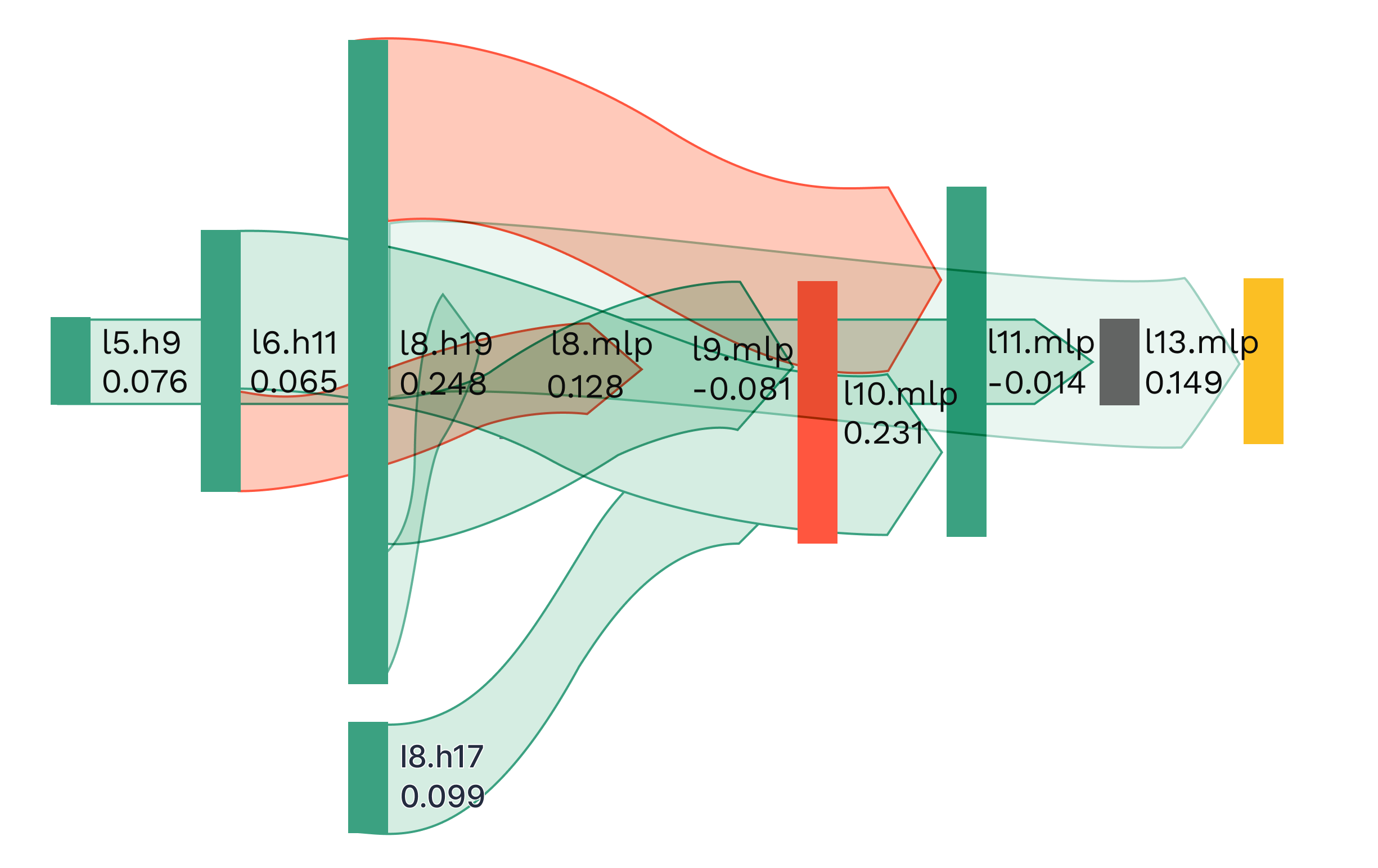}
        \label{fig:eap_fig14}
    \end{minipage}
    \hfill
    \begin{minipage}{0.24\textwidth}
        \centering
        \includegraphics[width=\linewidth]{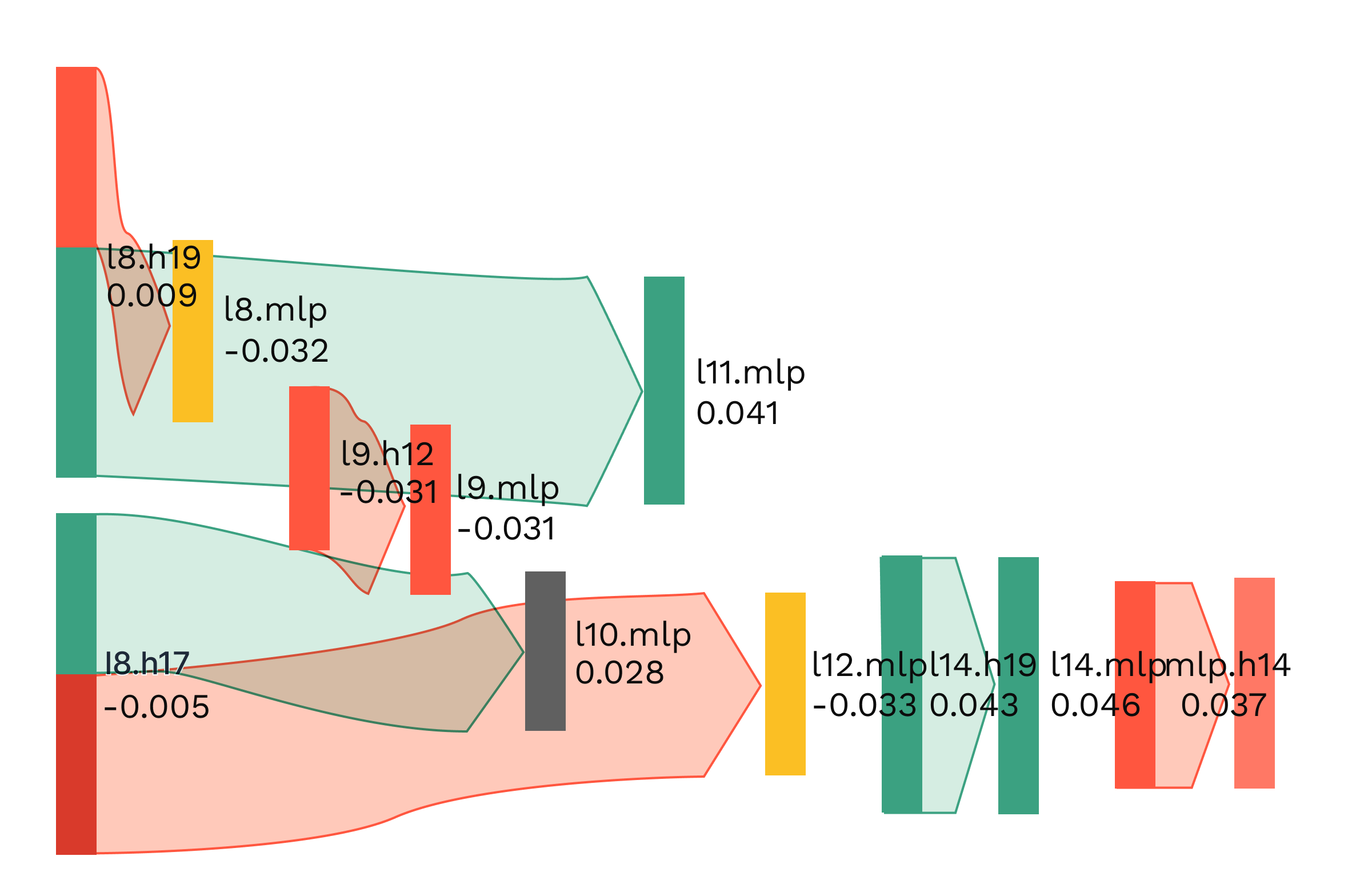}
        \label{fig:eap_fig15}
    \end{minipage}
    \hfill
    \begin{minipage}{0.24\textwidth}
        \centering
        \includegraphics[width=\linewidth]{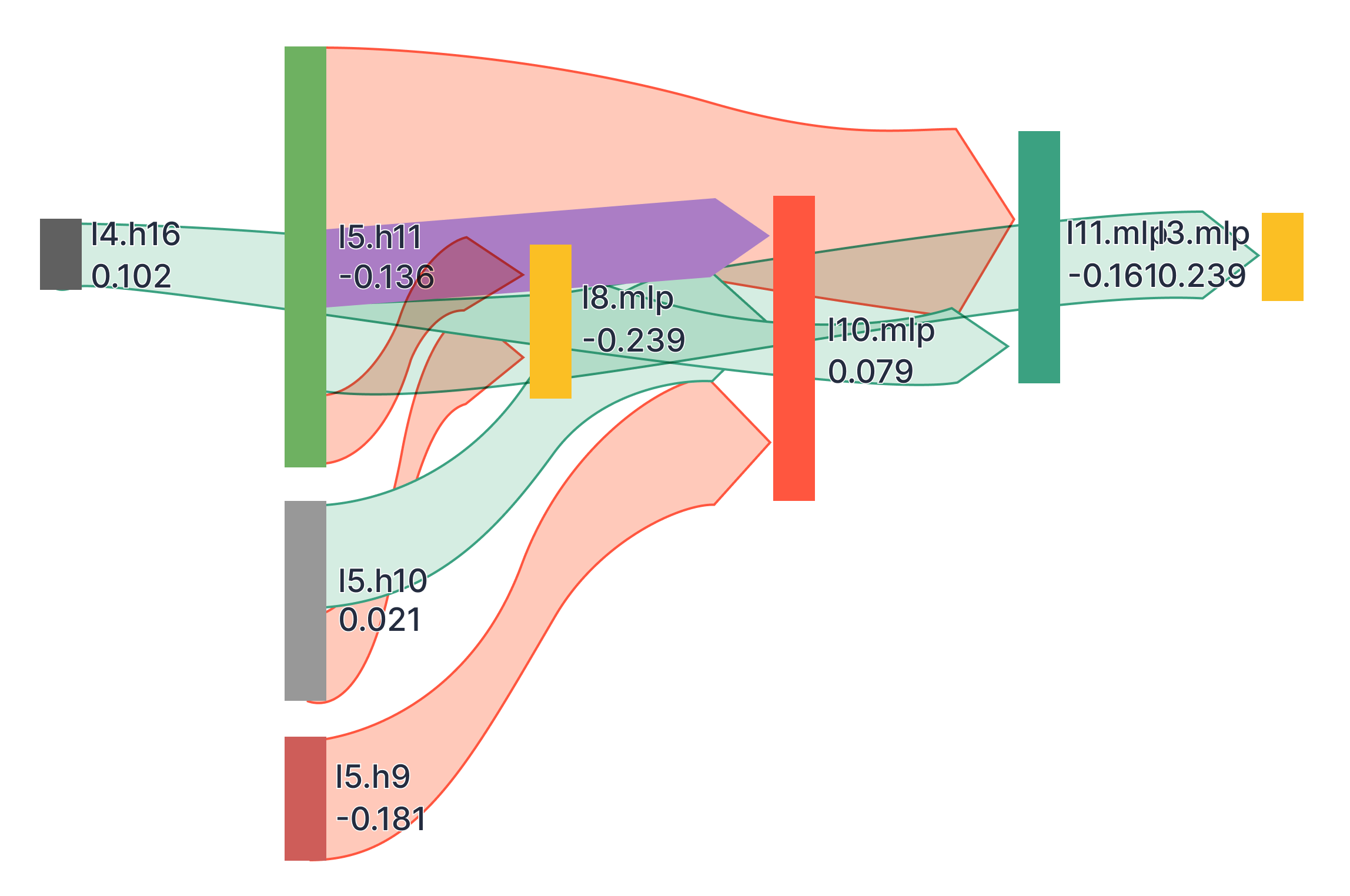}
        \label{fig:eap_fig16}
    \end{minipage}
    \caption{EAP results for different models and command set CS1: BM1\_CS1\_Syn (row 1), BM1\_CS3\_Syn (row 2), BM2\_CS1\_Syn (row 3) and BM3\_CS1\_Syn (row 4). Each row presents results for EngTableName, EngFieldName, DefTableName, and DefFieldName in that order.}
    \label{fig:eap_grid_cs1}
\end{figure}

\begin{figure}
    \centering
    \begin{minipage}{0.24\textwidth}
        \centering
        \includegraphics[width=\linewidth]{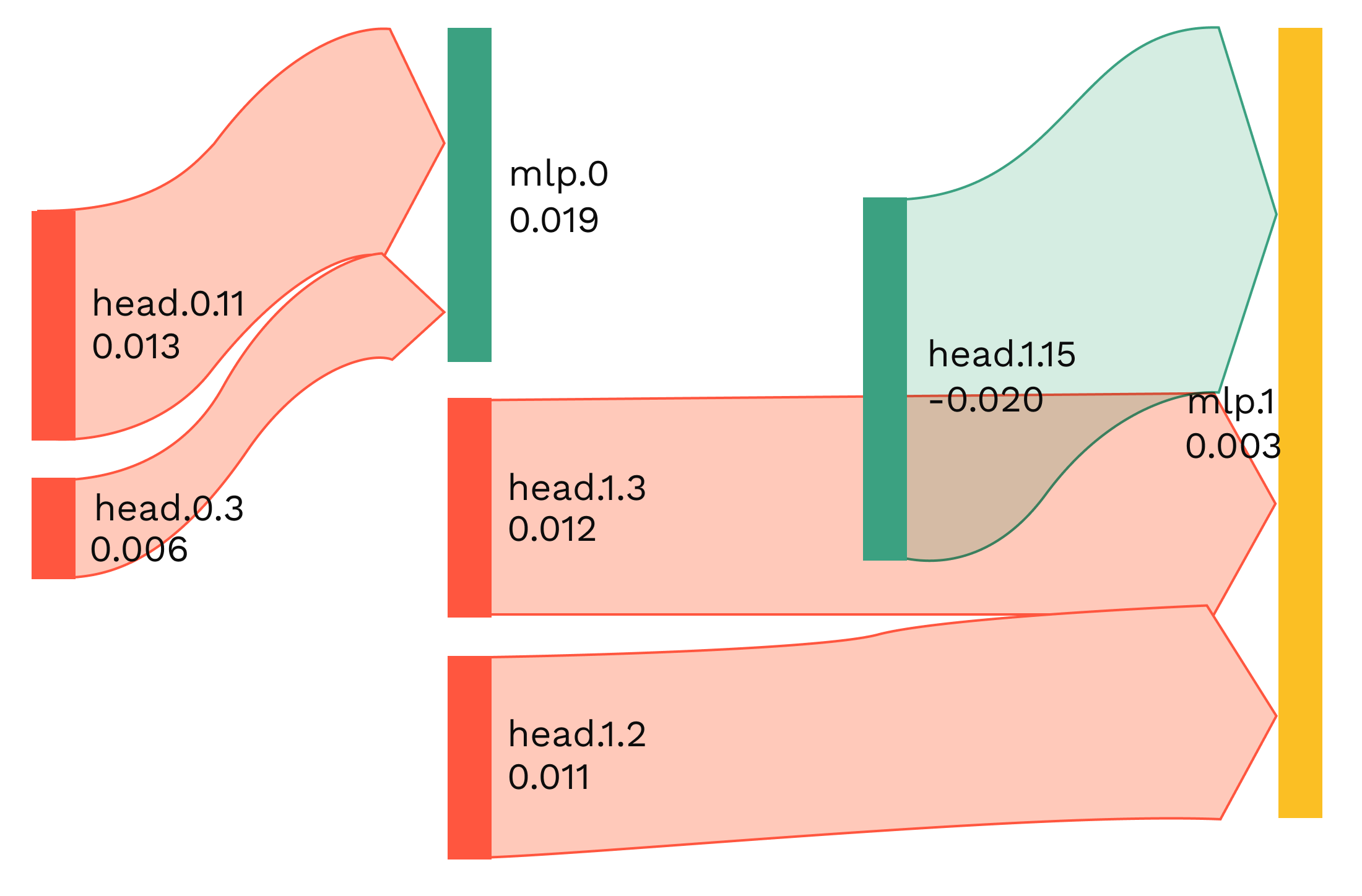}
        \label{fig:eap_cs2_fig1}
    \end{minipage}
    \hfill
    \begin{minipage}{0.24\textwidth}
        \centering
        \includegraphics[width=\linewidth]{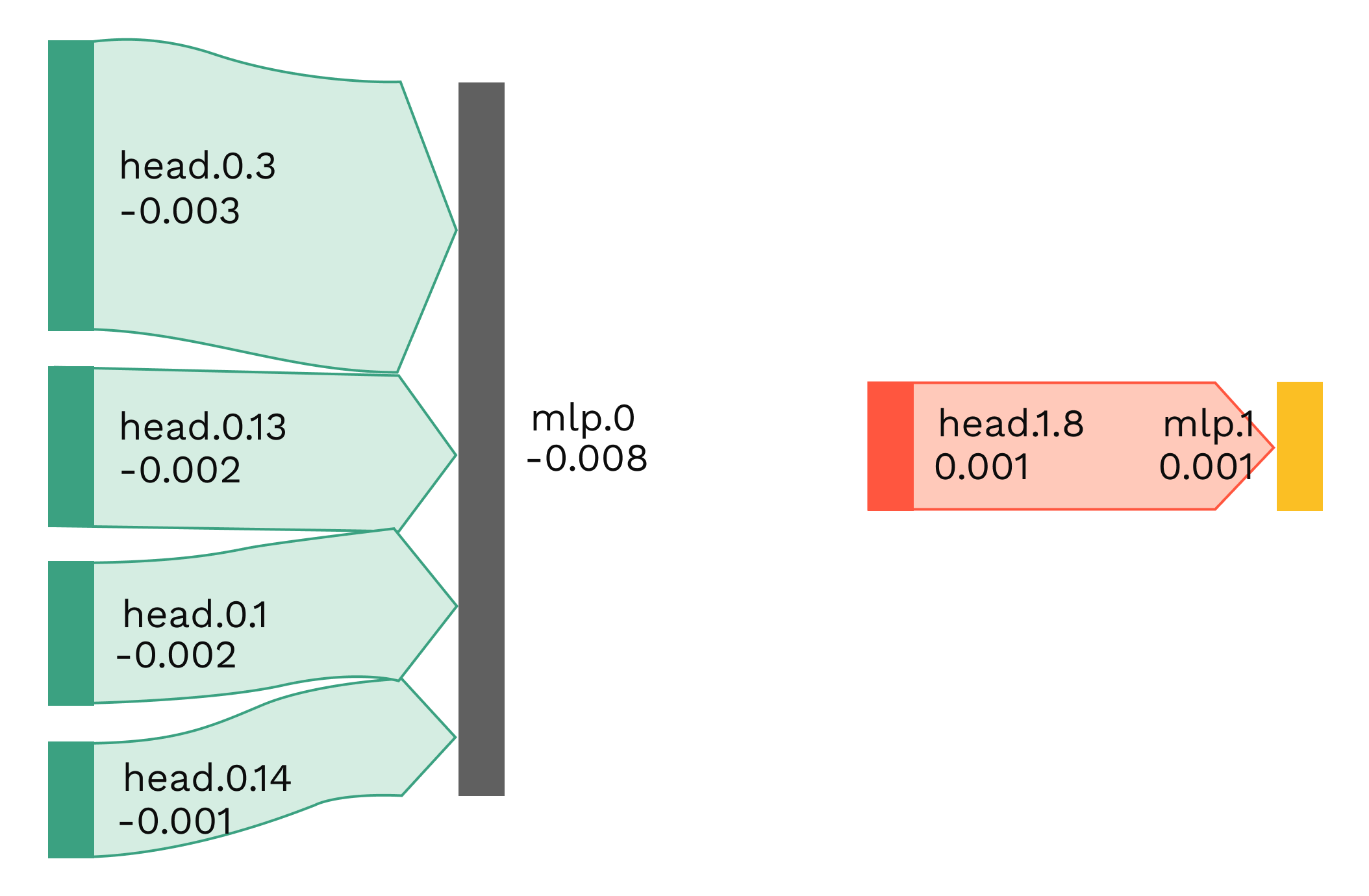}
        \label{fig:eap_cs2_fig2}
    \end{minipage}
    \hfill
    \begin{minipage}{0.24\textwidth}
        \centering
        \includegraphics[width=\linewidth]{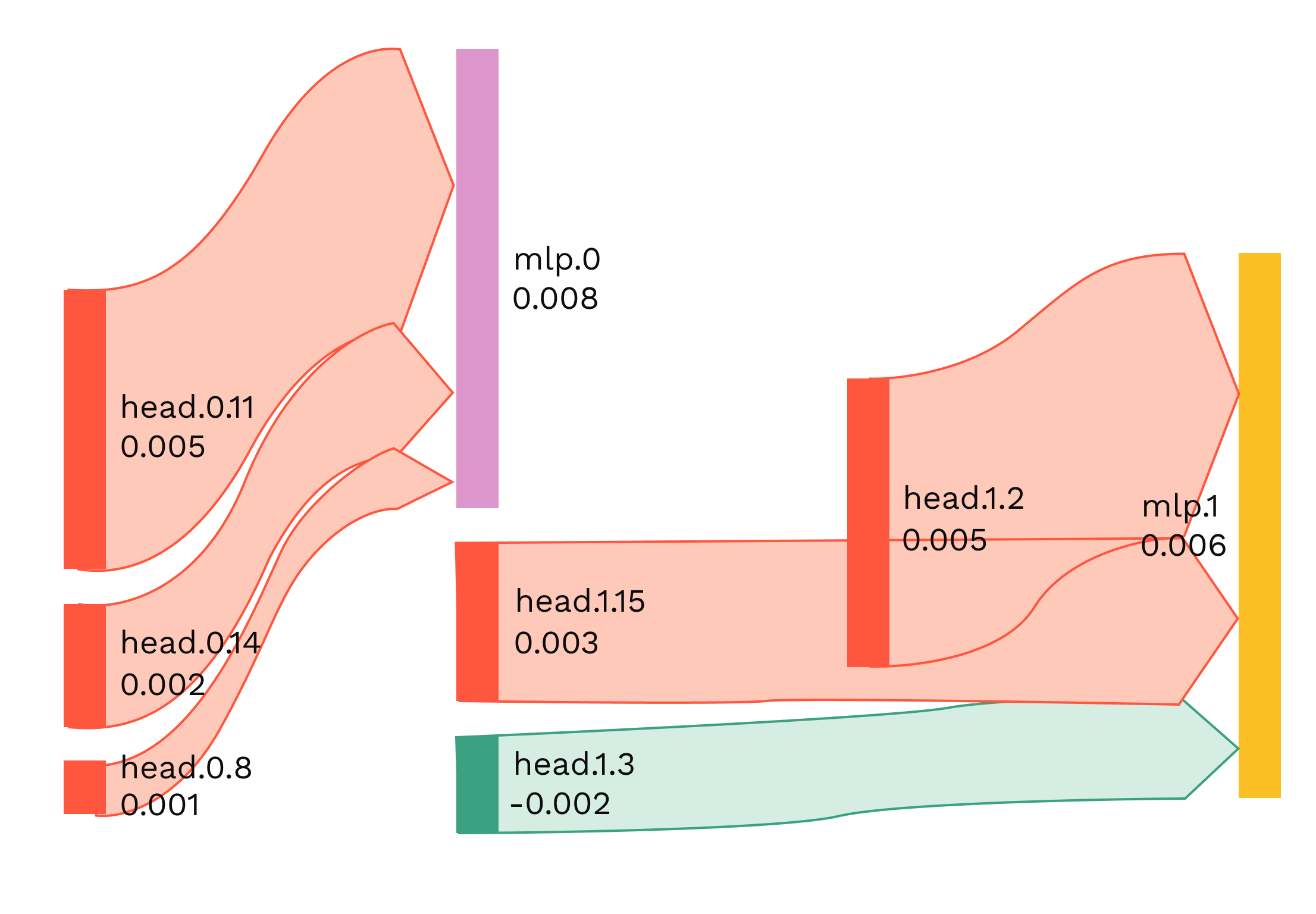}
        \label{fig:eap_cs2_fig3}
    \end{minipage}
    \hfill
    \begin{minipage}{0.24\textwidth}
        \centering
        \includegraphics[width=\linewidth]{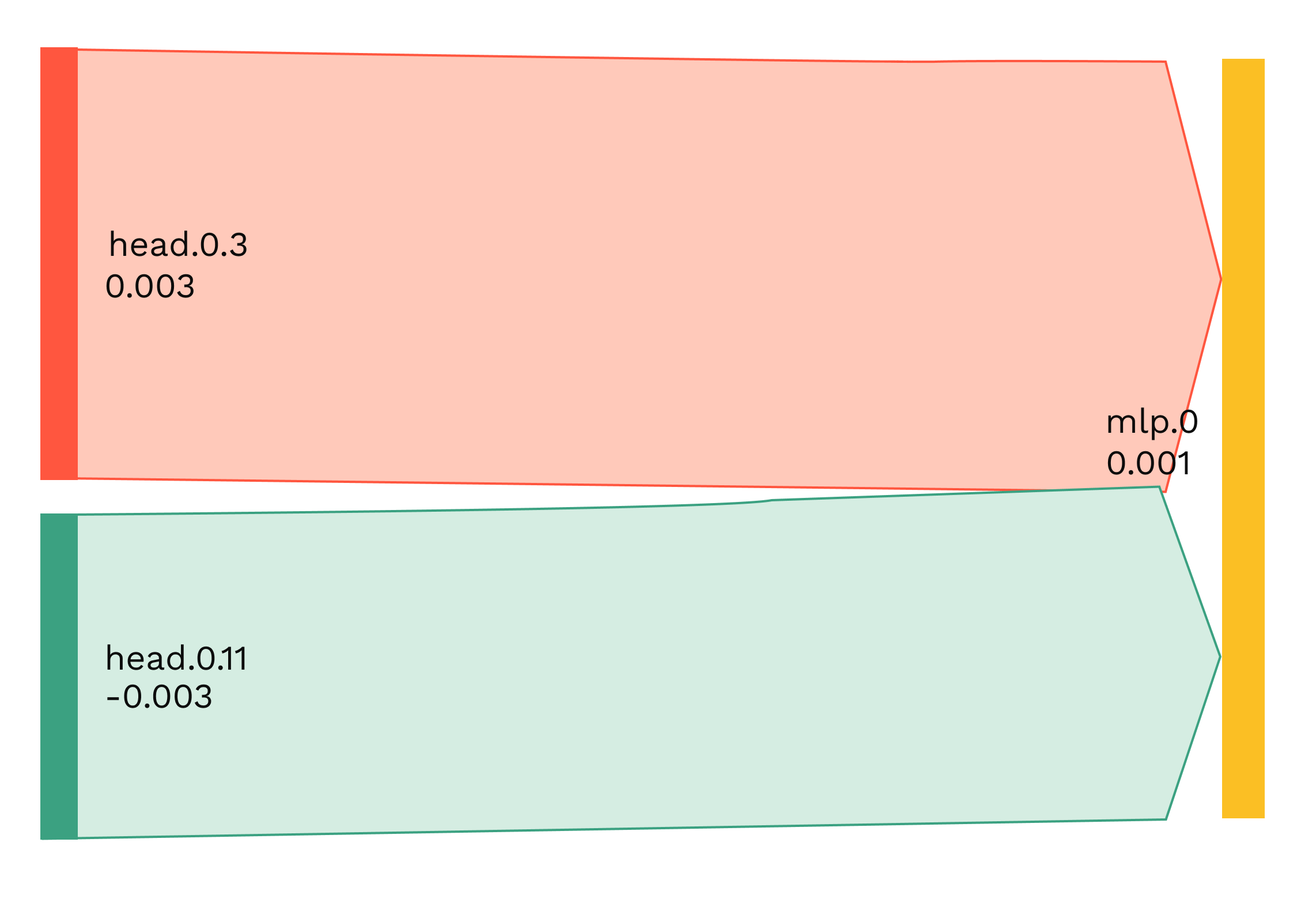}
        \label{fig:eap_cs2_fig4}
    \end{minipage}
    \hfill
    \begin{minipage}{0.24\textwidth}
        \centering
        \includegraphics[width=\linewidth]{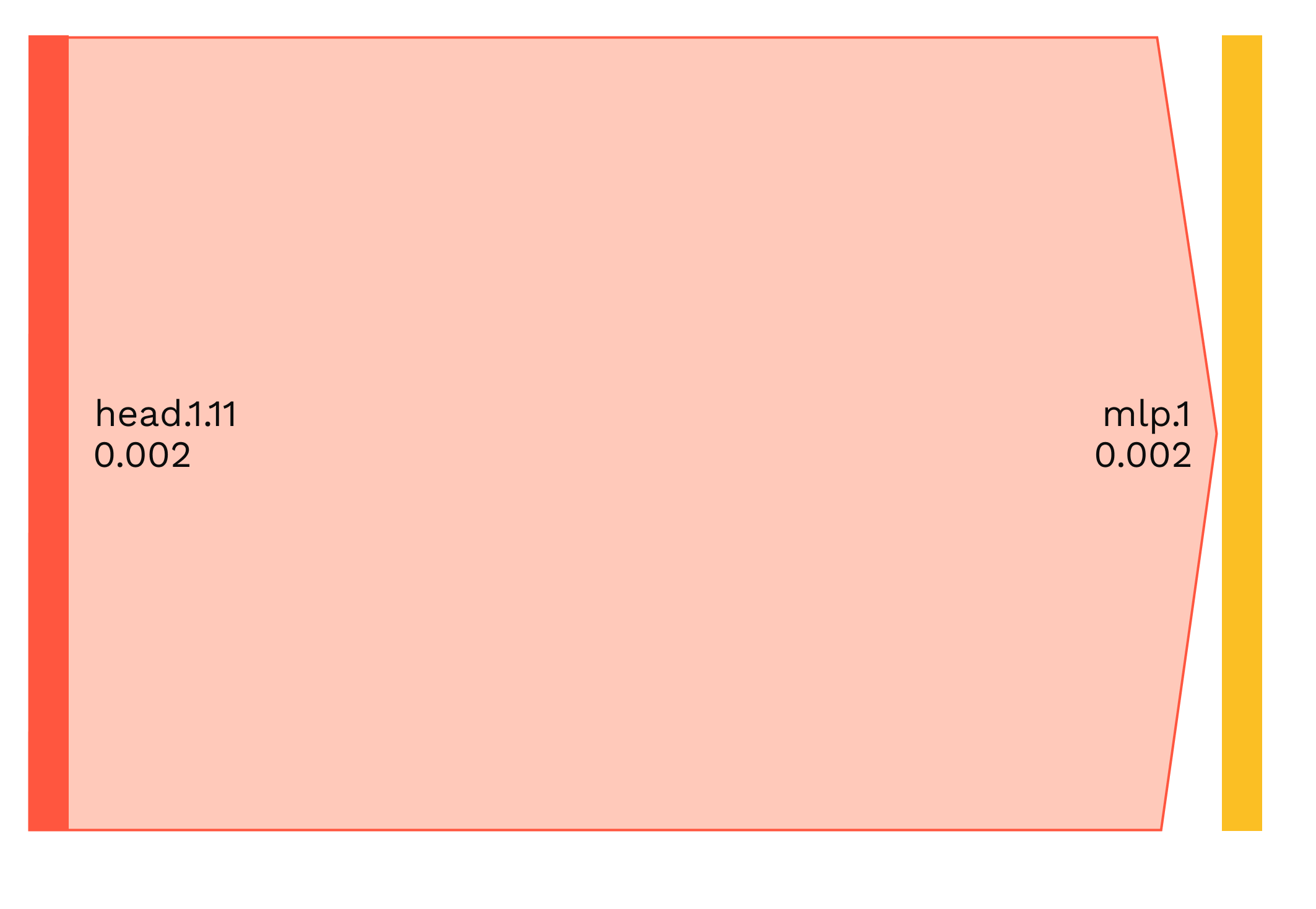}
        \label{fig:eap_cs2_fig5}
    \end{minipage}
    \caption{EAP results for model BM1\_CS3\_Syn on command set CS2: Results are for EngTableName, EngFieldName, DefTableName, DefFieldName and OrderBy in that order.}
    \label{fig:eap_grid_cs2}
\end{figure}

\begin{figure}
    \centering
    \begin{minipage}{0.24\textwidth}
        \centering
        \includegraphics[width=\linewidth]{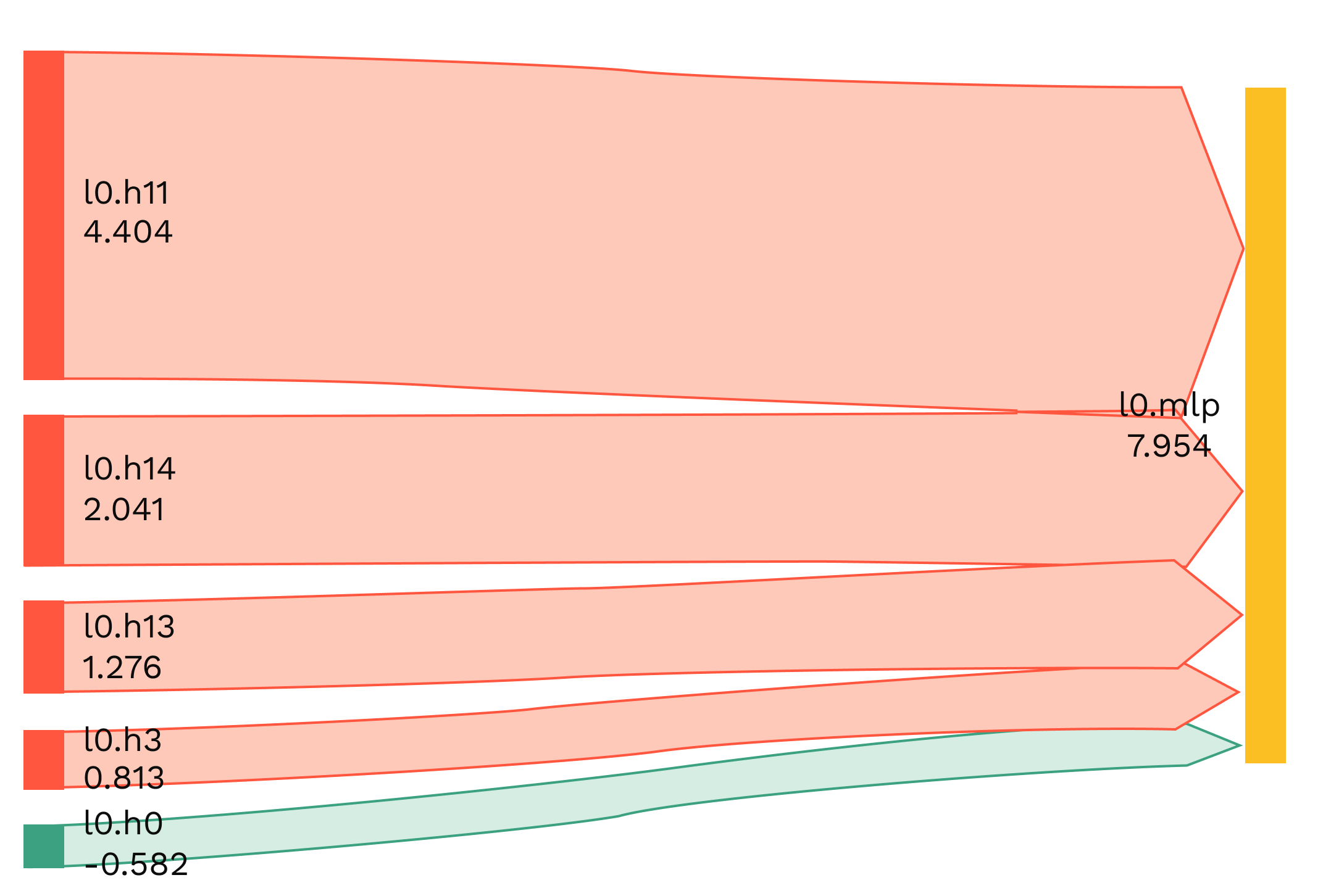}
        \label{fig:eap_cs3_fig1}
    \end{minipage}
    \hfill
    \begin{minipage}{0.24\textwidth}
        \centering
        \includegraphics[width=\linewidth]{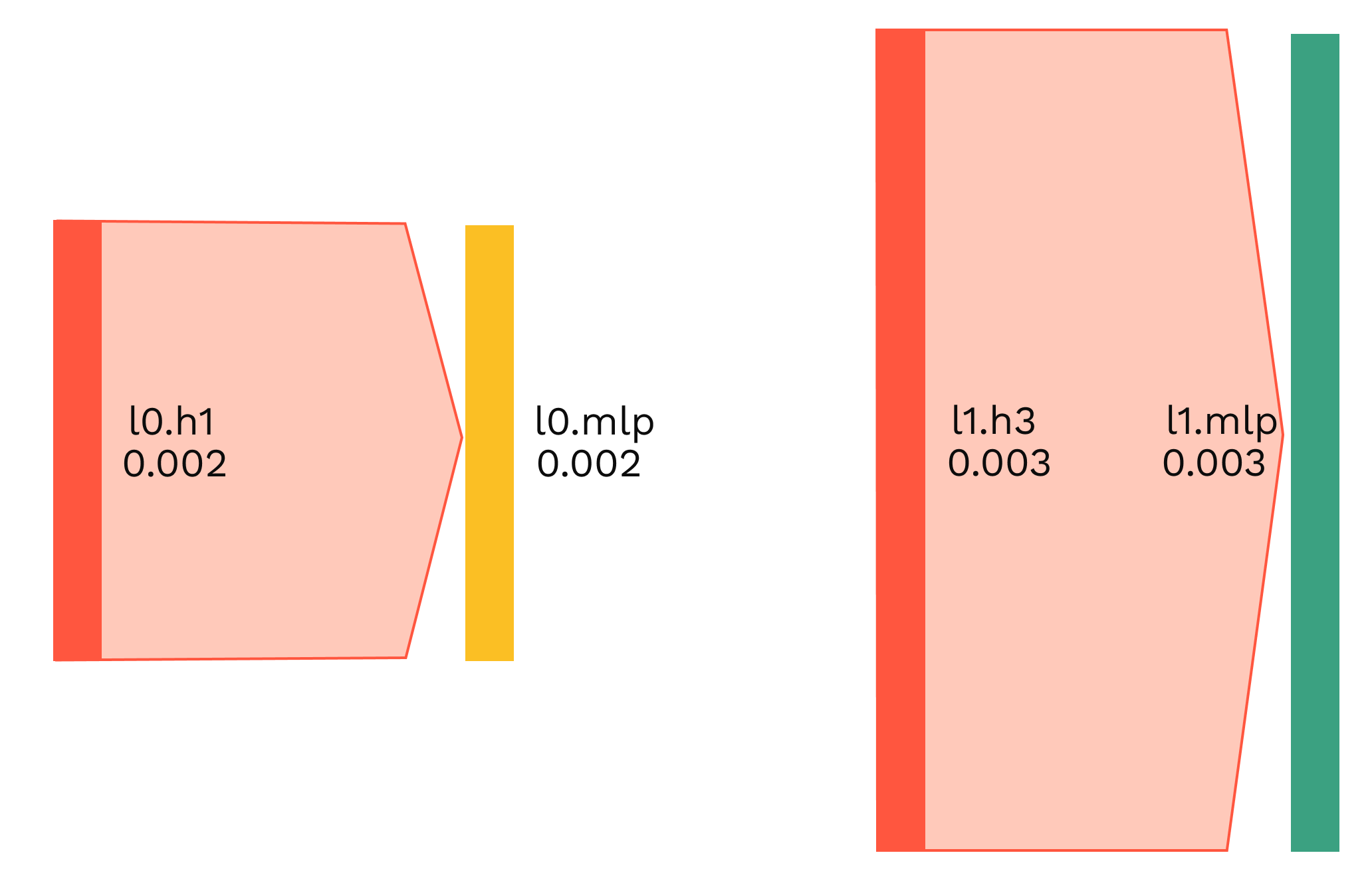}
        \label{fig:eap_cs3_fig2}
    \end{minipage}
    \hfill
    \begin{minipage}{0.24\textwidth}
        \centering
        \includegraphics[width=\linewidth]{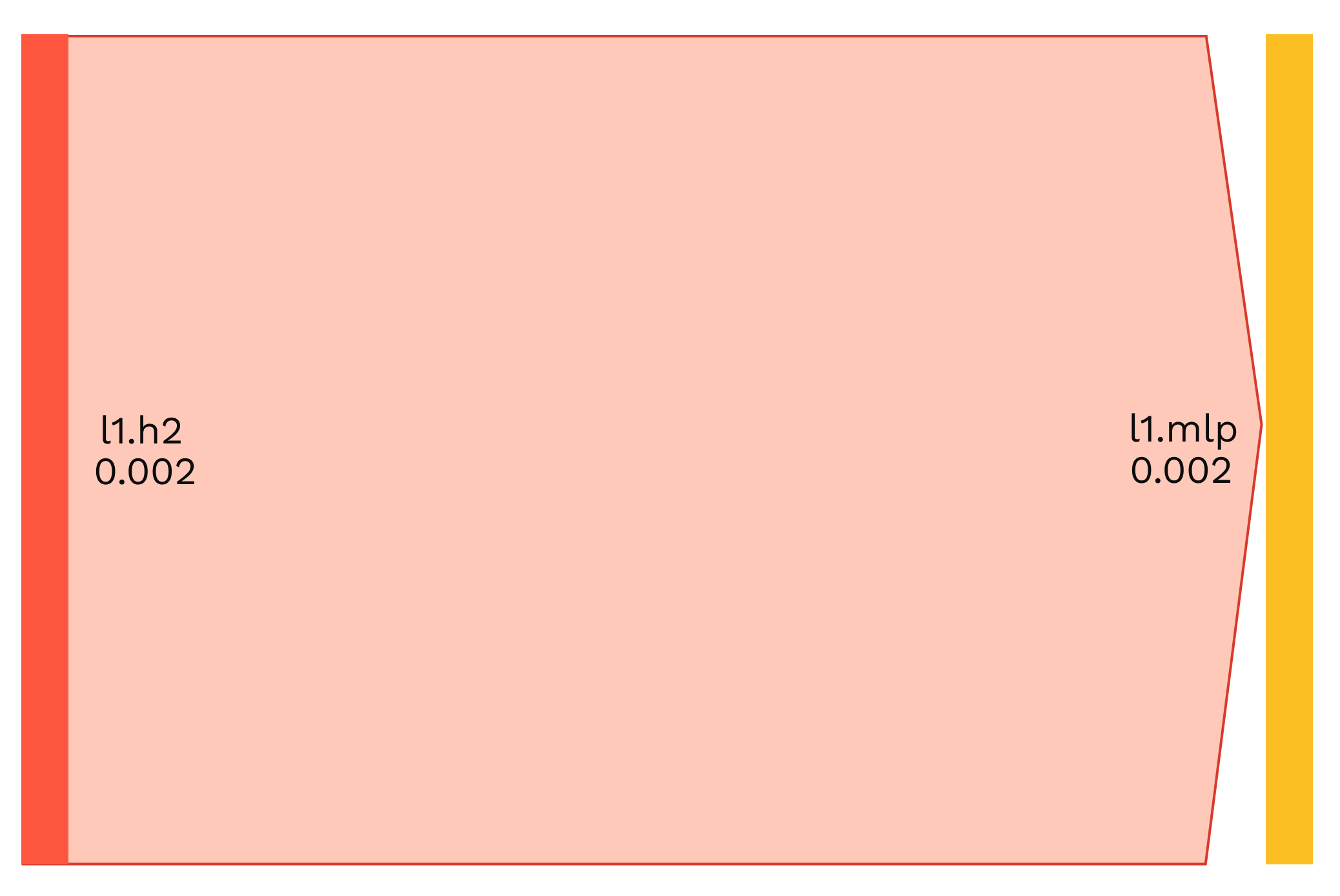}
        \label{fig:eap_cs3_fig3}
    \end{minipage}
    \hfill
    \begin{minipage}{0.24\textwidth}
        \centering
        \includegraphics[width=\linewidth]{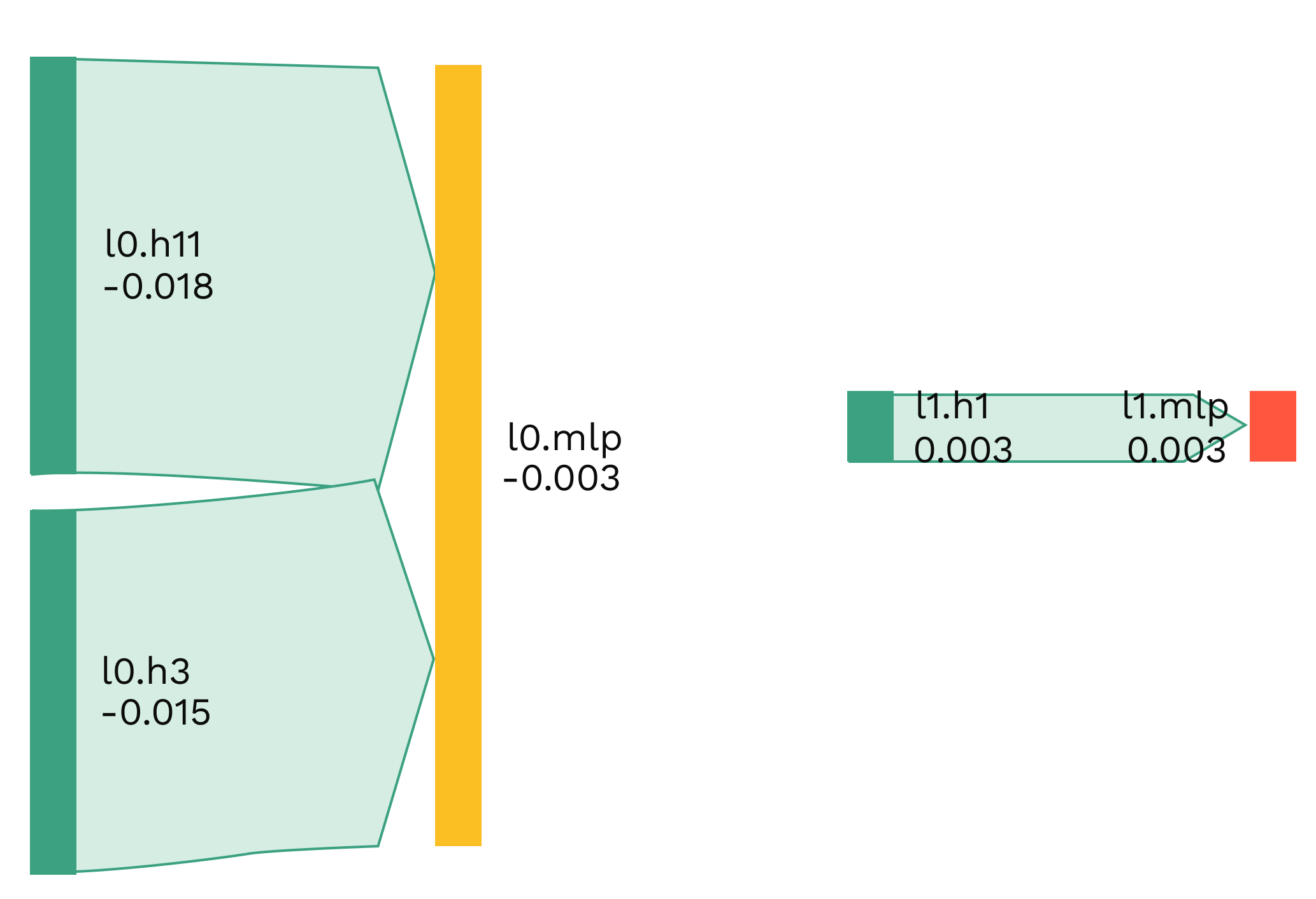}
        \label{fig:eap_cs3_fig4}
    \end{minipage}
    \hfill
    \begin{minipage}{0.24\textwidth}
        \centering
        \includegraphics[width=\linewidth]{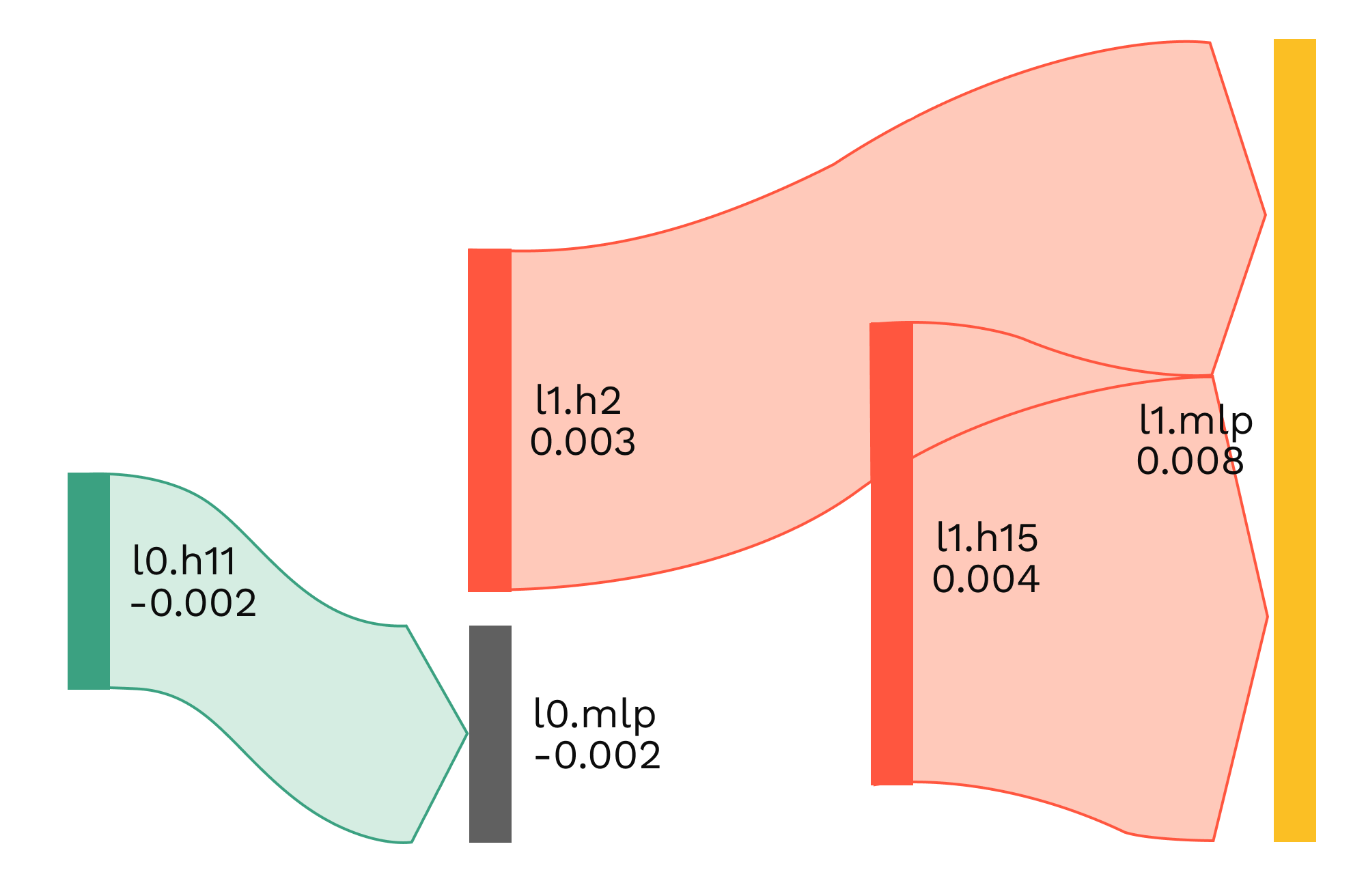}
        \label{fig:eap_cs3_fig5}
    \end{minipage}
    \hfill
    \begin{minipage}{0.24\textwidth}
        \centering
        \includegraphics[width=\linewidth]{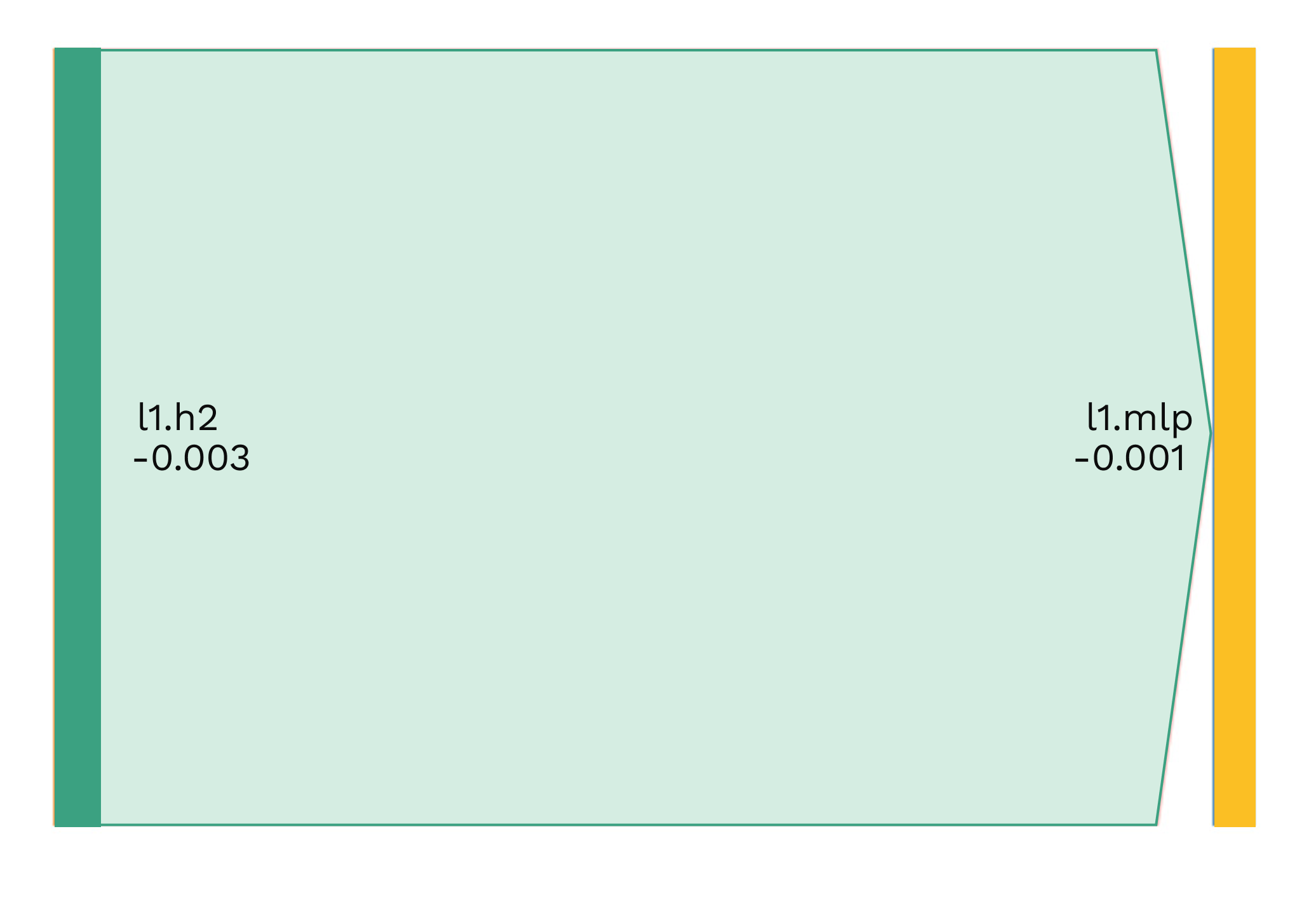}
        \label{fig:eap_cs3_fig6}
    \end{minipage}
    \hfill
    \begin{minipage}{0.24\textwidth}
        \centering
        \includegraphics[width=\linewidth]{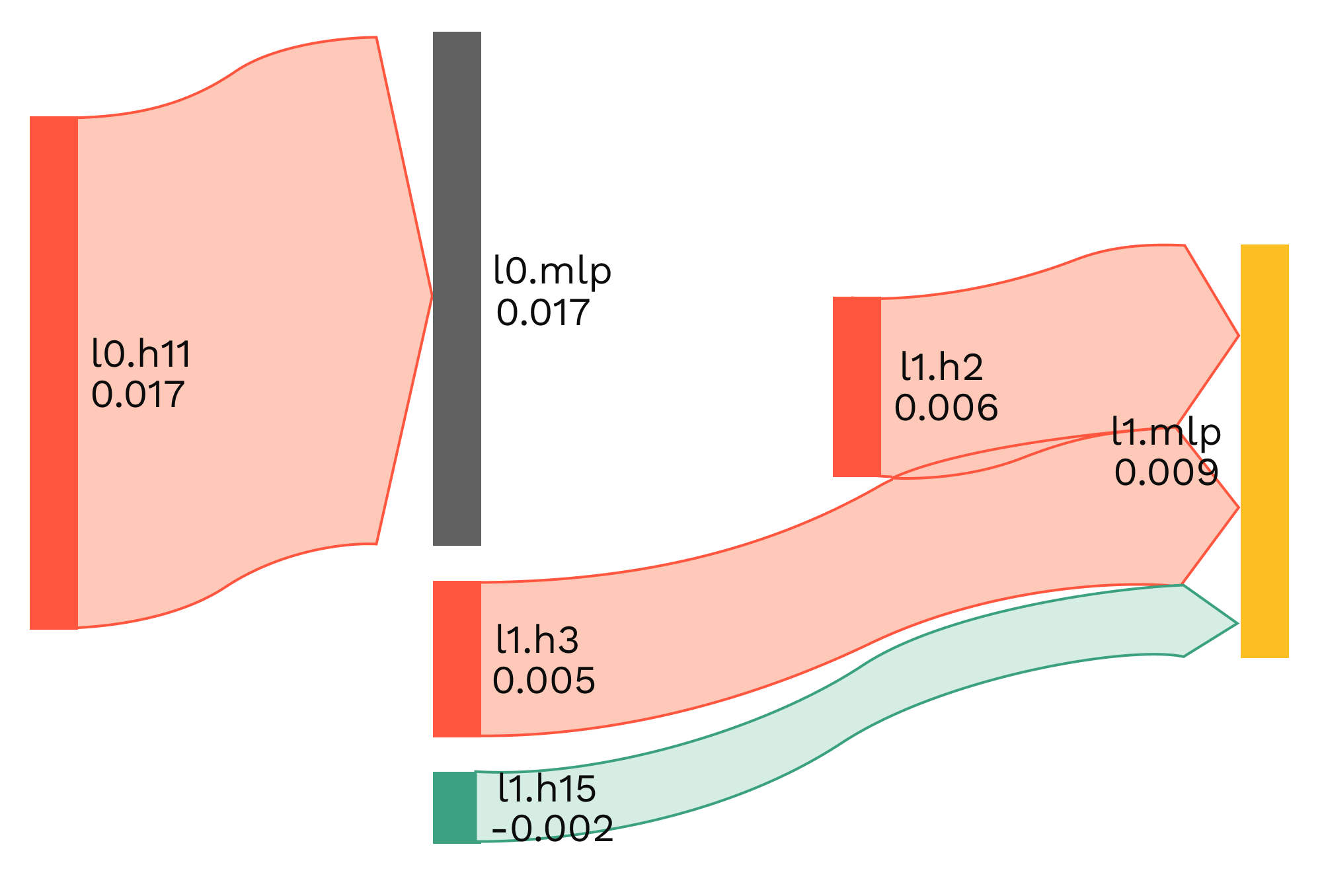}
        \label{fig:eap_cs3_fig7}
    \end{minipage}
    \hfill
    \begin{minipage}{0.24\textwidth}
        \centering
        \includegraphics[width=\linewidth]{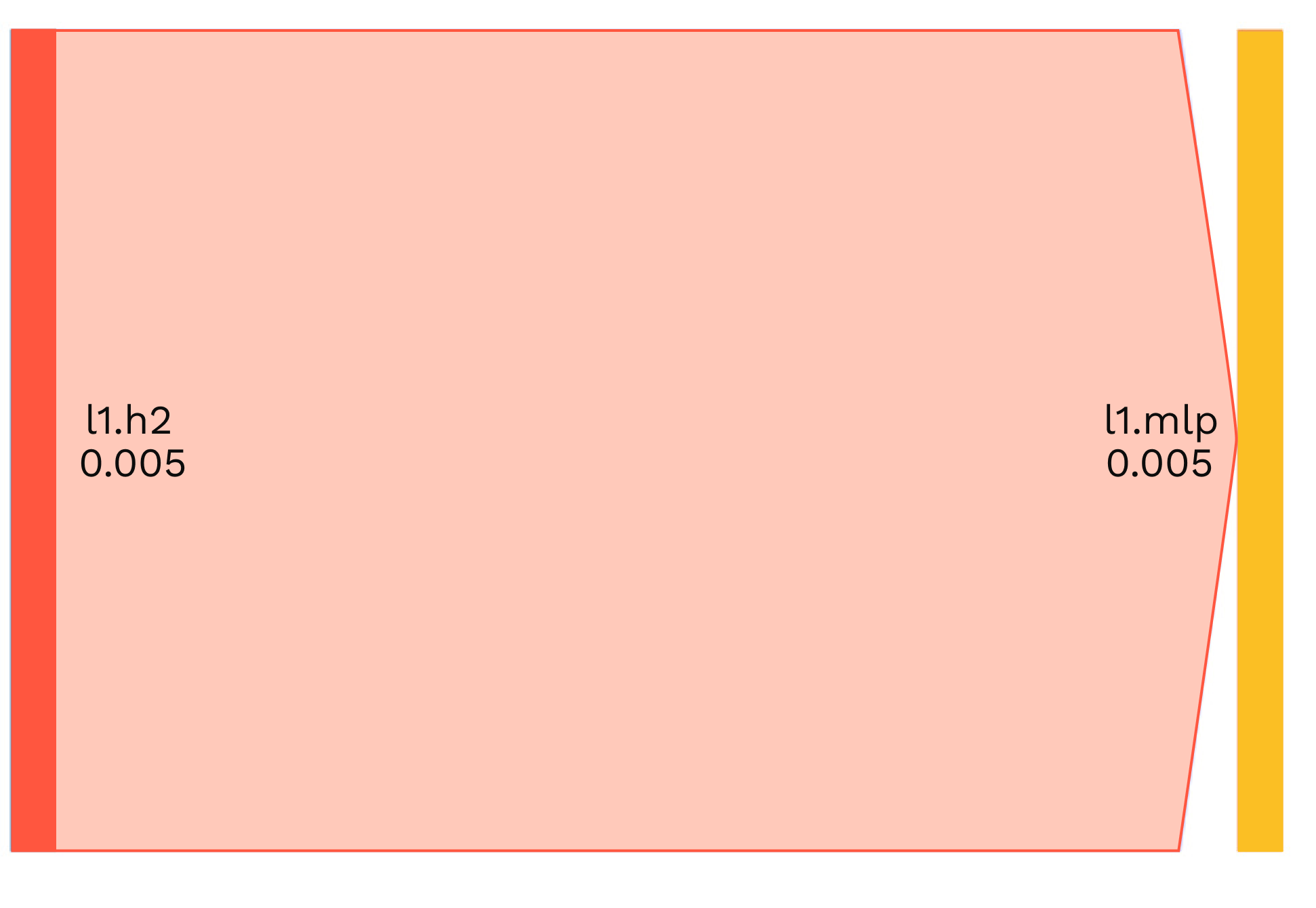}
        \label{fig:eap_cs3_fig8}
    \end{minipage}
    \caption{EAP results for model BM1\_CS3\_Syn on command set CS3: Results are for EngTableName, EngFieldName, DefTableName, DefFieldName, OrderByField, OrderByDirection, AggegrateField and AggregateFunction in that order.}
    \label{fig:eap_grid_cs3}
\end{figure}

After running EAP for all features in both cases, we obtain a set of edges per feature. We then aggregate these edges, extract their corresponding nodes, and use them later in the Ablation Study.
Results of EAP are shown in \hyperref[fig:eap_grid_cs1]{Fig.~\ref{fig:eap_grid_cs1}} for CS1, \hyperref[fig:eap_grid_cs2]{Fig.~\ref{fig:eap_grid_cs2}} for CS2 and \hyperref[fig:eap_grid_cs3]{Fig.~\ref{fig:eap_grid_cs3}} for CS3. We only show results where use\_synonyms\_field and use\_synonyms\_table are both set to False.

\subsection{Ablation Study Results}

\begin{figure*}
    \centering
    \begin{minipage}{0.48\textwidth}
        \centering
        \includegraphics[width=\textwidth]{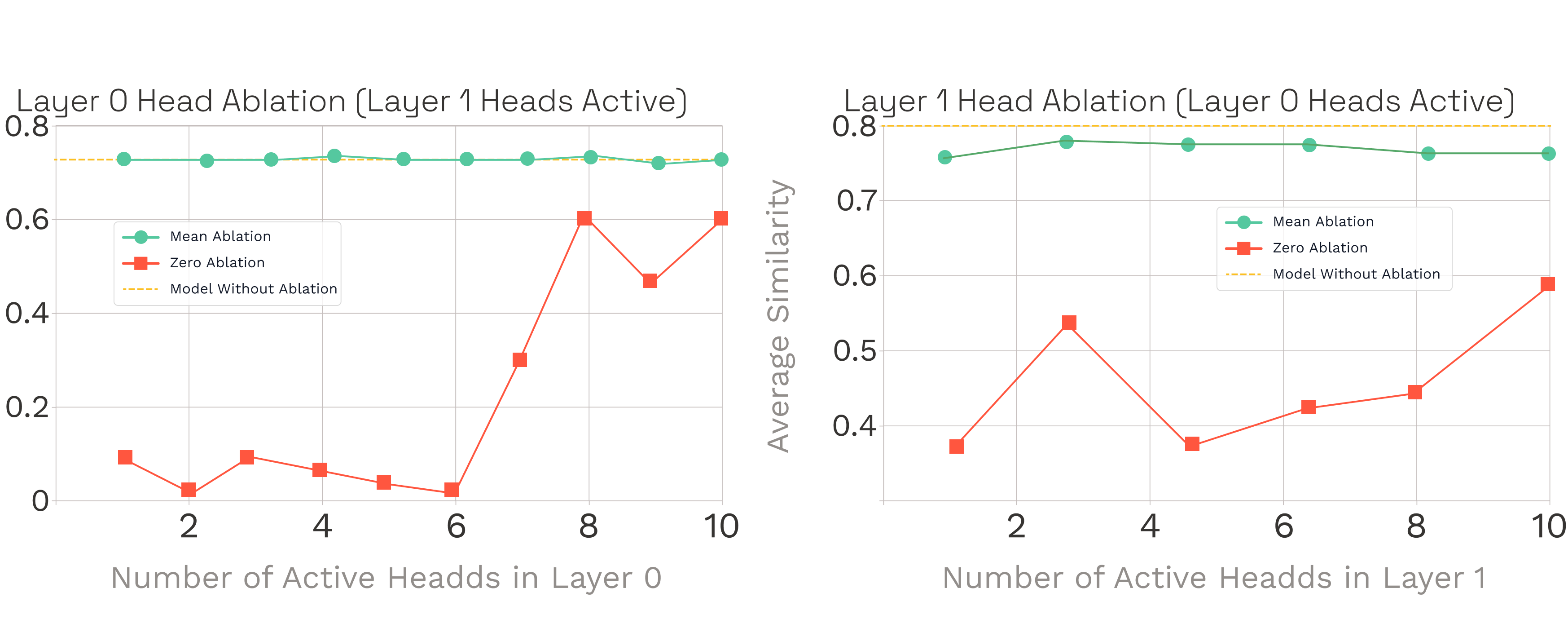}
    \end{minipage}
    \hfill
    \begin{minipage}{0.48\textwidth}
        \centering
        \includegraphics[width=\textwidth]{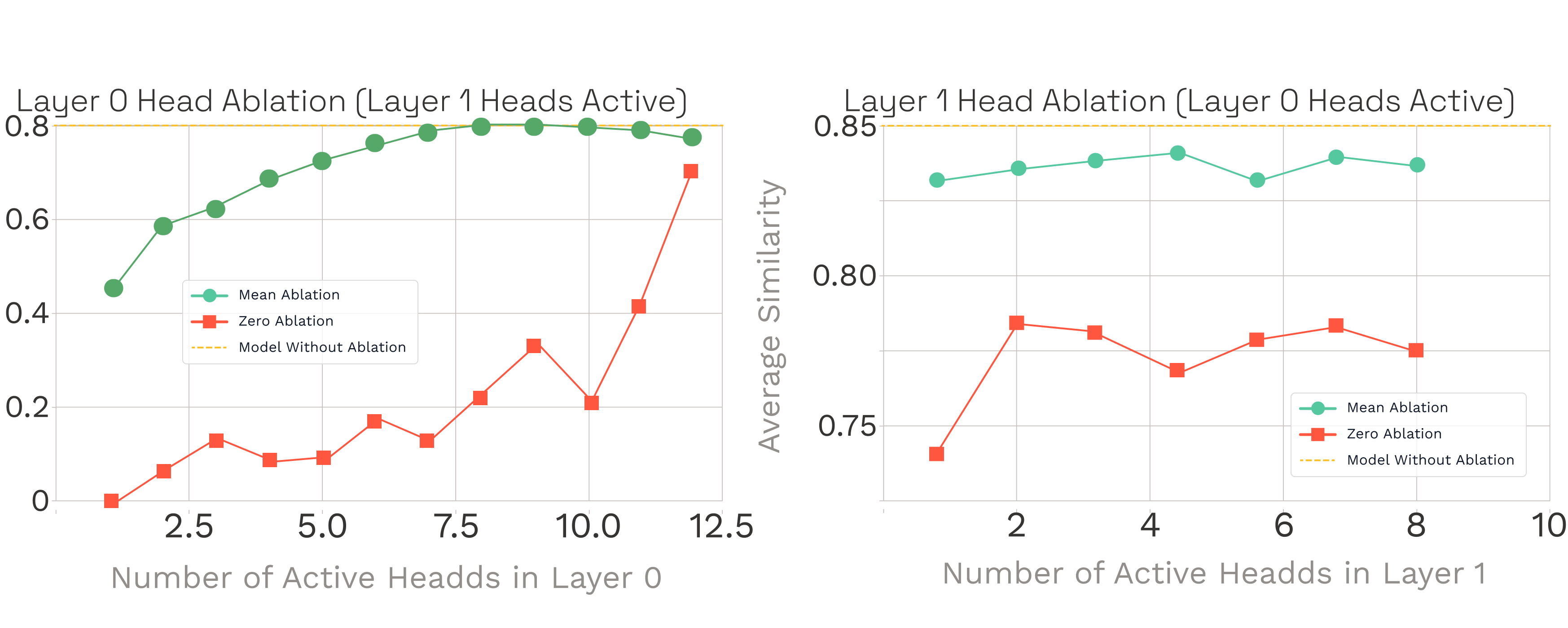}
    \end{minipage}
    
    \vspace{0.5cm} 

    \begin{minipage}{0.48\textwidth}
        \centering
        \includegraphics[width=\textwidth]{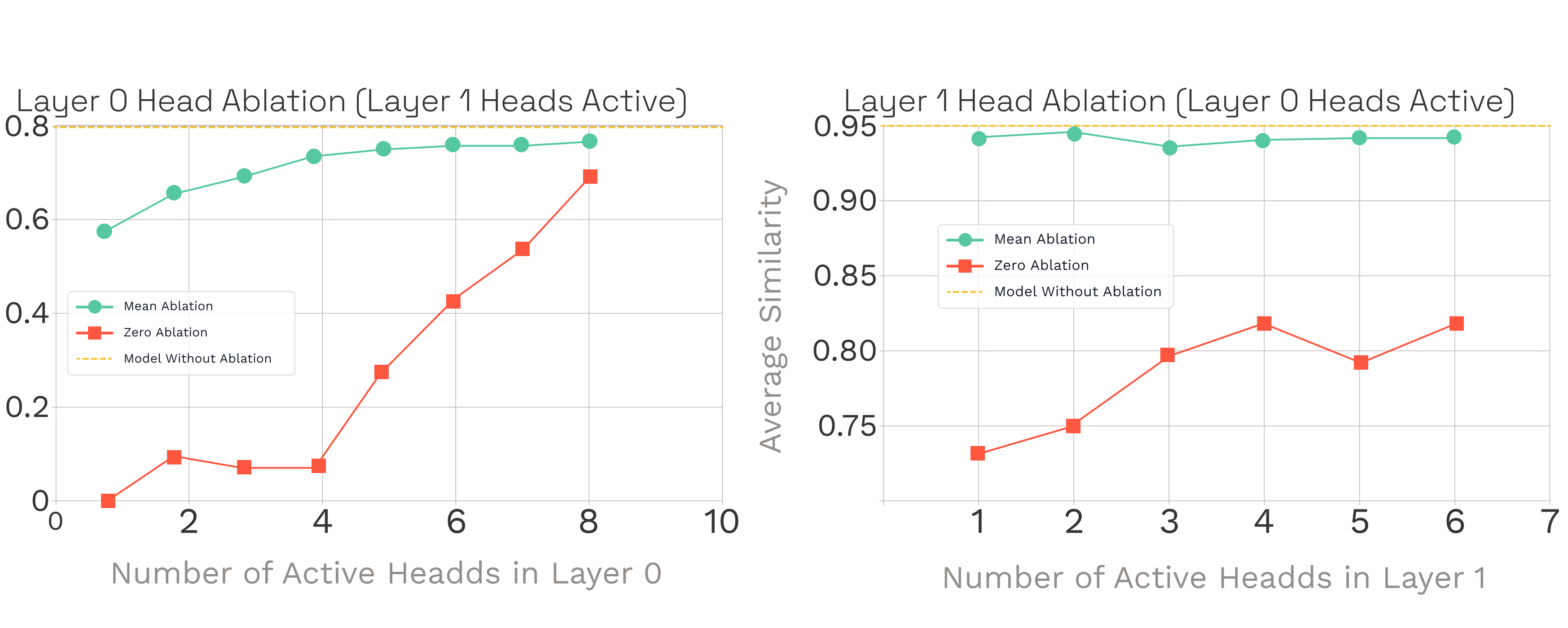}
    \end{minipage}
    \hfill
    \begin{minipage}{0.48\textwidth}
        \centering
        \includegraphics[width=\textwidth]{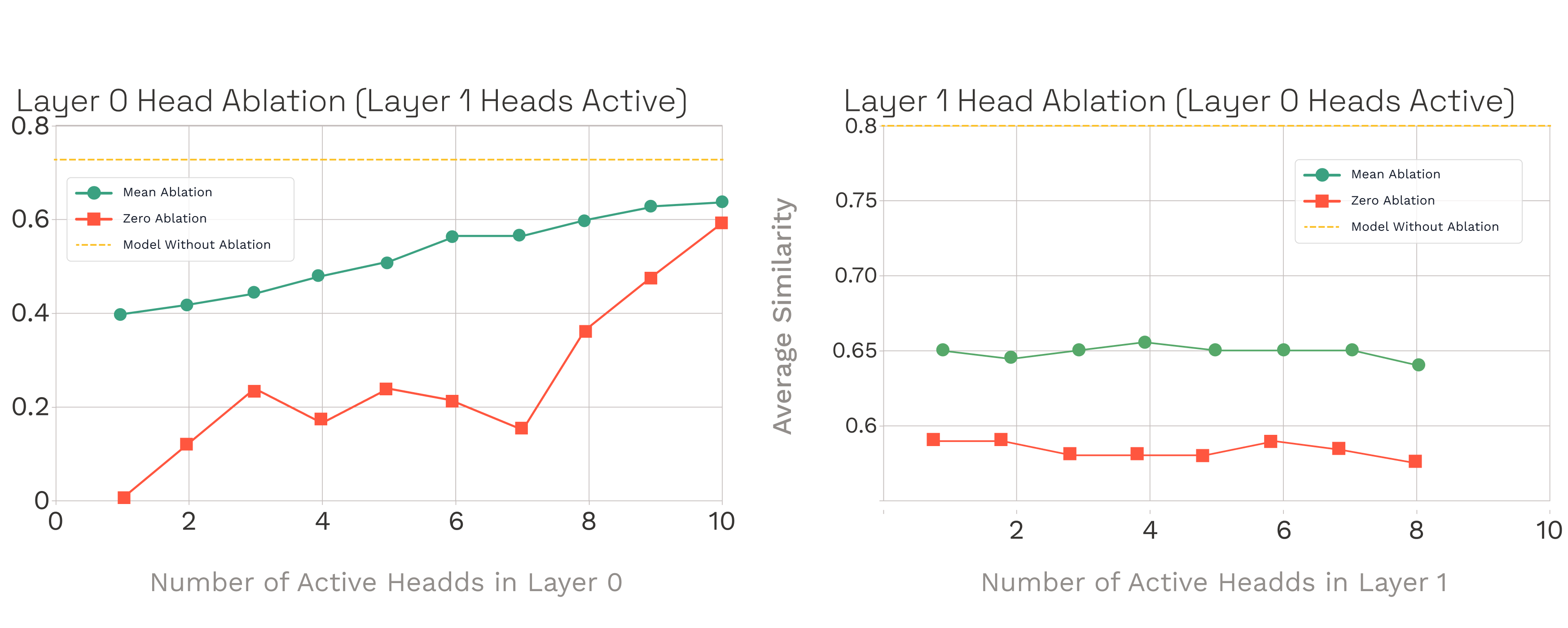}
    \end{minipage}

    \vspace{0.5cm} 

    \begin{minipage}{0.4\textwidth}
        \centering
        \includegraphics[width=\textwidth]{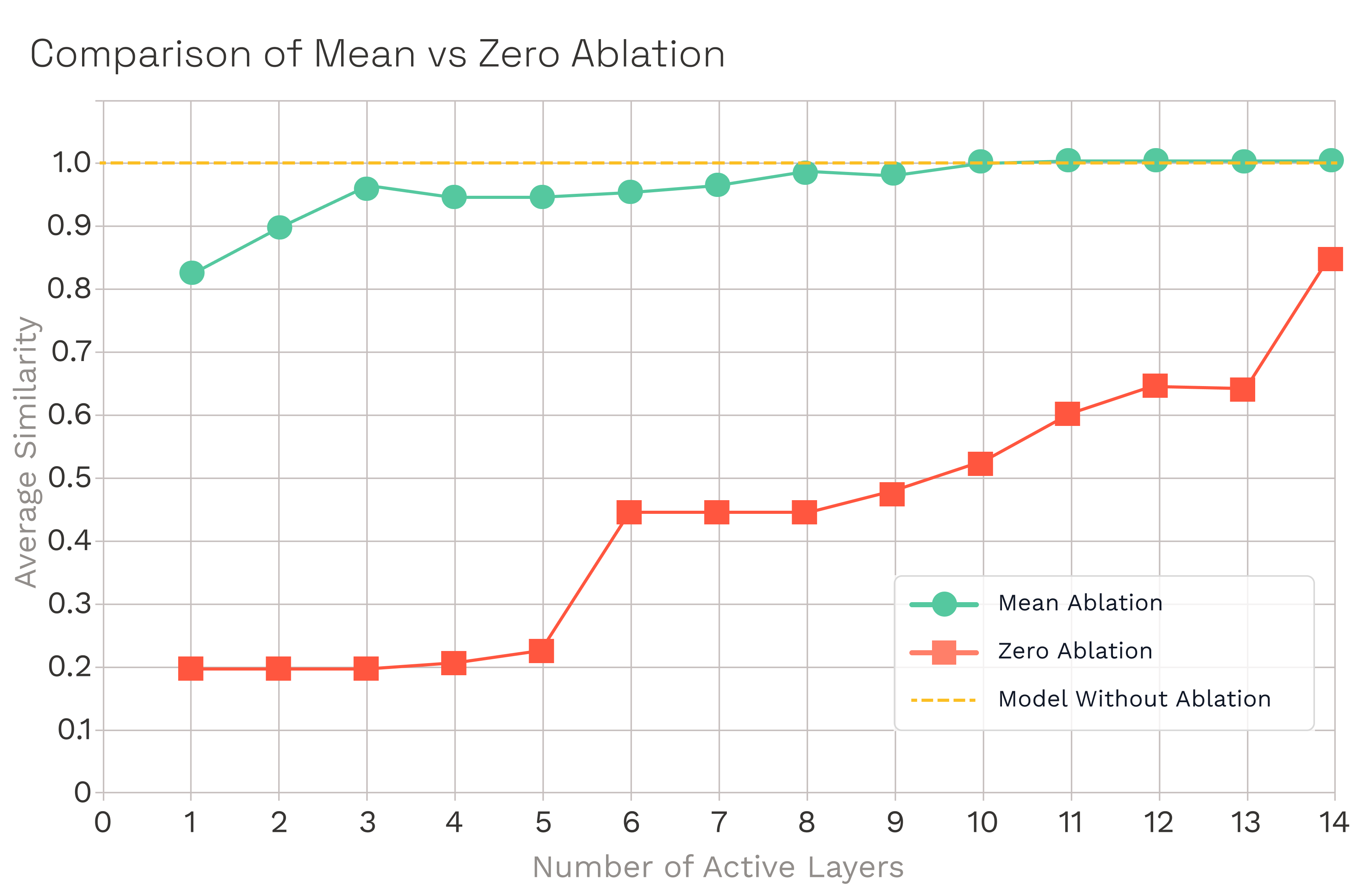}
    \end{minipage}

    \caption{Ablation results showing accuracy trends as more attention heads remain unablated. 
    (Top-left) BM1\_CS1\_Syn with CS1 data, 
    (Top-right) BM1\_CS3\_Syn with CS1 data, 
    (Middle-left) BM1\_CS3\_Syn with CS2 data, 
    (Middle-right) BM1\_CS3\_Syn with CS3 data, 
    (Bottom) BM2\_CS1\_Syn with CS1 data.}
    \label{fig:ablation_results}
\end{figure*}

After selecting the set of attention head nodes using EAP, we ablate all attention heads that are not part of this set. We apply both zero ablation and mean ablation, then compare the results, as shown in \hyperref[fig:ablation_results]{Fig.~\ref{fig:ablation_results}}. Our conclusions are presented in Sec.~\ref{sec:minimal_circuits}.

\subsection{Multi-Layer Perceptron (MLP) Ablations}

The next step was to keep the attention heads intact and instead ablate MLP outputs to determine whether mean ablation produces the same results and whether the model maintains its performance when using average activations instead of actual output activations. To compute the average activations, we use a batch of 200 examples.

\begin{figure}
    \centering
    \begin{minipage}{0.45\textwidth}
        \centering
        \includegraphics[width=\textwidth]{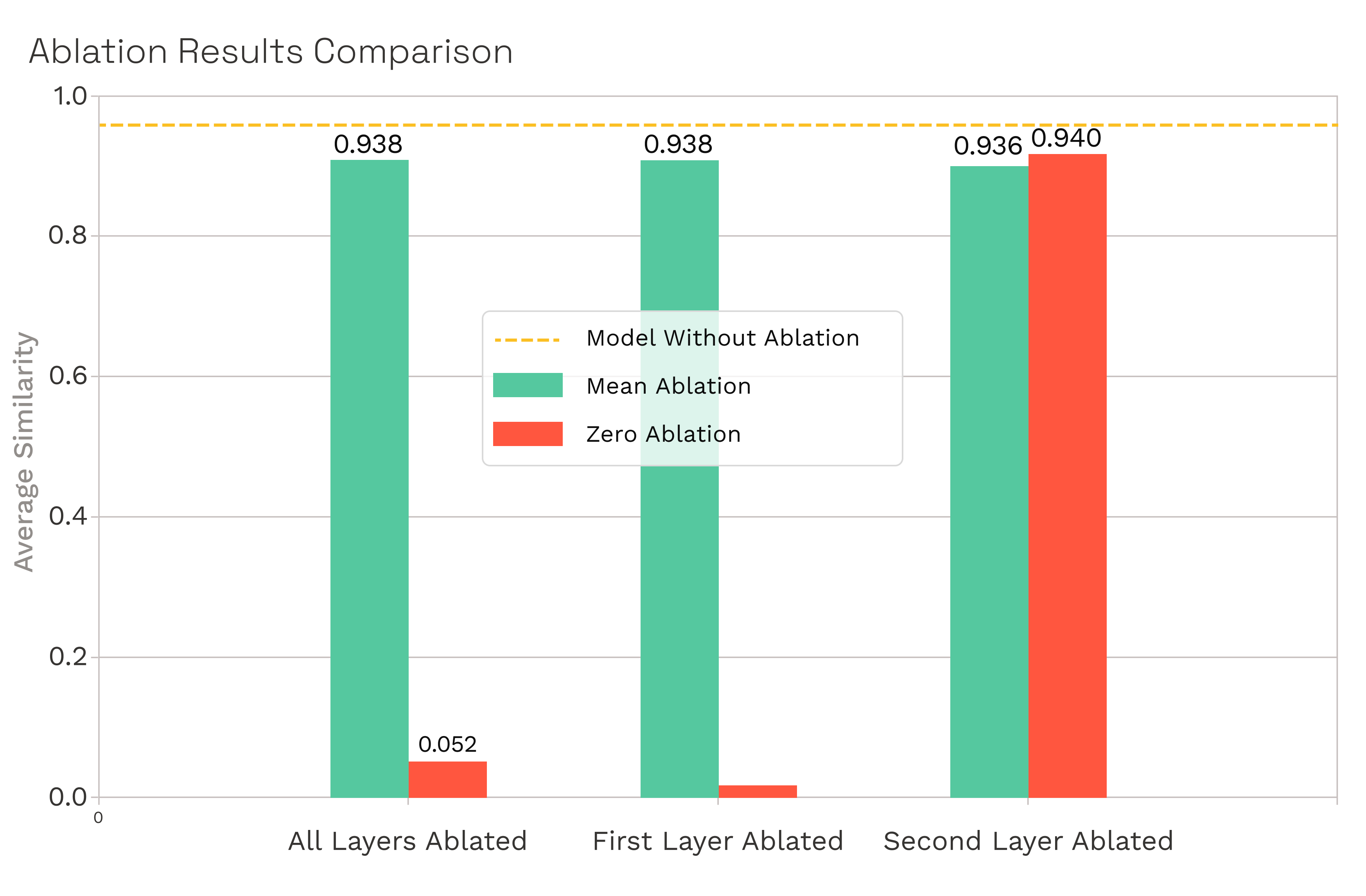}  
        \subcaption{BM1\_CS1 with CS1 data}
        \label{fig:bm1_cs1_cs1_mlp}
    \end{minipage}
    \begin{minipage}{0.45\textwidth}
        \centering
        \includegraphics[width=\textwidth]{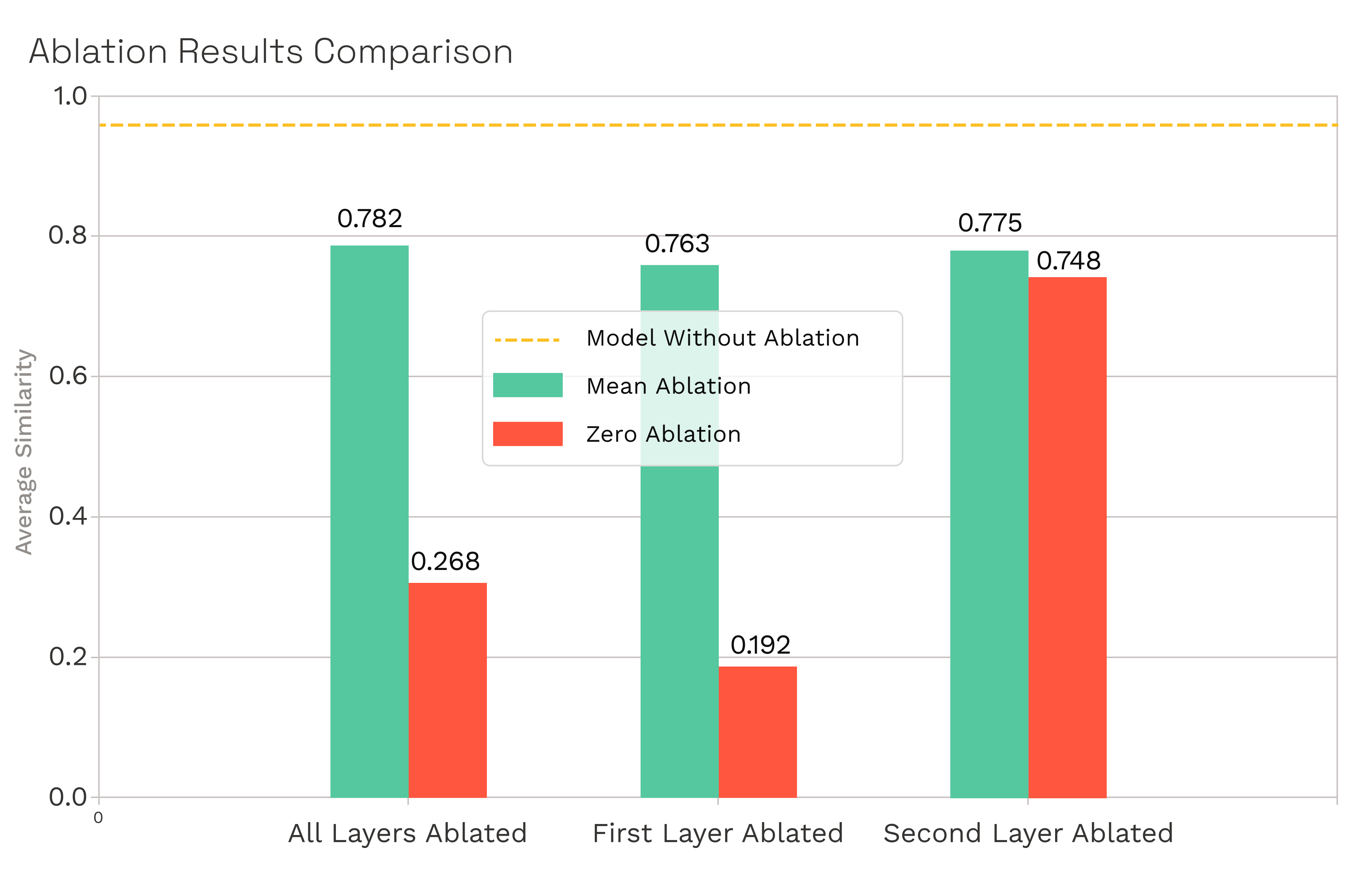}  
        \subcaption{BM1\_CS3 with CS1 data}
        \label{fig:bm1_cs3_cs1_mlp}
    \end{minipage}
    
    \vspace{0.5cm} 

    \begin{minipage}{0.45\textwidth}
        \centering
        \includegraphics[width=\textwidth]{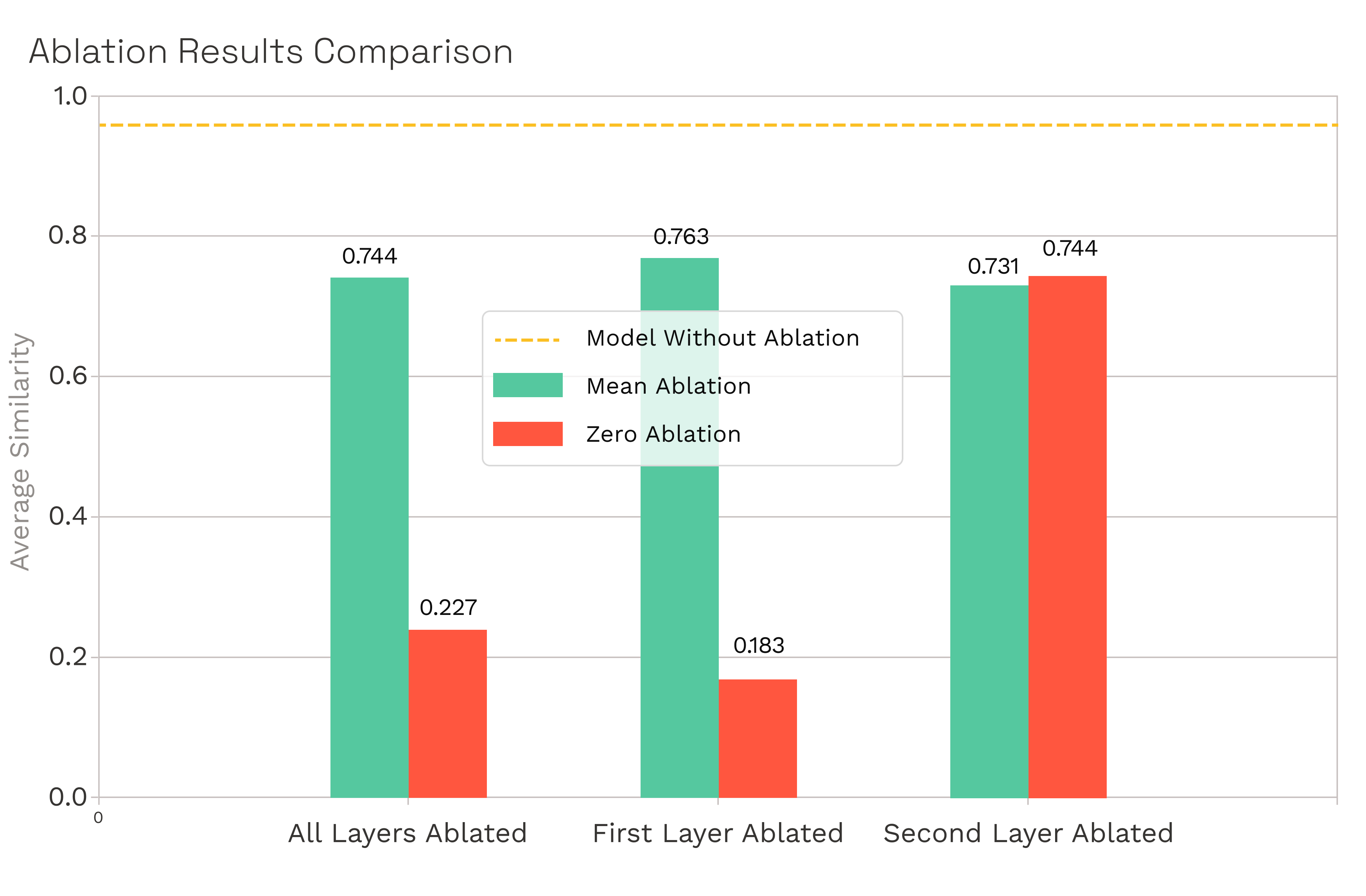}  
        \subcaption{BM1\_CS3 with CS2 data}
        \label{fig:bm1_cs3_cs2_mlp}
    \end{minipage}
    \begin{minipage}{0.45\textwidth}
        \centering
        \includegraphics[width=\textwidth]{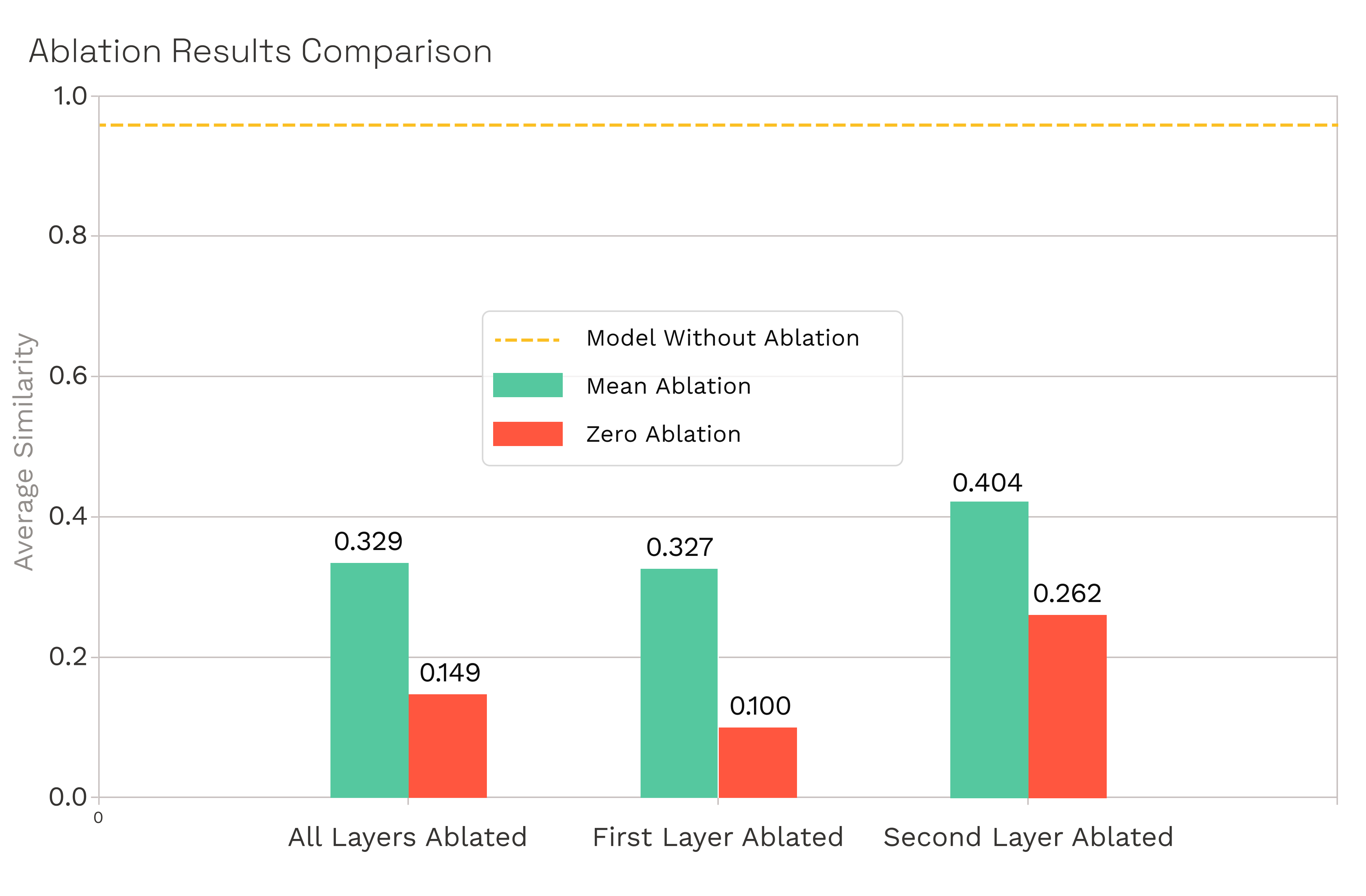}  
        \subcaption{BM1\_CS3 with CS3 data}
        \label{fig:bm1_cs3_cs3_mlp}
    \end{minipage}

    \vspace{0.5cm} 

    \begin{minipage}{0.45\textwidth}
        \centering
        \includegraphics[width=\textwidth]{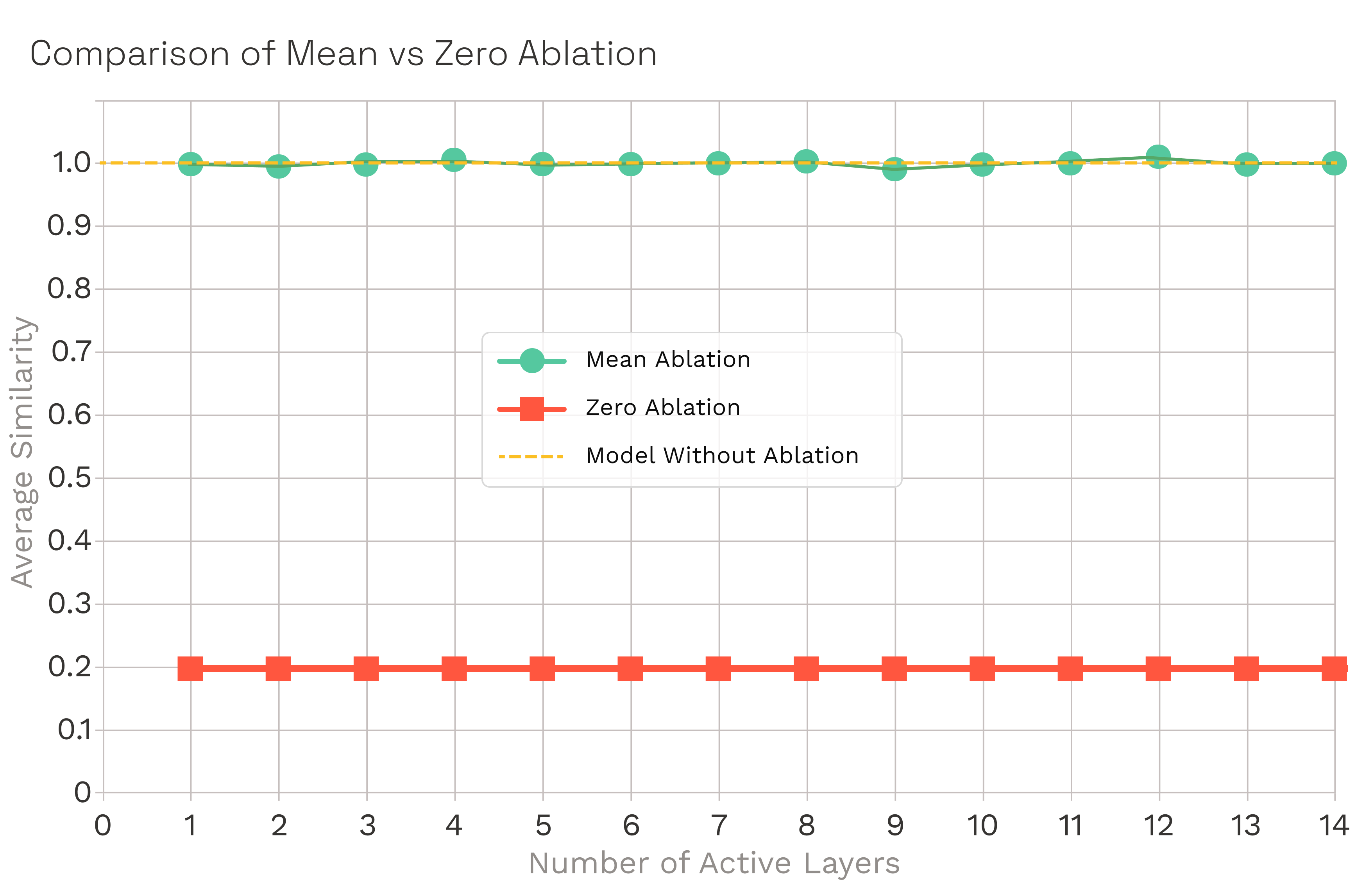} 
        \subcaption{BM2\_CS1 data}
        \label{fig:bm2_cs1_mlp}
    \end{minipage}

    \caption{MLP ablation results across different models and datasets. Each subfigure presents accuracy changes when ablating MLP outputs for specific configurations.}
    \label{fig:mlp_ablation_results}
\end{figure}

Our conclusions were presented in Sec.~\ref{sec:minimal_circuits}. However, the figures in \hyperref[fig:all_ablation_results]{Fig.~\ref{fig:mlp_ablation_results}} show that the first layer plays a major role in all models. Additionally, when a complex dataset is used (CS3), both layers become important, and mean ablation no longer preserves performance. This trend is consistently observed across all cases of MLP ablations.

\subsection{Full-Layer Ablations}

\begin{figure}
    \centering
    \begin{minipage}{0.45\textwidth}
        \centering
        \includegraphics[width=\textwidth]{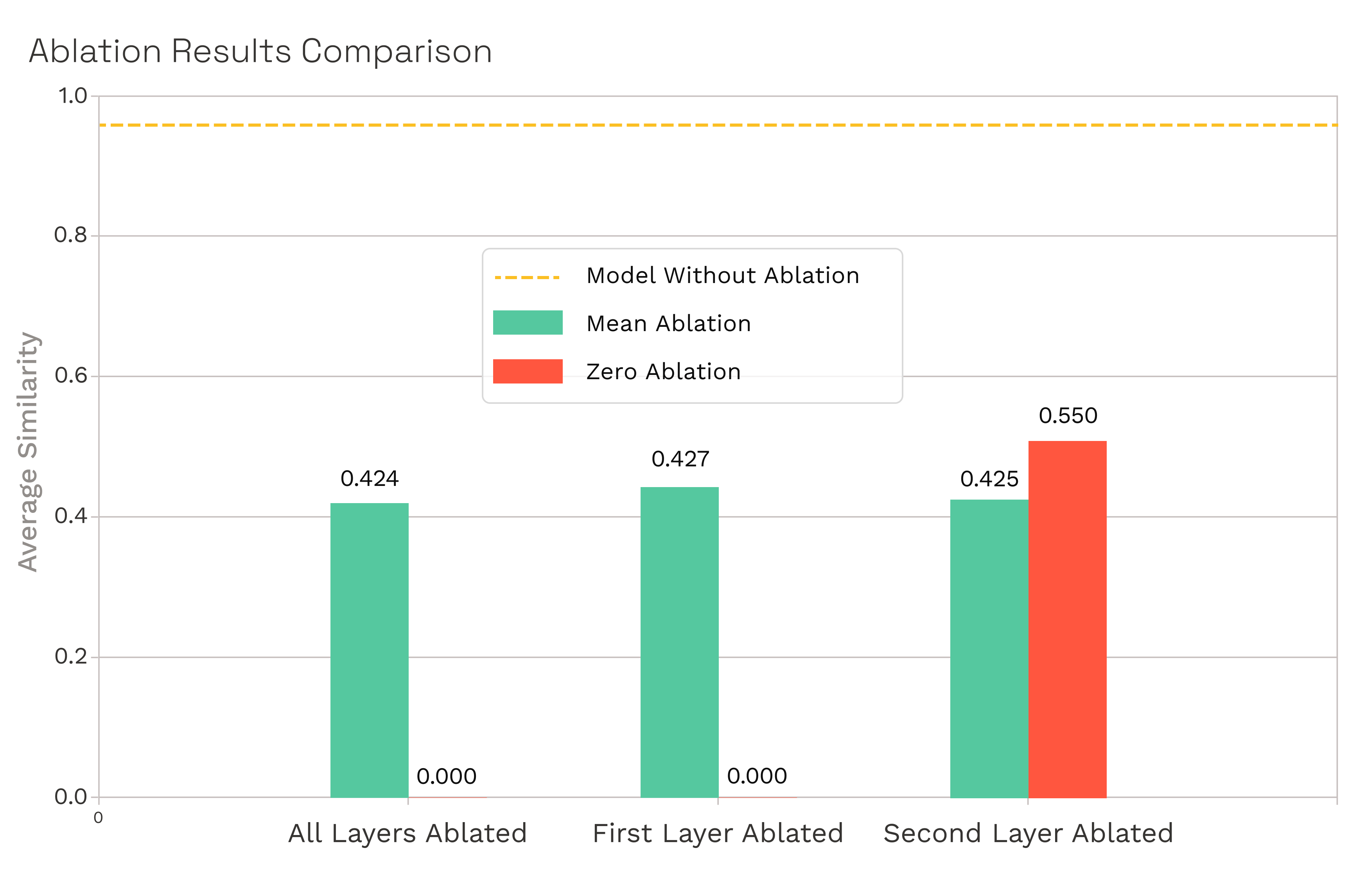}  
        \subcaption{BM1\_CS1 with CS1 data}
        \label{fig:bm1_cs1_cs1_all}
    \end{minipage}
    \begin{minipage}{0.45\textwidth}
        \centering
        \includegraphics[width=\textwidth]{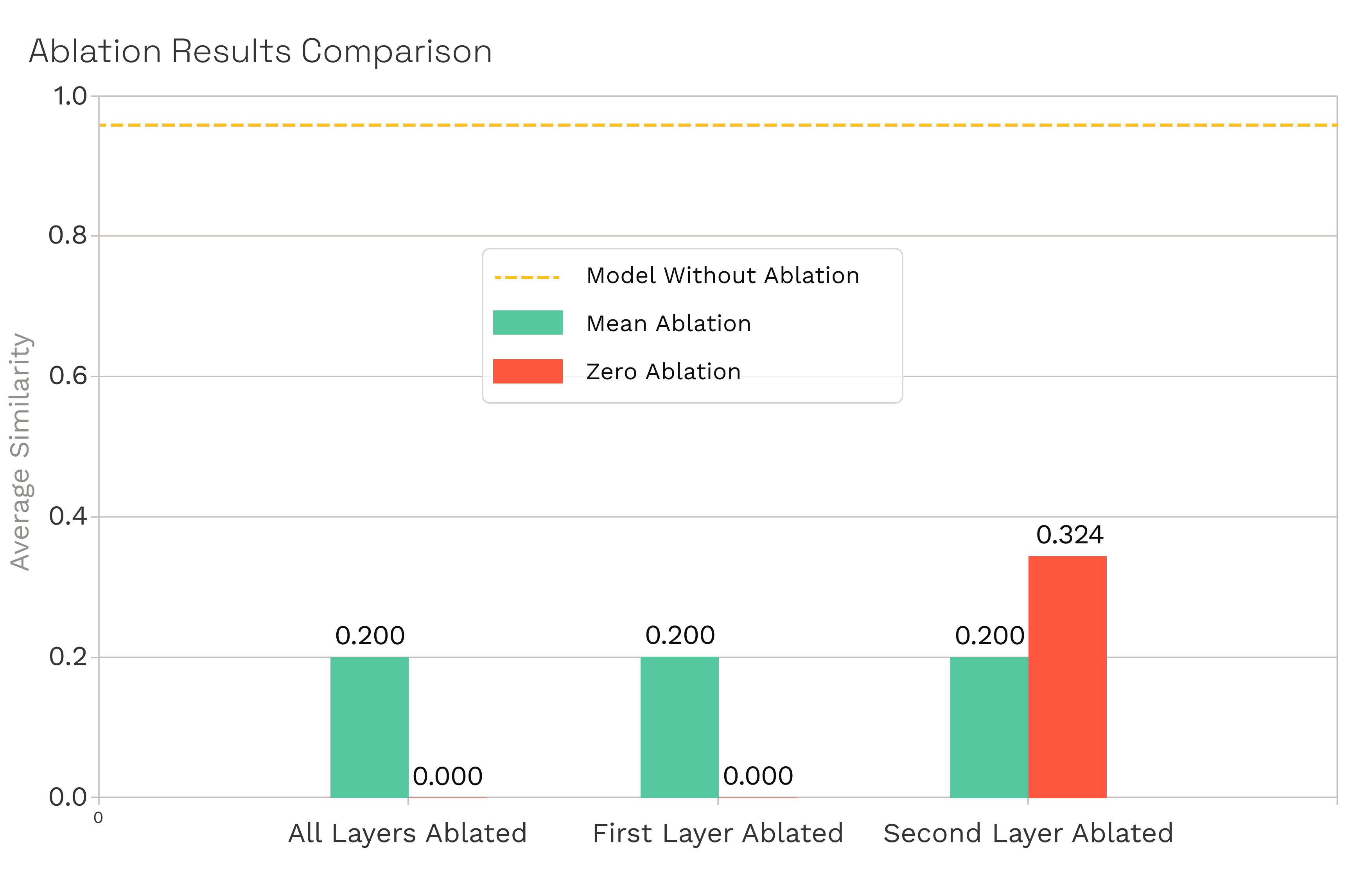}
        
        \subcaption{BM1\_CS3 with CS1 data}
        \label{fig:bm1_cs3_cs1_all}
    \end{minipage}
    
    \vspace{0.5cm} 

    \begin{minipage}{0.45\textwidth}
        \centering
        \includegraphics[width=\textwidth]{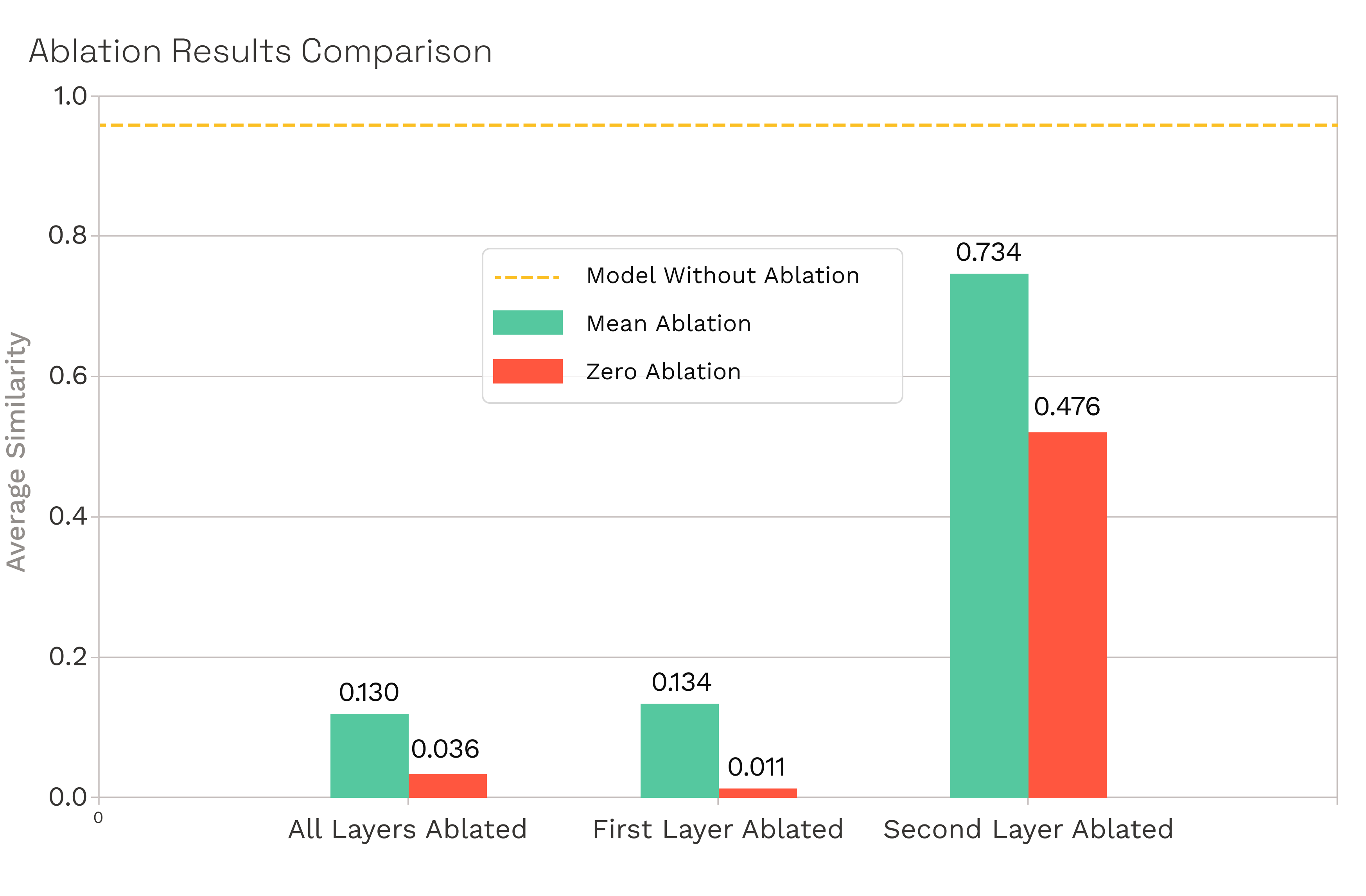}  
        \subcaption{BM1\_CS3 with CS2 data}
        \label{fig:bm1_cs3_cs2_all}
    \end{minipage}
    \begin{minipage}{0.45\textwidth}
        \centering
        \includegraphics[width=\textwidth]{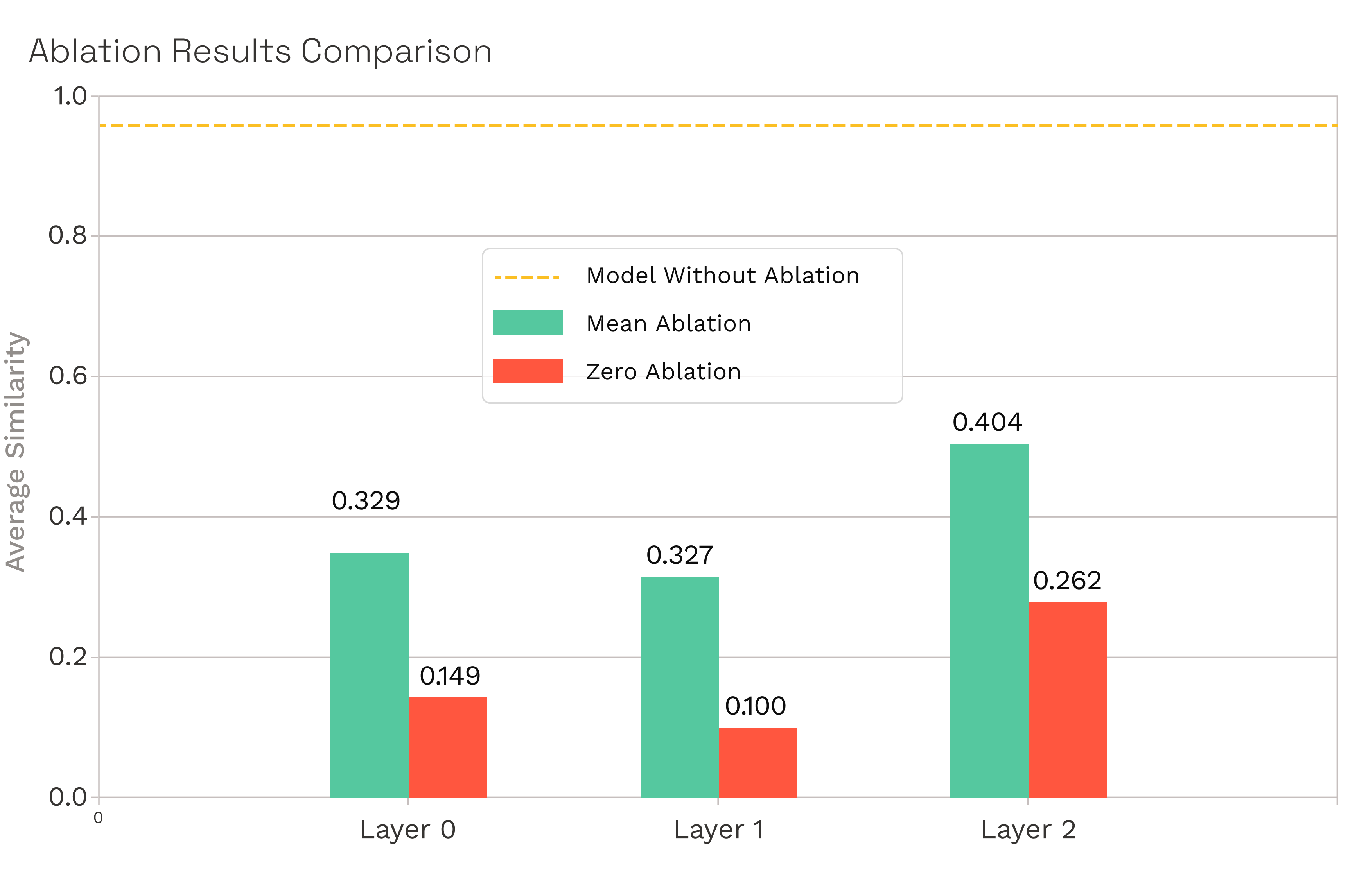}  
        \subcaption{BM1\_CS3 with CS3 data}
        \label{fig:bm1_cs3_cs3_all}
    \end{minipage}

    \vspace{0.5cm} 

    \begin{minipage}{0.45\textwidth}
        \centering
        \includegraphics[width=\textwidth]{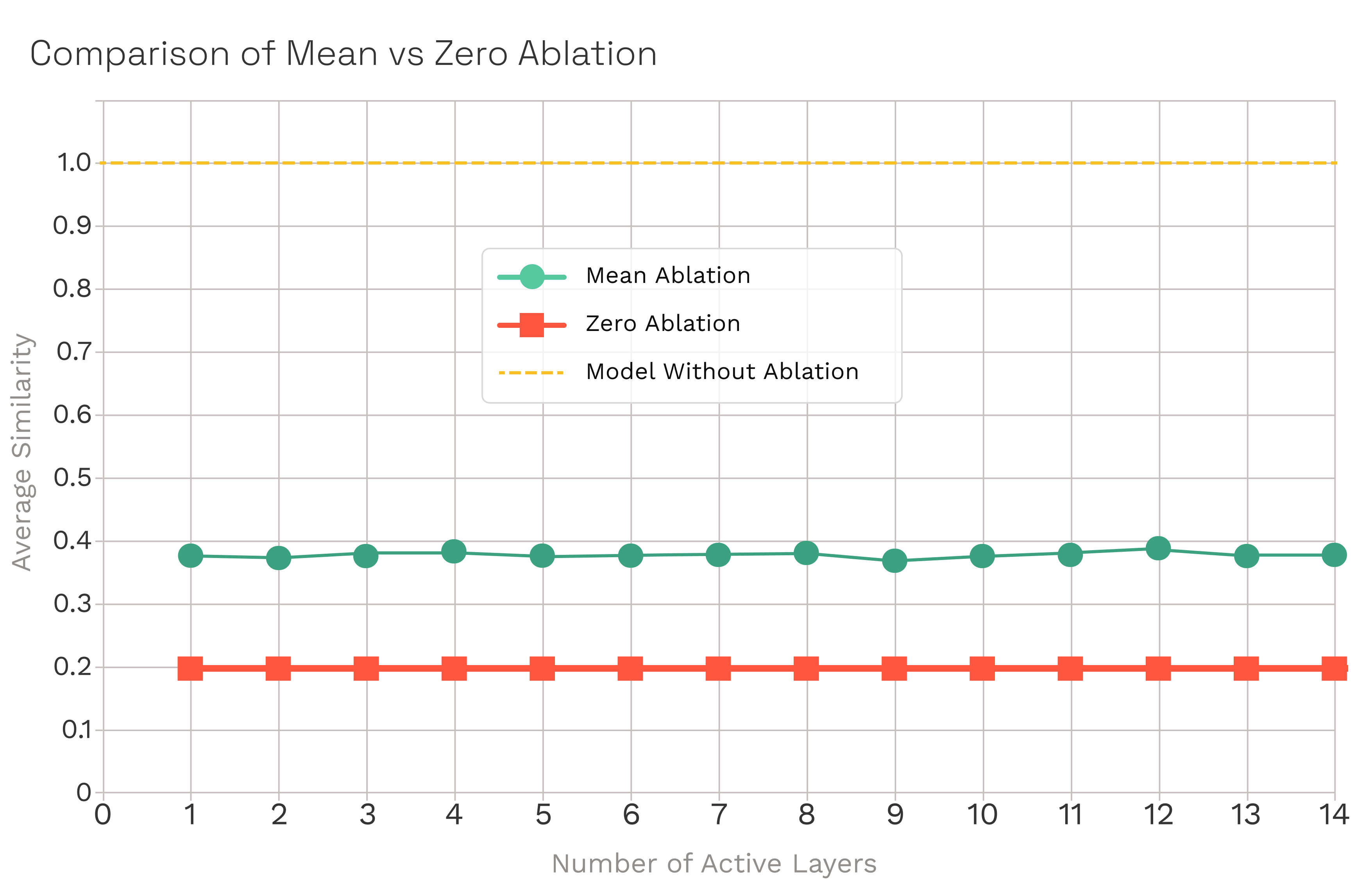} 
        \subcaption{BM2\_CS1 data}
        \label{fig:bm2_cs1_all}
    \end{minipage}

    \caption{Results of ablation across all outputs for different models and datasets. Each subfigure illustrates the impact of ablating all outputs in the specified configuration.}
    \label{fig:all_ablation_results}
\end{figure}

In the same manner, we now ablate all the outputs of a layer. Once again, we observe as shown in \hyperref[fig:all_ablation_results]{Fig.~\ref{fig:all_ablation_results}}  that layer 1 plays a more significant role in generation. However, this trend shifts with more complex command sets. When ablating entire layers, the model loses performance under both mean ablation and zero ablation across all layers.

For larger models like Qwen, accuracy drops to 20\% regardless of the number of layers ablated.

\section{SAE analysis}\label{app:saes}
We show the important heads in BM1 for processing table tokens next in Table \ref{tab:sae_important_heads_table_name_process}.
We also consider the important SAE features for generating the field tokens used for "ORDER BY" or "GROUP BY" type queries in 
Table \ref{tab:sae_important_heads_by_fields}.

\begin{table}
    \centering
    \resizebox{\columnwidth}{!}{
    \begin{tabular}{l|l|c|c}
        \toprule
        \textbf{Model} & \textbf{Dataset} & \textbf{Layer 0} & \textbf{Layer 1} \\
        \midrule
        BM1\_CS1\_Syn & CS1 & [1, 8, 10, 7] & [13, 4, 2, 1] \\
        BM1\_CS1\_Syn & CS1\_Syn & [1, 8, 10, 7] & [13, 4, 2, 8] \\
        \midrule
        BM1\_CS2\_Syn & CS2 & [1, 10, 8, 14] & [13, 8, 4, 1] \\
        BM1\_CS2\_Syn & CS2\_Syn & [1, 10, 8, 14] & [13, 8, 4, 1] \\
        \midrule
        BM1\_CS3\_Syn & CS3 & [1, 10, 8, 14] & [6, 2, 1, 9] \\
        BM1\_CS3\_Syn & CS3\_Syn & [1, 10, 8, 14] & [6, 2, 1, 9]\\
        \bottomrule
    \end{tabular}}
    \caption{From SAE feature analysis, important attention heads for processing table name tokens.}
    \label{tab:sae_important_heads_table_name_process}
\end{table}

Note that for response table names, Head 1 in Layer 0 is consistently the most important. Whereas we see in Layer 1, that BM1\_CS3\_* models consistently use Head 6 as opposed to BM1 and BM2 which use Layer 13.

We next look at important heads for generating table names, as determined by being immediately before the response table name, in Table \ref{tab:sae_important_heads_table_name_generate}.

\begin{table}
    \centering
    \resizebox{\columnwidth}{!}{
    \begin{tabular}{l|l|c|c}
        \toprule
        \textbf{Model} & \textbf{Dataset} & \textbf{Layer 0} & \textbf{Layer 1} \\
        \midrule
        BM1\_CS1\_Syn & CS1 & [1, 10, 8, 7] & [13, 4, 1, 12] \\
        BM1\_CS1\_Syn & CS1\_Syn & [1, 10, 8, 7] & [13, 4, 12, 1] \\
        \midrule
        BM1\_CS2\_Syn & CS2 & [1, 10, 8, 14] & [13, 8, 4, 1] \\
        BM1\_CS2\_Syn & CS2\_Syn & [1, 10, 8, 14] & [13, 8, 4, 1] \\
        \midrule
        BM1\_CS3\_Syn & CS3 & [1, 10, 8, 14] & [6, 2, 1, 13] \\
        BM1\_CS3\_Syn & CS3\_Syn & [1, 10, 8 , 14] & [6, 2, 1, 9]\\
        \bottomrule
    \end{tabular}}
    \caption{From SAE feature analysis, important attention heads for generating table name tokens.}
    \label{tab:sae_important_heads_table_name_generate}
\end{table}

\begin{table}
    \centering
    \resizebox{\columnwidth}{!}{
    \begin{tabular}{l|l|c|c}
        \toprule
        \textbf{Model} & \textbf{Dataset} & \textbf{Layer 0} & \textbf{Layer 1} \\
        \midrule
        BM1\_CS2\_Syn & CS2 & [1, 10, 8, 14]  & [13, 8, 1, 4]\\
        BM1\_CS2\_Syn & CS2\_Syn & [1, 10, 8, 14] & [13, 8, 4, 1] \\
        \midrule
        BM1\_CS3\_Syn & CS3 & [1, 10, 8, 14] & [6 ,2 ,1 , 13] \\
        BM1\_CS3\_Syn & CS3\_Syn & [1, 10, 8, 14] & [6, 2, 1, 13] \\
        \bottomrule
    \end{tabular}}
    \caption{From SAE feature analysis, important attention heads for generating order by and sort by tokens accurately.}
    \label{tab:sae_important_heads_by_fields}
\end{table}

The high degree of overlap between the heads used for processing the table names, and the field tokens, suggests that either these heads are polysemantic, or that the model uses a similar mechanism for selecting parts of the prompt to be used in the final response.

\begin{figure}
    \centering
    \includegraphics[width=0.48\textwidth]{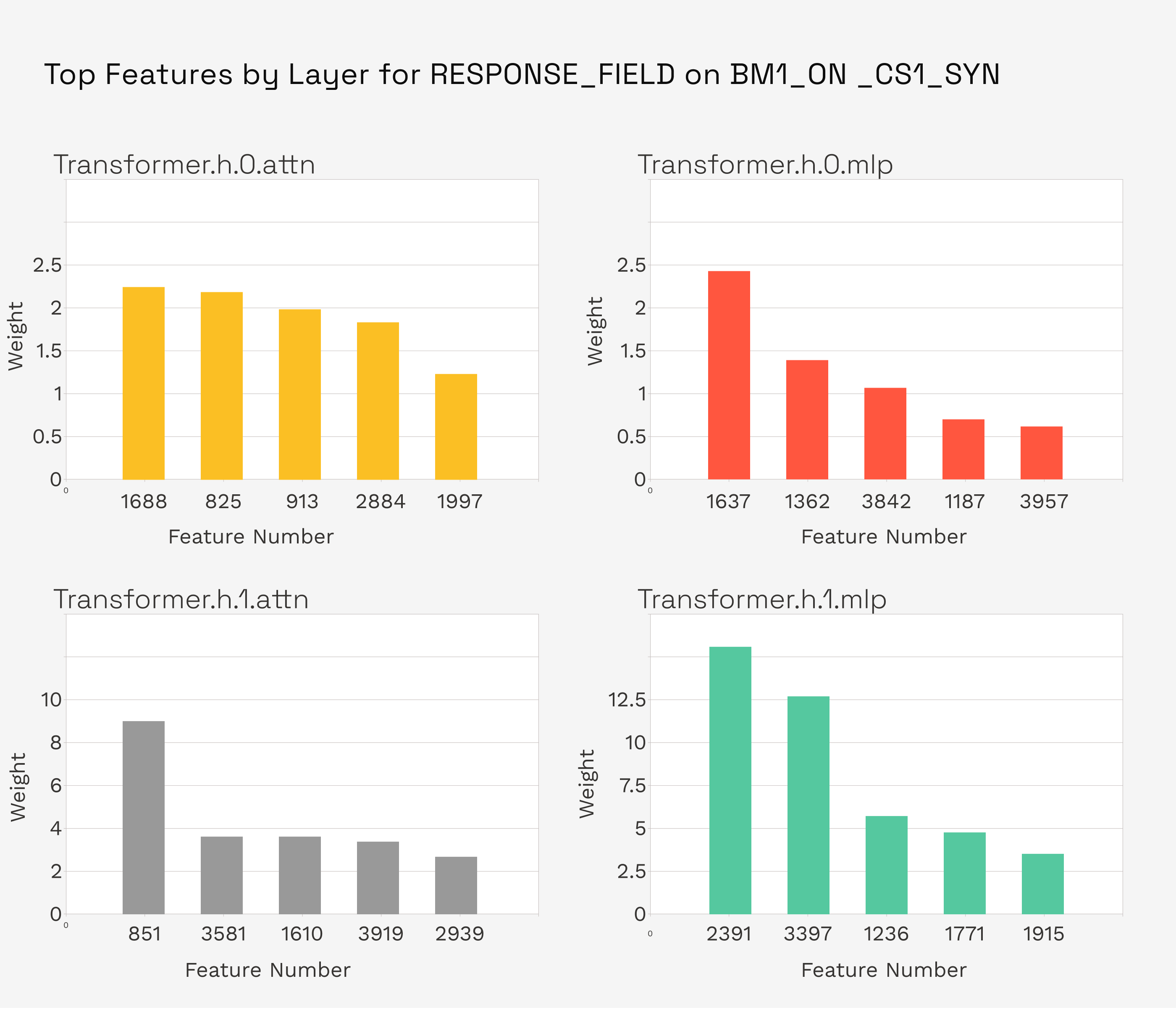} 
    \caption{Average SAE activations for BM1\_CS1\_Syn on CS1\_Syn table fields.}
    \label{fig:bm1_cs1_sae_table_fields}
\end{figure}

\begin{figure}
    \centering
    \includegraphics[width=0.48\textwidth]{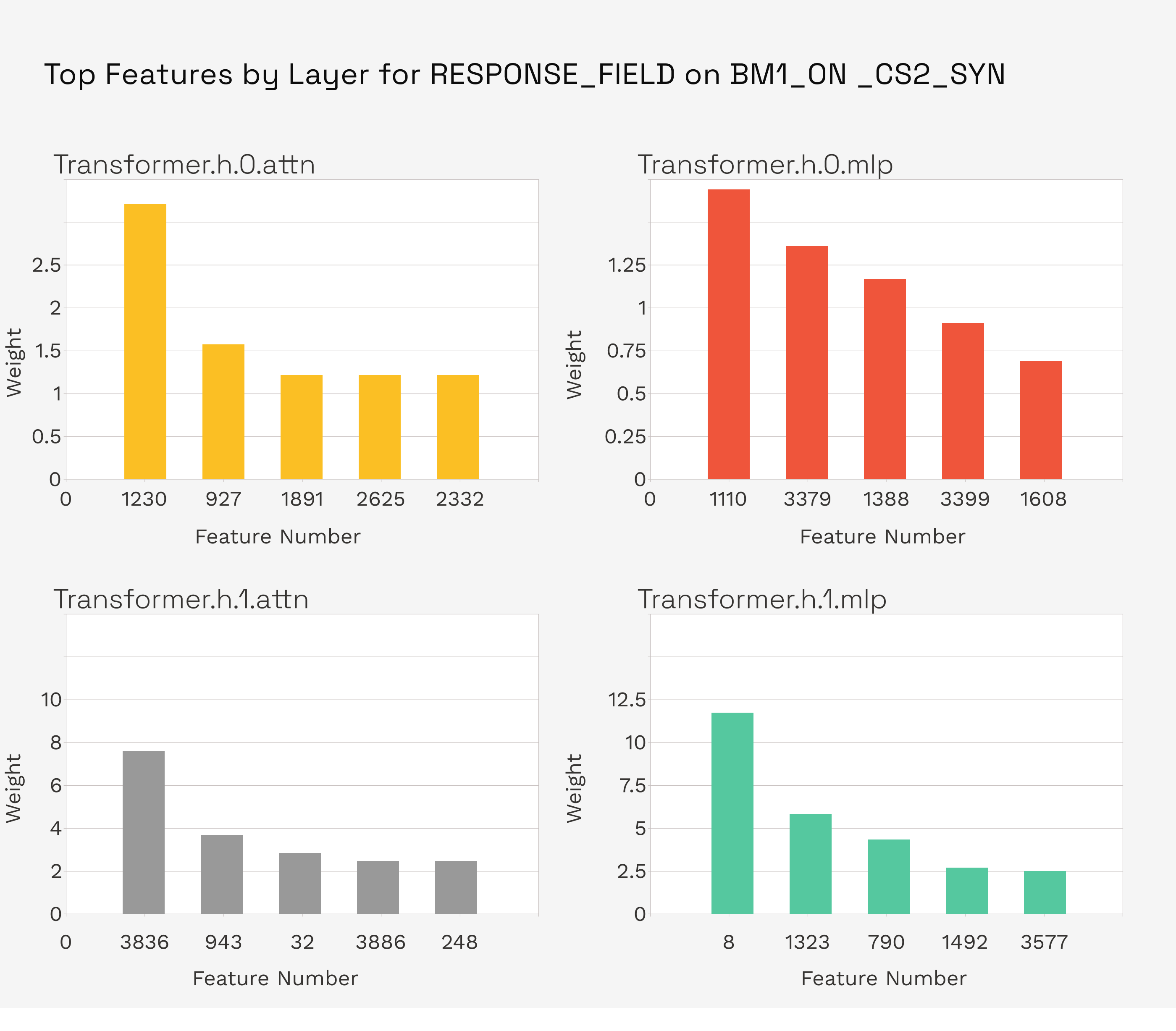} 
    \caption{Average SAE activations for BM1\_CS2\_Syn on CS2\_Syn table fields.}
    \label{fig:bm1_cs2_sae_table_fields}
\end{figure}

\begin{figure}
    \centering
    \includegraphics[width=0.48\textwidth]{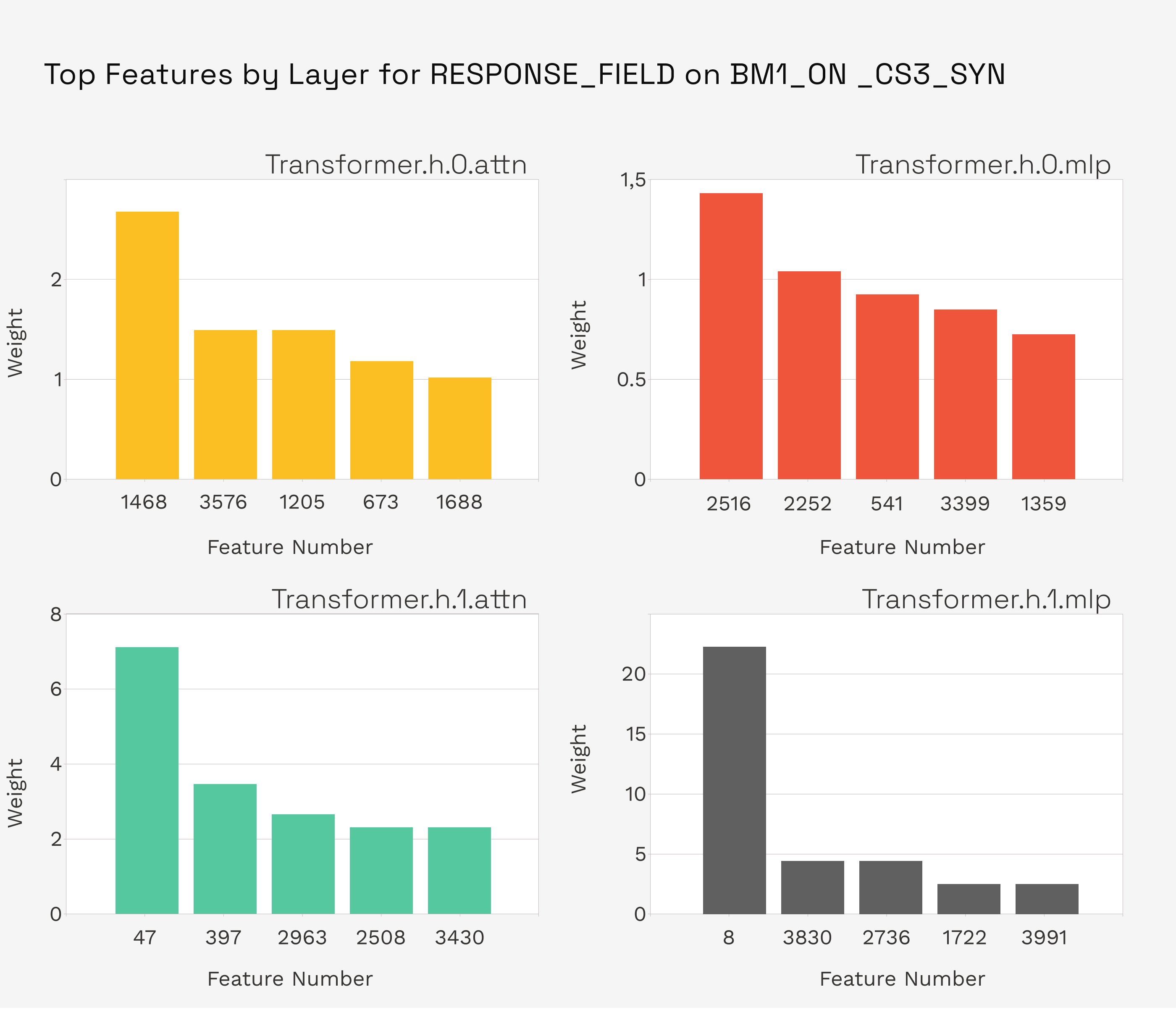} 
    \caption{Average SAE activations for BM1\_CS3\_Syn on CS3\_Syn table fields.}
    \label{fig:bm1_cs3_sae_table_fields}
\end{figure}

We show now the SAE features most involved in filling out the table name tokens, in particular for BM1\_CS1\_Syn, BM1\_CS2\_Syn and BM1\_CS3\_Syn.
 See Figures \ref{fig:bm1_cs1_sae_table_fields}, 
\ref{fig:bm1_cs2_sae_table_fields} and \ref{fig:bm1_cs3_sae_table_fields} for the top activating SAE features by average magnitude.

\begin{figure}
\centering
\begin{minipage}{0.48\textwidth}
    \centering
    \includegraphics[width=\textwidth]{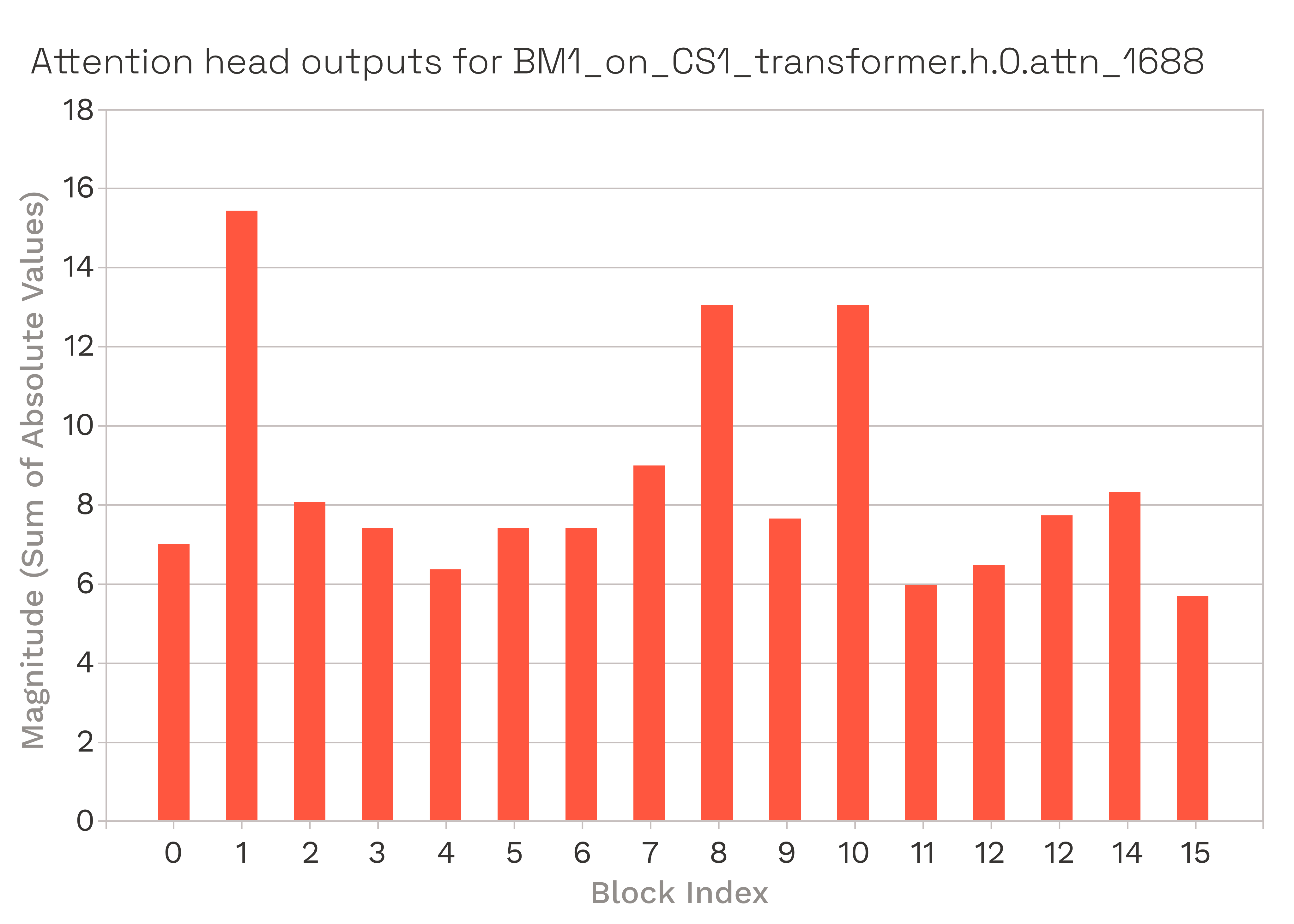}
    \subcaption{BM1\_CS1 Attention outputs by head for SAE feature 1688 on Layer 0.}
    \label{fig:bm1_cs1_sae_attn}
\end{minipage}
\hfill
\begin{minipage}{0.48\textwidth}
    \centering
    \includegraphics[width=\textwidth]{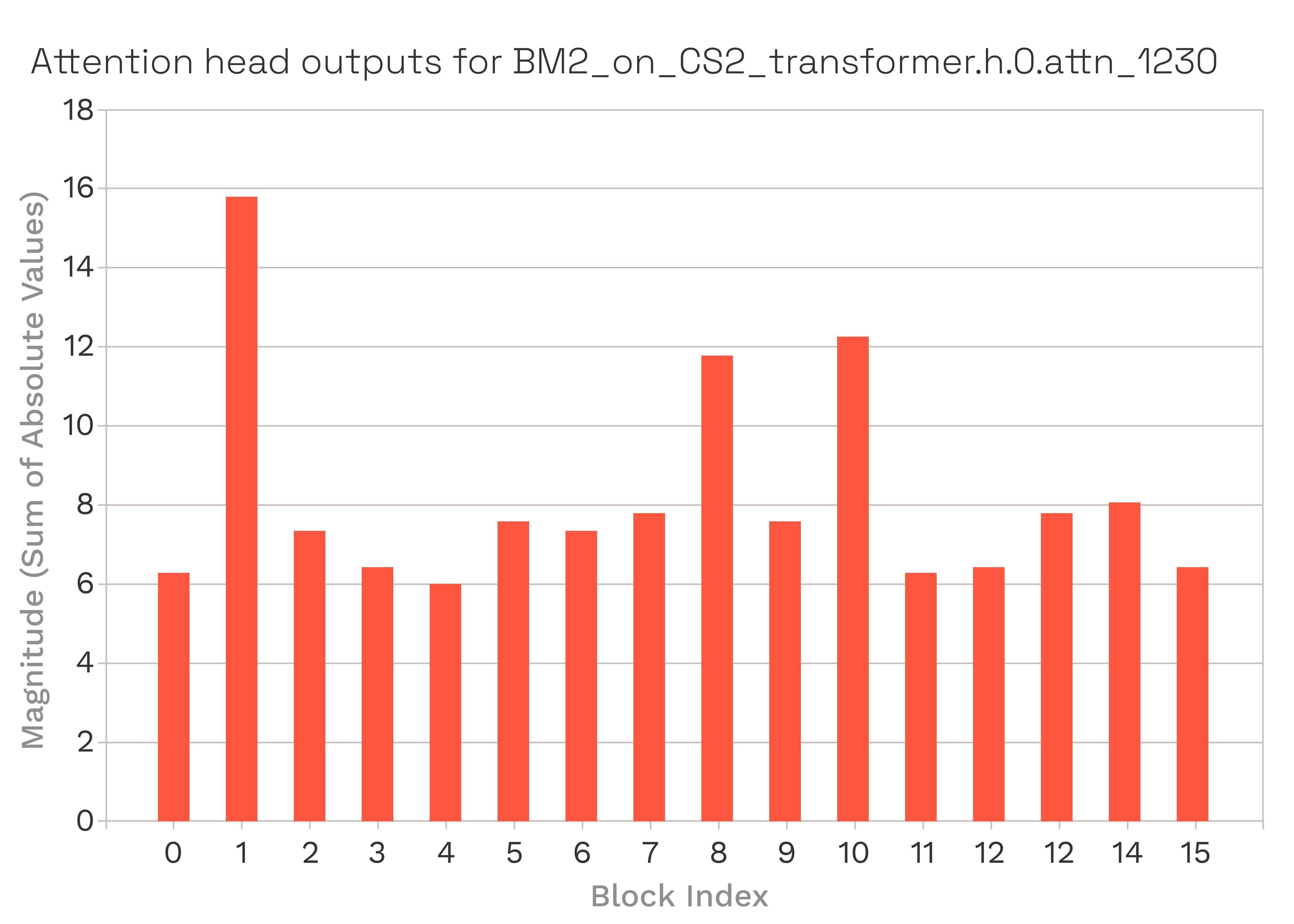}
    \subcaption{BM1\_CS2 Attention outputs by head for SAE feature 1230 on Layer 0.}
    \label{fig:bm1_cs2_sae_attn}
\end{minipage}

\vspace{1em}
\begin{minipage}{0.48\textwidth}
    \centering
    \includegraphics[width=\textwidth]{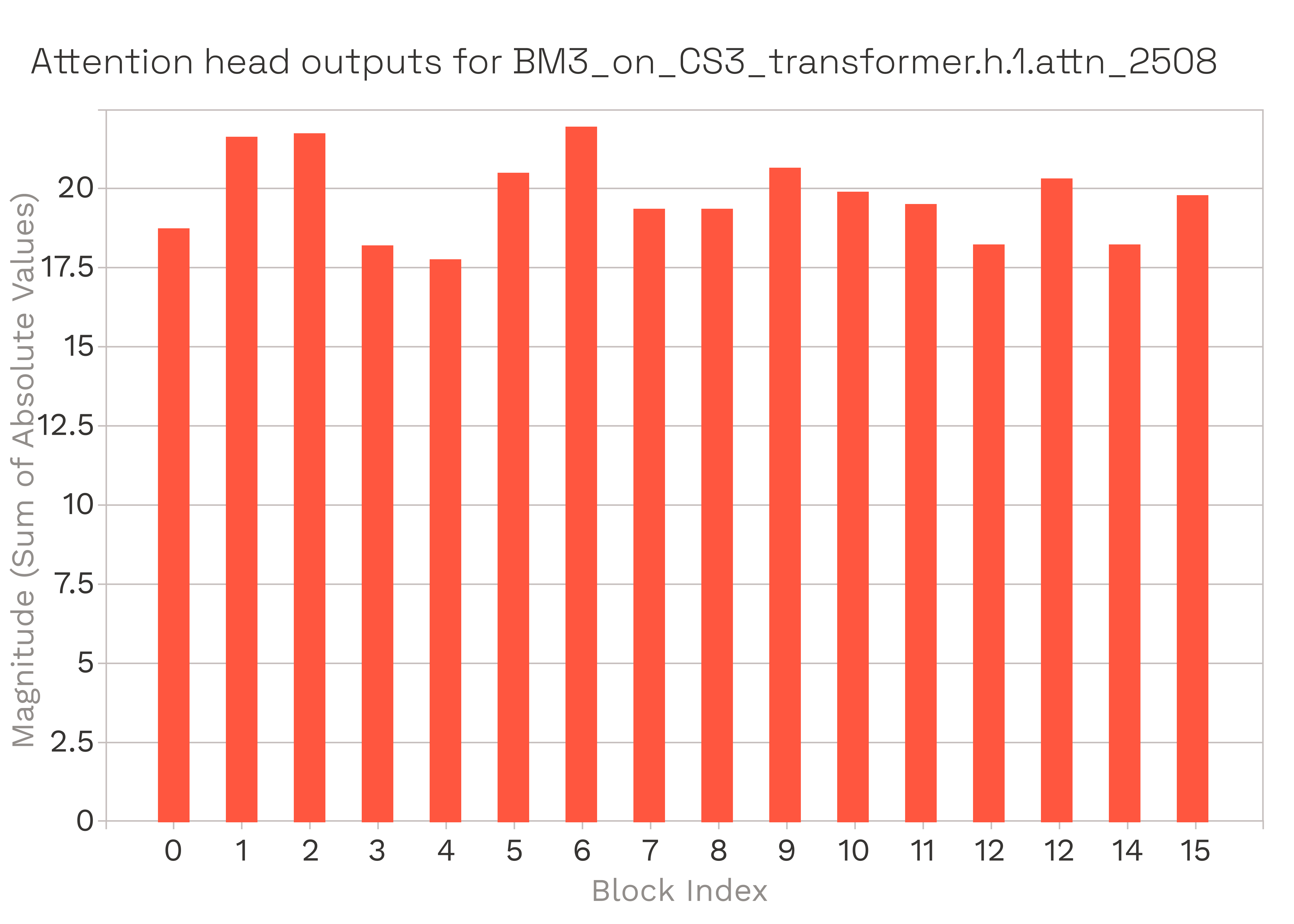}
    \subcaption{BM3\_CS3 Attention outputs by head for SAE feature 2508 on Layer 1.}
    \label{fig:bm1_cs3_sae_attn}
\end{minipage}
\caption{Attention outputs by head for selected SAE features across models and layers. Top row: BM1 with two different contexts; Bottom: BM3 highlighting deeper-layer behavior.}
\label{fig:sae_attn_outputs}
\end{figure}

We can map these to the actual attention outputs, see for instance Figures 
\ref{fig:bm1_cs1_sae_attn}, \ref{fig:bm1_cs2_sae_attn} and \ref{fig:bm1_cs3_sae_attn}. We average over these to determine the influential attention heads.

\section{SAE Circuits Results}
\label{app:saes_circuits_res}

We present additional results from our Synthetic Activation Experiments (SAEs) highlighting the identified circuit components and their recovered accuracy after zero-ablation across different model and dataset configurations (Table \ref{tab:bm1_extra_sae}).

\begin{table*}[t]
\centering
\small
\begin{tabularx}{\textwidth}{l|l|X|X|c}
\toprule
\textbf{Model} & \textbf{Dataset} & \multicolumn{2}{c|}{\textbf{Heads and MLPs}} & \textbf{Recovered Accuracy} \\
\cmidrule(r){3-4}
 & & \textbf{Layer 0} & \textbf{Layer 1} &  \\
\midrule
\multirow{2}{*}{BM1\_CS3\_Syn} & \multirow{2}{*}{CS2} & 13, 1, 8, 15, 10, 12, 14, 3, 7 & -- & \multirow{2}{*}{81.41\%} \\
 & & All MLP neurons & All MLP neurons & \\
\midrule
\multirow{2}{*}{BM1\_CS4\_Syn} & \multirow{2}{*}{CS4} & 8, 1, 10, 13, 3, 6, 12, 9, 2, 15 & -- & \multirow{2}{*}{82.19\%} \\
 & & 750 MLP neurons & -- & \\
\midrule
\multirow{2}{*}{BM1\_CS5\_Syn} & \multirow{2}{*}{CS5} & 8, 10, 1, 13, 3, 12, 6, 9, 2, 7, 4 & -- & \multirow{2}{*}{70.00\%} \\
 & & All MLP neurons & -- & \\
\bottomrule
\end{tabularx}
\caption{Additional SAE-based circuit results for BM1-CS3→CS2, CS4, and CS5 configurations. Recovered accuracy reflects performance after zero-ablation of all model components not selected by high-AUC SAE features.}
\label{tab:bm1_extra_sae}
\end{table*}

Table.\ref{tab:bm2_sae} details the layer-wise circuits identified by Synthetic Activation Experiments (SAEs) for the \texttt{BM2\_CS1\_Syn} model, demonstrating more granular selection of individual attention heads compared to the broader patterns found by EAP.

\begin{table*}
\centering

\begin{tabular}{llcc}
\toprule
\textbf{Model} & \textbf{Layer} & \textbf{Selected Heads} & \textbf{Recovered Accuracy} \\
\midrule
\multirow{24}{*}{BM2\_CS1\_Syn} 
  & L0  & 12, 7, 6, 0, 10, 2, 1, 9, 4           &        \\
  & L1  & All Heads                            &        \\
  & L2  & 0, 11, 4, 5, 9, 8, 13, 1, 2           &        \\
  & L3  & 0, 11, 10, 12, 3, 1, 9, 4             &        \\
  & L4  & All Heads                            &        \\
  & L5  & All Heads                            &        \\
  & L6  & 5, 0, 2, 13, 6, 8, 4, 12, 7, 9        &        \\
  & L7  & 9, 1, 6, 0, 2, 4, 11, 8, 3            &        \\
  & L8  & All Heads                            &        \\
  & L9  & 0, 3, 8, 4, 13, 9, 7, 10, 1           &        \\
  & L10 & All Heads                            &        \\
  & L11 & All Heads                            & 88.1\% \\
  & L12 & 0, 8, 7, 11, 3, 1, 5, 13             &        \\
  & L13 & 0, 3, 11, 7, 1, 2, 4, 10, 9           &        \\
  & L14 & All Heads                            &        \\
  & L15 & 9, 7, 10, 1, 4, 3, 8, 13, 5, 0        &        \\
  & L16 & All Heads                            &        \\
  & L17 & 13, 5, 4, 10, 7, 0, 8, 11             &        \\
  & L18 & 3, 0, 5, 13, 11, 8, 4, 9, 1, 6        &        \\
  & L19 & 3, 0, 7, 4, 11, 10, 13, 1, 5, 12      &        \\
  & L20 & 0, 3, 7, 2, 1, 13, 10                &        \\
  & L21 & 0, 3, 7, 1, 10, 6, 11, 9             &        \\
  & L22 & 0, 3, 7, 1, 2, 11, 10, 6             &        \\
  & L23 & 12, 1, 9, 13, 5, 2, 11               &        \\
\bottomrule
\end{tabular}
\caption{Layer-wise SAE-selected circuits for \textit{BM2\_CS1\_Syn}. Compared to EAP, the circuits identified by SAEs exhibit finer granularity, isolating individual attention heads rather than entire layers.}
\label{tab:bm2_sae}
\end{table*}

\section{Mean-Ablation Circuits}
\label{appendix:mean_circuits}

\paragraph{Circuits in Small 2-Layer Models.}
We begin by analyzing models with only two transformer layers (e.g., BM1 and BM3 variants). The mean ablation results for these models reveal small, interpretable sets of attention heads responsible for accurate task performance. Table~\ref{tab:small-models-attn} summarizes the critical heads involved in each model-task pair and their corresponding recovered accuracy after ablation.

\begin{table*}
\centering

\begin{tabular}{lcccc}
\toprule
\textbf{Model} & \textbf{Dataset} & \textbf{Important Heads (Layer0)} & \textbf{Recovered Accuracy} & \textbf{Component \%} \\
\midrule
BM1-CS1 & CS1 & 11, 3, 1, 8 & 93.68\% & 12.5\% \\
BM1-CS3 & CS1 & 11, 14, 3, 8, 2 & 93.8\% & 15.62\% \\
BM1-CS3 & CS2 & 11 & 99.0\% & 3.12\% \\
BM1-CS3 & CS3 & 11, 14, 3, 0, 13 & 73.43\% & 15.62\% \\
BM3-CS1 & CS1 & 0, 5 & 99.2\% & 8.3\% \\
\bottomrule
\end{tabular}
\caption{Mean Ablation Results for Attention Circuits in BM1 and BM3 Models}
\label{tab:small-models-attn}
\end{table*}

\paragraph{Circuits in a Deeper Model: BM2-CS1.}
To investigate whether similar sparse circuits exist in deeper models, we perform a layer-wise mean ablation analysis on BM2-CS1, which has 24 layers. For each layer, we identify key heads contributing to recovery and report both the area under the curve (AUC) and the recovered accuracy. Table~\ref{tab:bm2cs1-attn} presents these results, showing how different layers contribute to robust model behavior via distributed yet sparse mechanisms.

\begin{table*}
\centering

\begin{tabular}{lcc}
\toprule
\textbf{Layer} & \textbf{Key Heads} & \textbf{Recovered Accuracy} \\
\midrule
0  & 12 & 98.6\% \\
1  & 7 & 98.4\% \\
2  & 0 & 98.2\% \\
3  & 0, 11, 10, 12, 3, 1, 9, 4, 13 & 98.3\% \\
4  & 2 & 98.3\% \\
5  & 0 & 98.4\% \\
6  & 5 & 97.9\% \\
7  & 9 & 98.2\% \\
8  & 11 & 98.2\% \\
9  & 3 & 98.39\% \\
10 & 0 & 98.2\% \\
11 & 3 & 97.8\% \\
12 & 7 & 98.2\% \\
13 & 0 & 98.2\% \\
14 & 0 & 97.9\% \\
15 & 10, 7, 4, 13, 1, 9, 3, 8 & -- \\
16 & 0, 7, 13, 10, 3, 11, 2, 12, 1 & 97.9\% \\
17 & 13, 0, 5, 4, 1, 3, 10 & 98.2\% \\
18 & 3, 4, 5, 11, 0, 13, 12, 9, 6, 8 & 98.2\% \\
19 & 3, 0, 13, 10, 7, 12, 5, 4, 1 & 97.9\% \\
20 & 0, 3, 7, 2, 6, 4, 10, 1, 8 & 97.9\% \\
21 & 7, 13, 1, 6, 3, 8, 10, 9, 0 & 98.2\% \\
22 & 0, 3, 1, 6, 7, 8, 9, 12, 10 & 97.9\% \\
23 & 12, 13, 1, 6, 9, 11, 5, 0, 2, 7 & 98.8\% \\
\bottomrule
\end{tabular}
\caption{Mean Ablation Results Across Layers for BM2-CS1}
\label{tab:bm2cs1-attn}
\end{table*}

\paragraph{Efficiency of Model Component Usage.}
Beyond identifying which layers and heads are important, we quantify how much of the model is actually needed for successful task completion. Table~\ref{tab:component-percentage} reports the percentage of total components (e.g., attention heads or MLP neurons) that were retained in each ablation configuration while still achieving high performance. These results indicate that high accuracy can often be recovered using only a small fraction of the model’s total capacity.

\begin{table*}
\centering
\begin{tabular}{lclc}
\toprule
\textbf{Model} & \textbf{MI Method} & \textbf{Component Type} & \textbf{Percentage of Model Components} \\
\midrule
BM1-CS1-Syn         & CS1 & EAP & 12.5\% \\
BM1-CS3-Syn        & CS1 & EAP & 15.62\% \\
BM1-CS3-Syn         & CS2 & EAP & 3.12\% \\
BM1-CS3-Syn         & CS3 & EAP & 15.62\% \\
BM3-CS1-Syn         & CS1 & EAP & 8.3\% \\
BM2-CS1-Syn         & CS1 & EAP & 12.5\% \\

BM1-CS1-Syn     & CS1 & SAE       & 17.26\% \\
BM1-CS3-Syn     & CS1 & SAE       & 22.5\% \\
BM1-CS3-Syn     & CS2 & SAE       & 18.75\% \\
BM1-CS3-Syn     & CS3 & SAE       & 28.97\% \\
BM2-CS1-Syn     & CS1 & SAE       & 60.71\% \\
\bottomrule
\end{tabular}
\caption{Percentage of Model Components Used in Successful Recovery. In EAP, we only consider the percentage of Attention Heads as we only had access to those.}
\label{tab:component-percentage}
\end{table*}

\section{Comparison of EAP-Selected Circuits and Random Circuits}
\label{app:eap_vs_random}
Table \ref{tab:eap_random_acc} presents a comparison between the recovered accuracies of circuits selected by Edge Attribution Patching (EAP) and the average accuracies of randomly selected circuits of comparable size across different model and dataset configurations.

\begin{table*}
\centering
\begin{tabular}{lcc}
\toprule
\textbf{Model} & \textbf{EAP Circuit Accuracy} & \textbf{Average Random Circuit Accuracy} \\
\midrule
BM1-CS1-Syn $\to$ CS1 & 85.4\% & 54.55\% \\
BM1-CS3-Syn $\to$ CS1 & 97.6\% & 82.76\% \\
BM1-CS3-Syn $\to$ CS2 & 74.6\% & 57.83\% \\
BM1-CS3-Syn $\to$ CS3 & 78.3\% & 66.66\% \\
BM2-CS1-Syn $\to$ CS1 & 71.6\% & 30.0\% \\
\bottomrule
\end{tabular}
\caption{Recovered accuracy comparison between EAP-selected circuits and randomly selected circuits of similar size.}
\label{tab:eap_random_acc}
\end{table*}

\section{Logit Lens Analysis Case Study}
\label{appendix:logit-lens-case-study}
We performed a comprehensive logit lens analysis on three representative examples from CS2 and CS3 using the BM2 model (Qwen 0.5B, 24 layers) to better understand how SQL generation emerges across transformer layers.

\textbf{Progressive Token Emergence Pattern:}  
SQL keywords (e.g., \texttt{SELECT}, \texttt{FROM}, \texttt{WHERE}) achieve high confidence (top ranks) in earlier layers (17–21). In contrast, table and column names (e.g., \texttt{orders}, \texttt{order\_id}, \texttt{links}) only emerge with high confidence in the final layer (24).  
\textit{Example:} In Table \ref{tab:token_layer}, across all cases, \texttt{FROM} tokens reach rank 1–2 by layers 21–22, while table names like \texttt{orders} and \texttt{links} remain low until jumping to rank 1 at layer 24.

\textbf{Logit Trajectory Analysis:}  
A systematic reversal in relative logit strengths is observed.  
Early layers (0–16): SQL keywords maintain higher logits than table/column names.  
Later layers: Table/column names surpass SQL keywords in logit values.  
\textit{Example:} In Example 1, \texttt{SELECT} has logits around 0.4–0.5 through layers 8–9, while \texttt{orders} starts near 0.05 and only reaches comparable logits (>\ 0.3) in layer 24.

\textbf{Token-Type Stratification:}  
Three distinct emergence phases were found:  
\begin{itemize}
    \item Basic SQL tokens (\texttt{SELECT}, \texttt{FROM}) stabilize by layers 17–21
    \item Functional tokens (\texttt{COUNT}, \texttt{ORDER BY}) emerge in mid-layers
    \item Context-specific identifiers (table/column names) resolve in the final layer
\end{itemize}
This supports a hierarchical decoding pattern: from general syntax to specific schema grounding.

\textbf{Cross-Example Consistency:}  
The same patterns hold across diverse SQL operations: \texttt{COUNT} aggregation (Example 1), \texttt{MIN} functions (Example 2), and \texttt{ORDER BY} clauses (Example 3). Schema-specific tokens always emerge last (layer 24), while SQL structure stabilizes earlier.

\paragraph{Examples Analyzed}

\begin{itemize}  
  \item \textbf{Example 1} \\
  Instruction: In \texttt{orders}, list total count amount and instances of duration \\
  Context: \texttt{CREATE TABLE orders (duration TIME, amount INT)} \\
  Response: \texttt{SELECT COUNT(amount) AS COUNT\_amount, COUNT(duration) AS COUNT\_duration FROM orders}

  \item \textbf{Example 2} \\
  Instruction: Search for least duration in \texttt{links} beginning with the most duration \\
  Context: \texttt{CREATE TABLE links (price INT, duration TIME)} \\
  Response: \texttt{SELECT MIN(duration) AS MIN\_duration FROM links}

  \item \textbf{Example 3} \\
  Instruction: Show me the duration and size from the \texttt{orders} table ordered by duration in ascending order \\
  Context: \texttt{CREATE TABLE orders (duration TIME, size INT)} \\
  Response: \texttt{SELECT duration, size FROM orders ORDER BY duration ASC}
\end{itemize}

The results are summarized in Table \ref{tab:token_layer}

\paragraph{Summary Table: Layer Where Token First Achieves Top Rank}

\begin{table*}
\centering
\begin{tabular}{lccc}
\toprule
\textbf{Token Type} & \textbf{Example 1} & \textbf{Example 2} & \textbf{Example 3} \\
\midrule
FROM & Layer 21 & Layer 21 & Layer 22 \\
Aggregation (MIN, COUNT) & Layer 17 & Layer 21 & N/A \\
AS & Layer 24 & Layer 24 & N/A \\
GROUP BY & N/A & N/A & Layer 17 \\
Table Name (links, orders) & Layer 24 & Layer 24 & Layer 24 \\
Column Name & Layer 24 & Layer 24 & Layer 24 \\
\bottomrule
\end{tabular}
\caption{Layer where each token type first achieves top rank.}
\label{tab:token_layer}
\end{table*}

\paragraph{Logit Progression Pattern}  

\begin{itemize}
    \item Early Layers (0–16): SQL keywords dominate with a positive logit advantage
    \item Mid Layers (17–21): SQL structure tokens reach high ranks, schema tokens remain low
    \item Final Layer (24): Schema-specific tokens (e.g., table/column names) reach top logit values
\end{itemize}

This progression supports the hypothesis that transformer layers follow a hierarchical reasoning process, from abstract SQL syntax to grounded schema resolution.

We also present in Table \ref{tab:count_orders_rank}, a detailed analysis of the SQL Keyword \texttt{COUNT} ranks across layers compared to the table name \texttt{orders} , 


\begin{table}[htbp]
\centering
\begin{tabular}{ccc}
\toprule
Layer & COUNT Rank & orders Rank \\
\midrule
1 & 392 & 43497 \\
2 & 832 & 71069 \\
3 & 1032 & 67813 \\
4 & 5652 & 83330 \\
5 & 7892 & 118378 \\
6 & 4532 & 85341 \\
7 & 2209 & 112749 \\
8 & 3505 & 136323 \\
9 & 2101 & 144139 \\
10 & 2065 & 136692 \\
11 & 4047 & 136514 \\
12 & 2676 & 132739 \\
13 & 384 & 127971 \\
14 & 92 & 108117 \\
15 & 124 & 76418 \\
16 & 1 & 79962 \\
17 & 1 & 33234 \\
18 & 1 & 37628 \\
19 & 1 & 25178 \\
20 & 1 & 24058 \\
21 & 1 & 7664 \\
22 & 2 & 2430 \\
23 & 2 & 166 \\
24 & 1 & 1 \\
\bottomrule
\end{tabular}
\caption{Token rank comparison between \texttt{COUNT} and \texttt{orders} across layers for Example 1.}
\label{tab:count_orders_rank}
\end{table}

\section{Additional LogitLens Visualizations}
\label{appendix:logitlens}

To further support our interpretation of two-phase SQL query generation, we include additional LogitLens visualizations across other benchmarks and model configurations.

In each case, we consistently observe the same dynamics: an early spike in the logit probabilities of SQL-specific keywords, followed by a later emergence of table name tokens, indicating a transition from syntactic intent planning to context-aware content grounding.

These findings generalize across schema types and query contexts, reinforcing the idea that language models internally decompose structured generation tasks into distinct stages aligned with human-interpretable reasoning processes.

\begin{figure}
    \centering
    \begin{subfigure}[b]{\linewidth}
        \includegraphics[width=\linewidth]{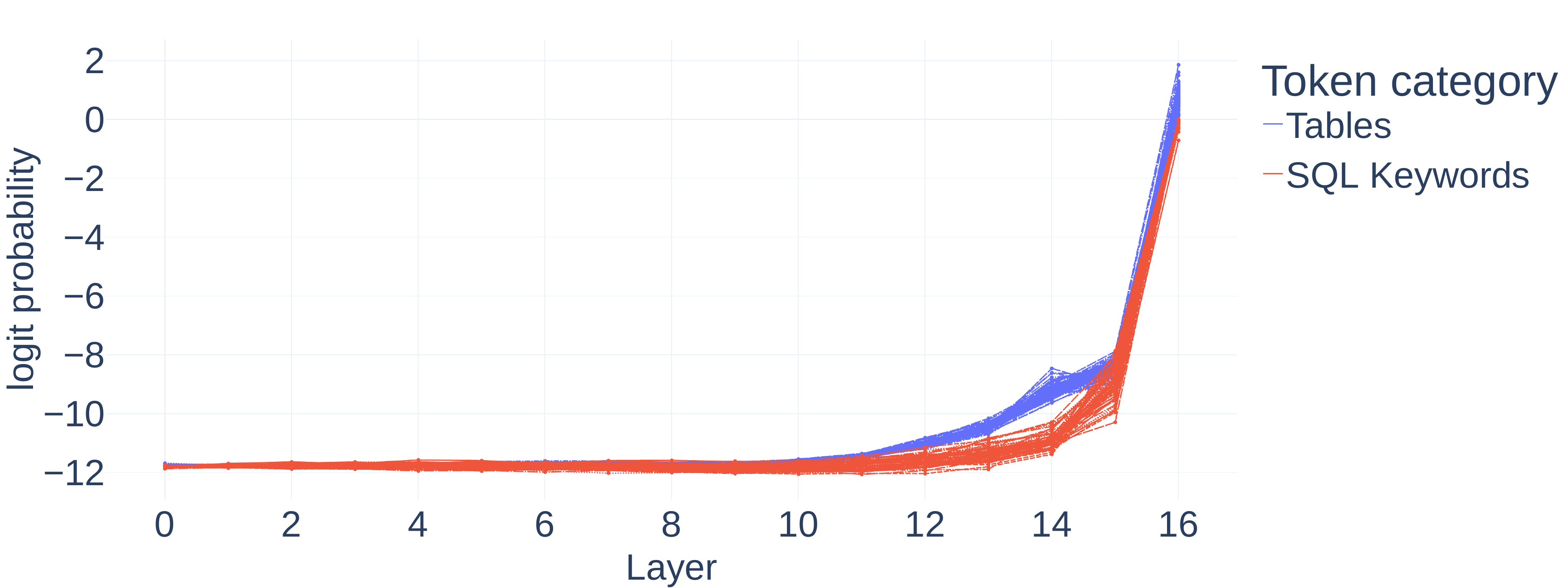}
        \caption{BM3-CS3}
        \label{fig:logitlens-fig1}
    \end{subfigure}
    \hfill
    \begin{subfigure}[b]{\linewidth}
        \includegraphics[width=\linewidth]{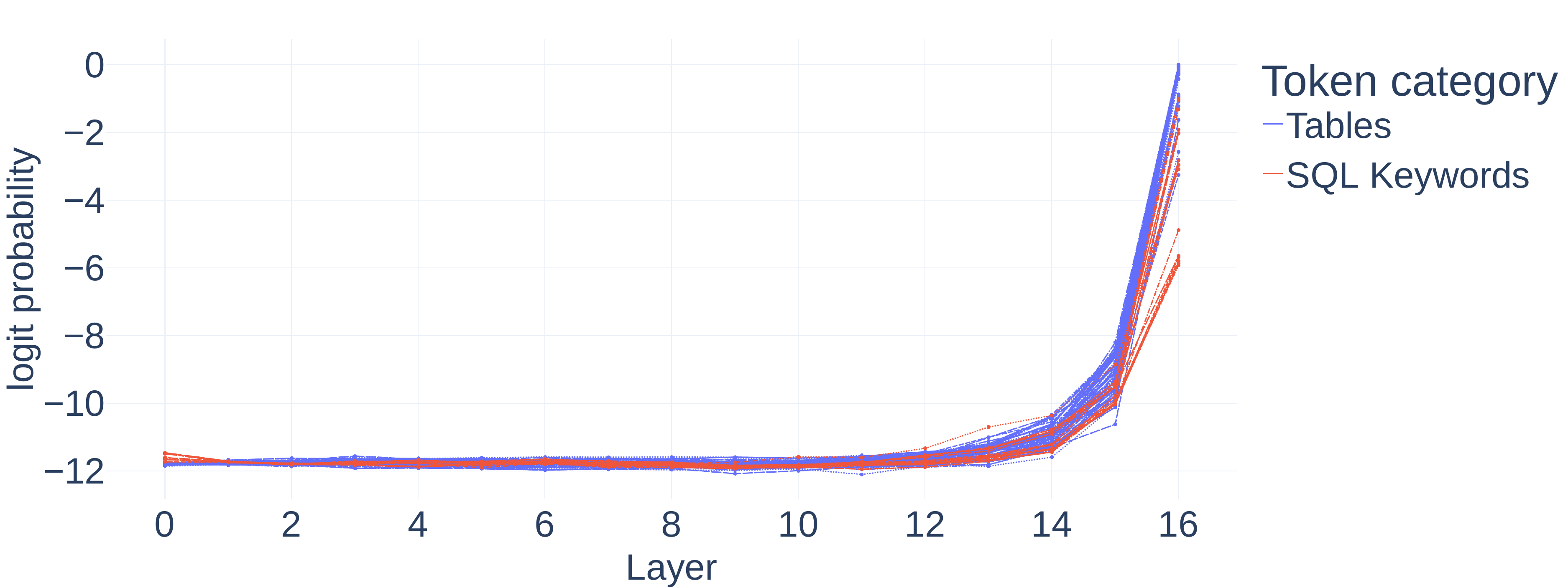}
        \caption{BM3-CS1}
        \label{fig:logitlens-fig2}
    \end{subfigure}
    
    \caption{Additional LogitLens visualizations across six benchmarks, illustrating consistent two-phase behavior: early SQL intent formation followed by table resolution.}
    \label{fig:logitlens-grid}
\end{figure}

\clearpage
\onecolumn

\end{document}